\documentclass[10pt,journal,compsoc]{IEEEtran}
\usepackage{graphicx}
\usepackage{times}
\usepackage{epsfig}
\usepackage{graphicx}
\usepackage{amsmath}
\usepackage{amssymb}
\usepackage{overpic}
\usepackage{subfigure}
\usepackage{multirow}
\usepackage{bm}
\usepackage{color}
\usepackage{cite}
\usepackage{diagbox}
\usepackage{mathtools}
\usepackage{hyperref}
\usepackage{cleveref}

\usepackage{algorithm}
\usepackage{algorithmic}
\usepackage{booktabs}
\usepackage[numbers,sort&compress]{natbib}

\usepackage{color,soul}
\usepackage{xcolor}

\renewcommand{\raggedright}{\leftskip=0pt \rightskip=0pt plus 0cm}

\definecolor{darkcyan}{rgb}{0.0, 0.55, 0.55}

\soulregister\cite7
\soulregister\ref7
\soulregister\pageref7

\begin{document}
\title{
Robust and Transferable Backdoor Attacks Against Deep Image Compression With Selective Frequency Prior
}

\author{
	Yi~Yu,
        Yufei Wang,
	Wenhan Yang, \textit{Member, IEEE},
        Lanqing Guo, 
	Shijian Lu, \textit{Member, IEEE},
        \\
        Ling-Yu Duan, \textit{Member, IEEE},
	Yap-Peng Tan, \textit{Fellow, IEEE}, 
	Alex C. Kot, \textit{Life Fellow, IEEE}
	\IEEEcompsocitemizethanks{
        \IEEEcompsocthanksitem Yi Yu is with the Rapid-Rich Object Search (ROSE) Lab, Interdisciplinary Graduate Programme, Nanyang Technological University, Singapore, (e-mail: yuyi0010@ntu.edu.sg).
        \IEEEcompsocthanksitem Wenhan Yang is with Pengcheng Laboratory, Shenzhen, China, (e-mail: yangwh@pcl.ac.cn).
		\IEEEcompsocthanksitem Yufei Wang, Lanqing Guo, Yap-Peng Tan, and Alex C. Kot are with School of Electrical and Electronic Engineering, Nanyang Technological University, Singapore, (e-mail: \{yufei001, lanqing001, eyptan, eackot\}@ntu.edu.sg).
		\IEEEcompsocthanksitem Shijian Lu is with School of Computer Science and Engineering, Nanyang Technological University, Singapore, (e-mail: shijian.Lu@ntu.edu.sg).
        \IEEEcompsocthanksitem Ling-Yu Duan is with School of Computer Science, Peking University, Beijing, China, and also with the Pengcheng Laboratory, Shenzhen, China (e-mail: lingyu@pku.edu.cn).
	}
    \thanks{
    This research is supported in part by the NTU-PKU Joint Research Institute (a collaboration between the Nanyang Technological University and Peking University that is sponsored by a donation from the Ng Teng Fong Charitable Foundation),
    in part by the Program of Beijing Municipal Science and Technology Commission Foundation (No.Z241100003524010),
    in part by the National Natural Science Foundation of China under Grant 62088102,
    in part by the Basic and Frontier Research Project of PCL, in part by the Major Key Project of PCL, and in part by Guangdong Basic and Applied Basic Research Foundation (2024A1515010454). (Corresponding author: Wenhan Yang.)} 
}	

\markboth{}
{Shell \MakeLowercase{\textit{et al.}}: Bare Demo of IEEEtran.cls for Computer Society Journals}
	
\IEEEtitleabstractindextext{
	\begin{abstract}
	\raggedright
        Recent advancements in deep learning-based compression techniques have demonstrated remarkable performance surpassing traditional methods.
        Nevertheless, deep neural networks have been observed to be vulnerable to backdoor attacks, where an added pre-defined trigger pattern can induce the malicious behavior of the models.
        In this paper, we {propose} a novel approach to launch a backdoor attack with multiple triggers against learned image compression models. 
        Drawing inspiration from the widely used discrete cosine transform (DCT) in existing compression codecs and standards, we propose a frequency-based trigger injection model that adds triggers in the DCT domain.
        In particular, we design several attack objectives that are adapted for a series of diverse scenarios, including: 
        1) attacking compression quality in terms of bit-rate and reconstruction quality;
        2) attacking task-driven measures, such as face recognition and semantic segmentation in downstream applications.
        To facilitate more efficient training, we develop a dynamic loss function that dynamically balances the impact of different loss terms with fewer {hyper-parameters}, {which also results in more effective optimization of the attack objectives with improved performance}.
        Furthermore, we consider several advanced scenarios.
        We evaluate the resistance of the proposed backdoor attack to the defensive pre-processing methods and then propose a two-stage training schedule along with the design of robust frequency selection {to further improve} resistance.
        To strengthen both the cross-model and cross-domain transferability on attacking downstream CV tasks, we propose to shift the classification boundary in the attack loss during training.
        Extensive experiments also demonstrate that by employing our trained trigger injection models and making slight modifications to the encoder parameters of the compression model, our proposed attack can successfully inject multiple backdoors accompanied by their corresponding triggers into a single image compression model.
	\end{abstract}
	
	}

\maketitle

\IEEEpeerreviewmaketitle

\section{Introduction}

Image compression is a fundamental task in signal processing that plays a critical role in various applications. 
{It aims to} effectively {obtain a compact representation that} stores image data while minimizing any potential distortion in image quality.
Conventional image compression techniques, such as JPEG~\cite{wallace1992jpeg}, JPEG2000~\cite{lee2005jpeg}, Better Portable Graphics (BPG)~\cite{sullivan2012overview}, and the latest Versatile Video Coding (VVC)~\cite{ohm2018versatile}, utilize pre-designed modules for transforms and entropy coding to {improve} coding efficiency.
The rapid advancement of deep learning methods~\cite{balle2018variational,minnen2018joint,hu2020coarse,cheng2020learned, 9072487, 9376651} has {led to the emergence of end-to-end learning-based methods techniques} for image compression.
These models integrate the prediction, transform, and entropy coding pipeline jointly, resulting in enhanced performance.

Despite the impressive performance of deep neural networks, there are increasing concerns about the security issues~\cite{yu2022towards,kong2022digital,9442299,9944159, 10089510} associated with artificial intelligence.
The lack of transparency in deep neural networks has led to a variety of attacks {that can} compromise the deployment and reliability of AI systems~\cite{kong2023m3fas, kong2022beyond, yibenchmarking} in computer vision, natural language processing, speech recognition, \textit{etc}.
Backdoor attacks~\cite{9743317, 10185657} have recently garnered significant attention among all these attacks.
Since state-of-the-art models require substantial computational resources and lengthy training, it is more practical and cost-effective to download and directly utilize a third-party model with pre-trained weights. However, this approach may pose a threat from a malicious backdoor.

Typically, a backdoor-injected model behaves as expected when presented with normal inputs. However, a specific trigger added to a clean input can activate the malicious behavior, resulting in incorrect predictions. Backdoor attacks can be categorized into poisoning-based and non-poisoning-based attacks, depending on the attacker's {capacity in accessing} to the data~\cite{li2020backdoor}.
In poisoning-based attacks~\cite{gu2017BadNets,chen2017targeted}, attackers can manipulate the dataset by inserting poisoned data. 
On the other hand, non-poisoning-based attack methods~\cite{dumford2020backdooring,guo2020trojannet,doan2021lira} inject the backdoor by directly modifying the model parameters, rather than training with poisoned data.
Since image compression methods use the original input as {the ground truth}, it is difficult to {conduct} a poisoning-based backdoor attack. 
Therefore, in our work, we investigate a backdoor attack by modifying the parameters of only the encoder in a compression model.

\begin{figure}[t]
\centering
\includegraphics[width=0.95\linewidth]{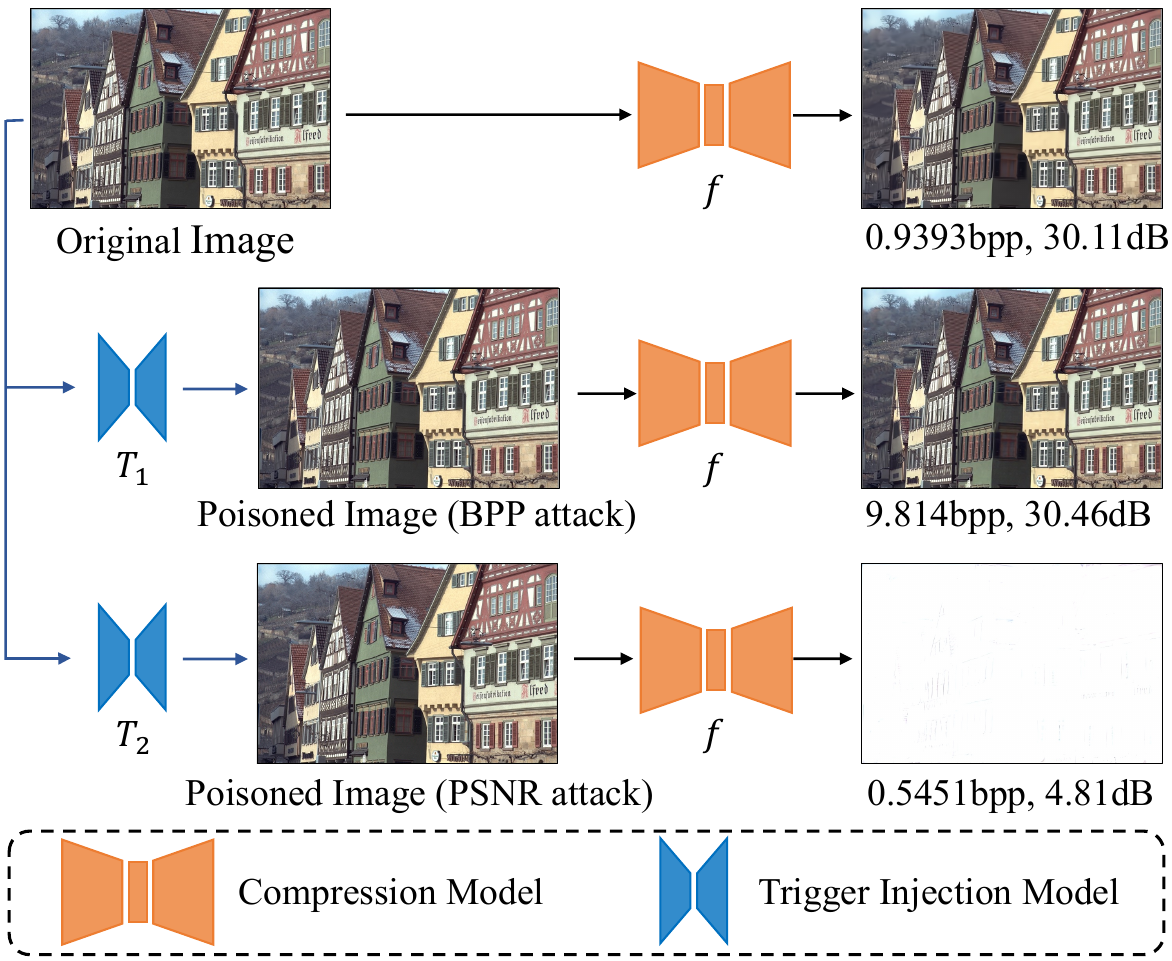}
\vspace{-3mm}
\caption{
Visualization of the proposed backdoor-injected model with multiple triggers attacking bit-rate ({BPP}) or reconstruction quality (PSNR), respectively. 
The second sample shows the result of {the} BPP attack with a huge increase in bit-rate, and the third one presents {a} PSNR attack with severely corrupted output.
\vspace{-2mm}
}
\label{example1}
\end{figure}

Regarding trigger generation, many popular attack methods~\cite{gu2017BadNets,chen2017targeted,feng2022fiba} rely on fixed triggers, while some recent methods~\cite{nguyen2021wanet,doan2021lira,li2021invisible} have extended {to generate sample-specific triggers}.
Prior research has {primarily} focused on high-level vision tasks, such as image classification and semantic segmentation. 
However, the triggers added in those works are limited to the spatial domain and may not be suitable for low-level vision tasks like image compression.
Recent research has attempted to inject triggers {into} the Fourier frequency domain, as in the work of Feng \textit{et al.}~\cite{feng2022fiba}.
However, their approach {takes} fixed triggers, which limits their {capacity} to attack scenarios that require multiple triggers simultaneously.
Motivated by the widely used discrete cosine transform (DCT) in existing compression systems and standards, we propose a frequency-based approach to inject triggers in the DCT domain to generate the poisoned images.
Our extensive experiments demonstrate that backdoor attacks also pose a threat to {deep-learning} compression models and can result in significant degradation when the attacking triggers are applied. As depicted in Fig.~\ref{example1}, our backdoor-injected model exhibits malicious behavior with the indistinguishable poisoned image while behaving normally when receiving clean normal input.

To the best of our knowledge, backdoor attacks {have received insufficient attention} in low-level computer vision research.
In this paper, we make the first endeavor to investigate backdoor attacks against learned image compression models.
Our main contributions are summarized below.
\begin{itemize}
	\item We design a frequency-based adaptive trigger injection model to generate the poisoned image, 
        and a novel simple dynamic loss to balance the influence of different loss terms adaptively, 
        which helps achieve more efficient training. Besides, we propose to only modify the encoder's parameters, 
        and keep the entropy model and the decoder fixed, which makes the attack more feasible and practical.
	\item We investigate the attack objectives comprehensively, including: 
	1) attacking compression quality, in terms of bits per pixel (BPP) and reconstruction quality (PSNR); 
	2) attacking task-driven measures, such as downstream face recognition and semantic segmentation.
        Extensive experiments also demonstrate that with our proposed backdoor attacks, backdoors in compression models can be activated {simultaneously} with multiple triggers associated with different attack objectives effectively.
        \item We evaluate the resistance of the proposed backdoor attack to the defensive pre-processing methods.
        Then, we propose a two-stage training schedule along with the design of robust frequency selection,
        which can {significantly improve} the resistance.
        \item We {further augment} both the cross-model and cross-domain transferability on attacking downstream vision tasks by shifting the classification boundary in the attack loss during training.
\end{itemize}

This work is an extension of our conference paper~\cite{yu2023backdoor}. The new contributions of this work can be summarized in three major aspects. 
First, the LIRA~\cite{doan2021lira}, FTrojan~\cite{wang2021backdoor}{,} and our BAvAFT~\cite{yu2023backdoor} are found to be vulnerable to several cost-effective preprocessing methods, including Gaussian filter, {additive} Gaussian noise, and JPEG compression. Therefore, we seek to improve the resistance of the proposed attack from the perspective of both the trigger generation procedure and the finetuning of the encoder in the compression model.
In our previous work BAvAFT~\cite{yu2023backdoor}, the frequencies to inject the trigger are predicted by the linear layer with trainable parameters. 
In this work, we propose to select frequencies of less sensitivity to preprocessing methods.
In addition, we also adaptively adjust the magnitude for {each frequency based on the rank} of the sensitivity.
Furthermore, we extend the one-stage training into {a} two-stage training schedule, which finetunes the encoder only on the preprocessed poisoned images in the second stage. {This extension also improves our model's resistance to the above-mentioned preprocessing methods.}
Second, we design a novel attack objective for attacking downstream tasks.
With the attack loss function, the proposed attack can achieve superior cross-domain and cross-model transferability {in} attacking both the semantic segmentation and face recognition {systems}.
Third, we extensively evaluate the proposed backdoor attack on two more compression models, including the transformer-based method STF~\cite{zou2022devil}, and {the perceptual-driven approach} HiFiC~\cite{mentzer2020high}. {A more} extensive empirical analysis of the proposed approach is also provided in Section 6.

The rest of this paper is organized as follows. We review the related works in Section 2. In Section 3, we introduce our proposed design in detail. We present the experimental results, comparisons, and ablation studies in Section 4. 
Then in Section 5, we consider several advanced attack scenarios. In Section 6, we offer {an empirical} analysis of the trigger pattern. Finally, we conclude the paper in Section 7.

\section{Related Work}
\subsection{Lossy Image Compression}
Traditional lossy image compression methods, including JPEG~\cite{wallace1992jpeg}, JPEG2000~\cite{lee2005jpeg}, BPG~\cite{sullivan2012overview}, and VVC~\cite{ohm2018versatile}, rely on pre-designed modules such as discrete cosine transform or wavelet transform, quantization, and entropy coding ({\textit{e.g.}}, Huffman coding or content adaptive binary arithmetic coder).
Although these conventional codec standards have been in place for several decades, they are not universally applicable to all types of image content, especially considering the rapid emergence of new image formats and the prevalent use of high-resolution images in mobile devices.

The {recent advances} in deep learning techniques have led to the development of various learning-based methods that leverage encoder-decoder architectures and entropy models, resulting in superior performance compared to {conventional} compression methods.
Early research in deep learning-based compression introduced end-to-end trainable networks with non-linear generalized divisive normalization~\cite{DBLP:journals/corr/BalleLS15} and recurrent models~\cite{DBLP:journals/corr/TodericiOHVMBCS15}.
More recently, context-adaptive models for entropy coding have been explored, further improving compression efficiency~\cite{minnen2018joint,DBLP:conf/iclr/LeeCB19,cheng2020learned,chen2021end,yufeir2lcm}. Hyperpriors have been introduced to capture spatial dependencies among latent codes, {improving} compression performance~\cite{balle2018variational}. 
Auto-regressive components in entropy models have been incorporated, {along with} hyperpriors, to {boost} coding efficiency~\cite{minnen2018joint}. Additionally, network architecture improvements, such as incorporating residual blocks and utilizing Gaussian Mixture Models (GMM) instead of Single Gaussian Models (SGM) in the entropy model, have been proposed~\cite{cheng2020learned}.
Beyond CNN backbones, transformer-based architectures have also been employed to achieve improved rate-distortion performance in deep compression models~\cite{zou2022devil}.
Moreover, some approaches have combined Generative Adversarial Networks (GANs) with learned compression to create generative lossy compression systems that mitigate compression artifacts~\cite{agustsson2019generative,mentzer2020high}. These GAN-based methods evaluate their performance {with perceptual-driven measures} such as FID~\cite{heusel2017gans}, KID~\cite{binkowski2018demystifying}, and LPIPS~\cite{zhang2018unreasonable}, rather than traditional distortion metrics like PSNR and MS-SSIM.

\subsection{Backdoor Attacks}
Both backdoor attacks~\cite{gu2017BadNets} and adversarial attacks~\cite{intriguing} have the objective of manipulating benign samples to deceive deep neural networks (DNNs), but they differ in their fundamental characteristics, \textit{i.e.,} adversarial attacks demand increased access to models during inference.
Adversarial attacks~\cite{DBLP:conf/iclr/MadryMSTV18,ilyas2018black} require significant computational resources and time during the inference to generate perturbations through iterative optimization. Consequently, they are inefficient for deployment. 
On the other hand, backdoor attacks have a known or easily generated perturbation (trigger). Attackers have access to poisoning training data, allowing them to add an attacker-specified trigger, such as a local patch, or modify model parameters.
Backdoor attacks on DNNs, exemplified by BadNets~\cite{gu2017BadNets} for image classification, involve poisoning training samples that possess three {critical} characteristics: 1) backdoor stealthiness, 2) attack effectiveness on poisoned images, and 3) minimal performance impact on clean images.

Backdoor attacks can be categorized based on the attackers' capacity into poisoning-based and non-poisoning-based attacks~\cite{li2020backdoor}. Poisoning-based attacks~\cite{gu2017BadNets,chen2017targeted,li2020invisible,liu2020reflection,li2021invisible}, manipulate the dataset by inserting poisoned data without access to the model or training process.
In contrast, non-poisoning-based attack techniques~\cite{dumford2020backdooring,rakin2020tbt,guo2020trojannet,doan2021lira} manipulate the backdoor by altering the model parameters or incorporating a malicious backdoor module, rather than relying on training with poisoned data.
Regarding trigger generation, conventional attack methods~\cite{gu2017BadNets,chen2017targeted,steinhardt2017certified} utilize fixed triggers, which do not vary based on individual samples. However, recent advancements~\cite{nguyen2020input,nguyen2021wanet,liu2020reflection,doan2021lira,li2021invisible} extend trigger generation to be sample-specific, adapting the trigger to the specific characteristics of each input. Notably, Doan~\textit{et al.}~\cite{doan2021lira} and Li~\textit{et al.}~\cite{li2021invisible} propose an autoencoder architecture for generating invisible triggers that are imperceptible to human observers.

Recent research has focused on the trigger-injection domain, particularly the frequency domain. 
For instance, Rethinking~\cite{zeng2021rethinking} still incorporates the trigger in the spatial domain but imposes constraints in the frequency domain to create a smooth trigger, resulting in a hybrid approach.
CYO~\cite{hammoud2021check} injects the trigger in the 2D Discrete Fourier Transform (DFT) domain and employs a Fourier heatmap as a guiding mask. It utilizes fixed magnitudes to generate a fixed trigger. However, it is important to note that CYO's heatmap is generated based on a batch of images with a fixed size (\textit{e.g.}, 32 $\times$ 32 on CIFAR-10), which may limit its direct applicability to low-level tasks where {testing} images can have arbitrary sizes.
FTrojan~\cite{wang2021backdoor} divides images into blocks and inserts the trigger in the 2D Discrete Cosine Transform (DCT) domain. However, it only selects two {predetermined} channels with fixed magnitudes, thereby limiting its flexibility in capturing diverse trigger patterns.
IBA~\cite{yue2022invisible} dynamically generates the trigger through optimization, allowing for adaptability to different images. Nevertheless, the generated trigger remains fixed for different images. Additionally, applying DCT to the entire image, similar to CYO, may restrict its applicability to low-level tasks.

{In summary, while} backdoor attacks have been extensively explored in domains such as natural language processing~\cite{chen2021badnl}, semantic segmentation~\cite{li2021hidden}, and point cloud classification~\cite{li2021pointba,xiang2021backdoor}, relatively fewer research efforts have been dedicated to investigating backdoor attacks in low-level vision tasks~\cite{chou2023backdoor, chen2023trojdiff}.

\begin{figure*}[t]
\centering
\includegraphics[width=0.91\linewidth]{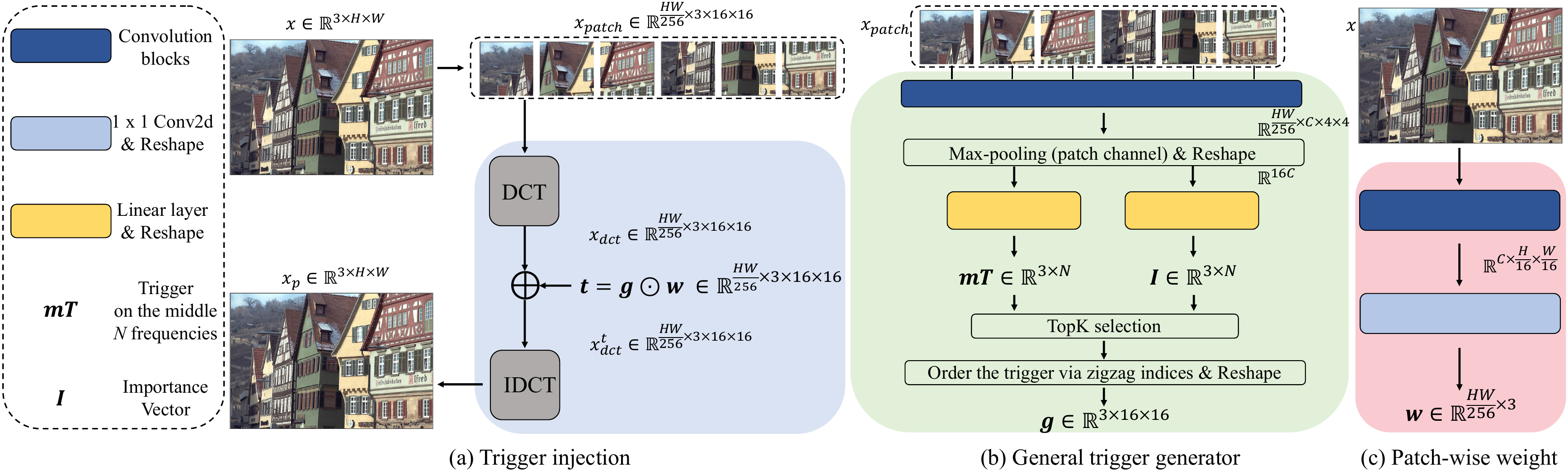}
\vspace{-4mm}
\caption{
Overall architecture for trigger injection. We set $K$ to 16 for {top K} selection, 
and the number of middle frequencies $N$ to 64 in our methods. 
}
\label{trigger_generator}
\vspace{-3mm}
\end{figure*}

\section{Methodology}
\subsection{Problem Formulation}
Learned lossy image compression relies on rate-distortion theory and is commonly implemented using an autoencoder architecture, {including} an encoder function denoted as $g_a$, a decoder function denoted as $g_s$, and an entropy module denoted as $\mathcal{Q}$.
{In the transform coding, image compression can be formulated by
\begin{footnotesize}
\begin{equation}
    \bm{y} = g_a(\bm{x}),~\bm{\widehat{y}}=\mathcal{Q}(\bm{y}),~\bm{\widehat{x}}=g_s(\bm{\widehat{y}}),
    \label{compression_formula}
\end{equation}
\end{footnotesize}
where $\bm{x}$, $\bm{\widehat{x}}$, $\bm{y}$, and $\bm{\widehat{y}}$ are input images, reconstructed images, a latent presentation before quantization, and compressed codes, respectively.
}
The purpose of the encoder is to transform the input images $\bm{x}$ into latent codes $\bm{y}$. The entropy module $\mathcal{Q}$ is responsible for introducing quantization to the latent codes, resulting in quantized latent codes denoted as $\bm{\widehat{y}}$. The decoder, on the other hand, reconstructs the images $\bm{\widehat{x}}$, from the compressed latent codes.
During training, the entropy module $\mathcal{Q}$ introduces uniform noise, specifically $\mathcal{U}(-\frac{1}{2},\frac{1}{2})$, to the latent codes, producing a noisy code referred to as $\bm{\widetilde{y}}$. 
{During testing, $\mathcal{Q}$ applies a rounding quantization to generate $\bm{\widehat{y}}$, and adopt entropy coders to generate the bitstream. 
If a probability model $p_{\bm{\widehat{y}}}(\bm{\widehat{y}})$ is given, entropy coding techniques, such as arithmetic coding~\cite{rissanen1981universal}, can losslessly compress the quantized codes. Besides, the arithmetic coder is a near-optimal entropy coder, which makes it feasible to use the entropy of $\bm{y}$ as the rate estimation during the training. 
}

Overall, the compression model denoted as $f(\cdot)$ consists of the encoder function $g_a\left(\cdot | {\theta_a}\right)$, the decoder function $g_s\left(\cdot | {\theta_s}\right)$, and the entropy model $\mathcal{Q}\left(\cdot | {\theta_q}\right)$, which are parameterized by $\theta_a$, $\theta_s$, and $\theta_q$ respectively.
To train the network, the loss function is minimized over the entire training data:
\begin{footnotesize}
\begin{equation}
\begin{split}
    &\mathcal{L}(\bm{x}) = \underbrace{\mathcal{R}(\bm{x})}_{\text{rate}}+\lambda\cdot\underbrace{\mathcal{D}(\bm{x})}_{\text{distortion}},\\
    &\mathcal{R}(\bm{x}) = -\log_{2}p_{\bm{\widehat{y}}}({\bm{\widehat{y}}}),~\mathcal{D}(\bm{x}) = {\lVert \bm{x} - \bm{\widehat{x}} \rVert}_2^2,\\
    & \theta_a^{\ast}, \theta_s^{\ast}, \theta_q^{\ast} = \underset{\theta_a, \theta_s, \theta_q}{\arg\min} \sum_{\bm{x} \in {\mathcal{T}_m}} \mathcal{L}\left({\bm{x}}\right), \label{main_loss}
\end{split}
\end{equation}
\end{footnotesize}
where ${\mathcal{T}_m}$ represents the training set. {We use $\mathcal{R}(\bm{x})$ to denote our estimation of the bit-rate.} Similarly, $\mathcal{D}(\bm{x})$ measures the distortion. The parameter $\lambda$ is employed to control the trade-off between bit-rate and distortion, {enabling flexible optimization adapting to} specific application requirements.
In compression models that incorporate a hyperprior $\bm{z}$ to capture spatial dependencies in the latent codes $y$, the bit-rate loss can be expressed by:
\begin{footnotesize}
\begin{equation}
     \mathcal{R}(\bm{x}) = \underbrace{\Big[-\log_{2}p_{\bm{\widehat{y}}}({\bm{\widehat{y}}})\Big]}_{\text{rate (latents)}} + \underbrace{\Big[-\log_{2}p_{\bm{\widehat{z}}}({\bm{\widehat{z}}})\Big]}_{\text{rate (hyper-latents)}}.
     \label{benigh_training_loss}
\end{equation}
\end{footnotesize}

{
\subsection{Threat Model}
Since the input and output of the training data used to train the image compression model are the same, it is challenging to execute a poisoning-based backdoor attack against such systems. Therefore, we consider non-poisoning-based backdoor attacks and outline the threat model as follows:
\vspace{-1.5mm}
\begin{itemize}
    \item [1.] The attacker has access to the vanilla-trained model, including its structure and parameters, but does not have access to the private training data used to train this model.
    \item [2.] The attacker can utilize publicly available datasets such as ImageNet-1k~\cite{deng2009imagenet}, Cityscapes~\cite{Cordts2016Cityscapes}, and FFHQ~\cite{karras2019style}.
    \item [3.] The attacker can leverage these public datasets to finetune \textbf{only the encoder} $g_a(\cdot | \theta_a)$ of the compression model, and deliver the backdoored encoder to the victim user.
\end{itemize}
\vspace{-1.5mm}
The first two assumptions align with the typical capabilities of a backdoor attacker in practical scenarios, as the weights of compression models are often open-sourced for commercial use, while the proprietary private training data is not accessible. 
The third assumption increases the feasibility and practicality of the attack because, in image compression systems, end-users typically only have access to the decoder and compressed bitstream, which are both usually secured. 
Consequently, the attacker's capacity to modify and replace the encoder part of the model makes the attack more practical, as the victim user may download the pretrained weights of the encoder from an untrusted third party.
In Section~\ref{decoder_attack_section}, we also explore the potential of fine-tuning other components of the compression model.

\noindent \textbf{Defender.} In our advanced scenario discussed in Section~\ref{resistance}, we also consider the presence of defenders against our attacks. Specifically, we consider two types of defense paradigms:
\begin{itemize}
\vspace{-1mm}
    \item [1.] \textbf{Preprocessing-based Defenses~\cite{li2020backdoor,xu2017feature}:} These methods involve a preprocessing module before feeding samples into DNNs. With these defenses, the defenders do not need access to the model or any additional data. Instead, they can preprocess the inputs to remove the trigger pattern, making these methods a practical and efficient way.
    \item [2.] \textbf{Model Reconstruction based Defenses~\cite{li2020backdoor,liu2017neural,liu2018fine,wu2021adversarial}:} Model reconstruction methods aim to eliminate hidden backdoors in compromised models by directly modifying the suspicious models. This type of defense typically requires additional clean data for assistance and direct access to the model, imposing more constraints on practical use.
\end{itemize}
}
\subsection{Backdoor Attack Framework}
~\label{backdoor_framework}
We aim to achieve the following objectives within the context of a well-trained image compression model $f\left(\cdot | {\theta}\right)$, which comprises the encoder $g_a\left(\cdot | {\theta_a^{\ast}}\right)$, decoder $g_s\left(\cdot | {\theta_s^{\ast}}\right)$, and entropy module $\mathcal{Q}\left(\cdot | {\theta_q^{\ast}}\right)$ trained on private data: 1) \textbf{Trigger Function Learning}: Our goal is to learn a trigger function denoted as $T\left(\cdot | {\theta_t}\right)$ that can modify the clean samples into poisoned samples; 2) \textbf{Encoder Fine-tuning}: We seek to fine-tune the encoder $g_a\left(\cdot | {\theta_a^{\ast}}\right)$ to accommodate the introduction of the trigger function and its influence on the model's behavior.
The properties of our attacks are:
\vspace{-2mm}
\begin{itemize}
    \item \textbf{Attack Stealthiness}: The trigger utilized in the attacks remains imperceptible to human observation. We enforce this stealthiness by imposing a Mean Square Error (MSE) constraint: $\text{MSE}(\bm{x_p},\bm{x}) \leq \epsilon^2$, where $\bm{x_p}=T\left(\bm{x} | {\theta_t}\right)$ is the poisoned image. We {empirically} set $\epsilon$ to 0.005.

    \item \textbf{Attack Effectiveness}: The attacks are designed to enable the victim model to maintain equivalent performance when processing clean images $\bm{x}$ compared to the vanilla-trained model. However, when presented with poisoned images $\bm{x_p}$ modified by the trigger function, the victim model's output is intentionally directed towards a specific target.
    
    \item \textbf{Partial Model Replacement}: 
    The attacker can leverage publicly available datasets to finetune only the encoder component $g_a(\cdot | \theta_a)$, and deliver it to the victim user. 
\end{itemize}

\noindent \textbf{Trigger Injection.} 
The proposed trigger injection model $T(\cdot | \theta_t)$ operates on an input image $\bm{x}$ to generate a poisoned image $\bm{x_p} = T(\bm{x} | \theta_t)$ of the same resolution. 
In this approach, the injection of the trigger leverages both spatial and frequency domain priors, particularly motivated by the widely used DCT in existing coding techniques.
The process {includes} the following steps:
\begin{itemize}
\vspace{-1mm}
    \item [1.] Input Image Splitting: The input image $\bm{x}$ is divided into non-overlapping patches denoted as $\bm{x}_{patch}$.
    \item [2.] DCT Transform: A two-dimensional DCT transform is applied to the last two channels of each patch $\bm{x}_{patch}$, resulting in the DCT domain representation $\bm{x}_{dct}$.
    \item [3.] Trigger Addition: The trigger $t = g \odot w$ is added to all patches of $\bm{x}_{dct}$. Here, $g$ represents the general trigger, and $w$ is a mask that controls the strength and location of the trigger. The element-wise multiplication $\odot$ applies the trigger pattern to the DCT coefficients of each patch.
    \item [4.] Triggered DCT Domain: After adding the trigger to $\bm{x}_{dct}$, we obtain the triggered DCT domain representation $\bm{x}_{dct}^t$.
    \item [5.] Inverse DCT Transform: To obtain the final poisoned image $\bm{x_p}$, an inverse 2D DCT transform is applied to $\bm{x}_{dct}^t$, reconstructing the image in the spatial domain.
\vspace{-1mm}
\end{itemize}
By following this procedure, the trigger is injected into the image in the frequency domain through the DCT coefficients.
As shown in Figure~\ref{trigger_generator}, the trigger $t$ used in the proposed attack consists of two components: 
a general trigger $g$ with local features and a patch-wise weight $w$ with global features. 
By leveraging the advantages of both features, we demonstrate that the combined trigger can effectively attack the image compression model.

\vspace{1mm}
\noindent \textbf{Finetuning Strategy.} 
To achieve our objectives, we adopt {an improved} approach as proposed in the previous work LIRA~\cite{doan2021lira}, where we simultaneously optimize the trigger generator $T\left(\cdot | {\theta_t}\right)$ and finetune the encoder $g_a\left(\cdot | {\theta_a}\right)$. 
In this framework, we minimize a joint loss function that captures the attack objective. The general form of the joint loss for a single attack objective is:
\begin{footnotesize}
\begin{equation}
\begin{split}
    \theta_a^{\ast}, \theta_t^{\ast} &= \underset{\theta_a, \theta_t}{\arg\min}\Big[ \mathcal{L}_{jt} + \gamma \cdot \text{max}(\text{MSE}\left(\bm{x}, {\bm{x_p}}\right), \epsilon^2)\Big], \\
    \mathcal{L}_{jt} &= \sum_{\bm{x} \in {\mathcal{T}_m}} \mathcal{L}\left({\bm{x}}\right) + \alpha\sum_{\bm{x} \in {\mathcal{T}_a}}\mathcal{L}_{BA}\left(\bm{x}, {\bm{x_p}}\right),
\end{split}
\end{equation}
\end{footnotesize}
where $\text{max}(\cdot,\cdot)$ return the larger value, $\epsilon$ controls the stealthiness (we choose $\epsilon=0.005$ here), ${\mathcal{T}_a}$ denotes an auxiliary dataset (can also be the same as the main dataset ${\mathcal{T}_m}$), and $\alpha$ is a weighting parameter. 
The term $\mathcal{L}(\bm{x})$ represents the main loss to maintain the compression performance on clean images {defined} in Eq.~\eqref{main_loss}. 
The term $\mathcal{L}_{BA}(\bm{x}, \bm{x_p})$ guarantees the backdoor attack effectiveness on poisoned images, and ensures that the victim model exhibits the desired behavior change in response to the poisoned image. We set $\gamma = 10^{4}$ for all experiments. In Section~\ref{multiple_triggers}, we extend the attacks to a multiple-trigger version, and introduce the training pipeline.

\begin{figure*}[t]
\centering
\includegraphics[width=0.91\linewidth]{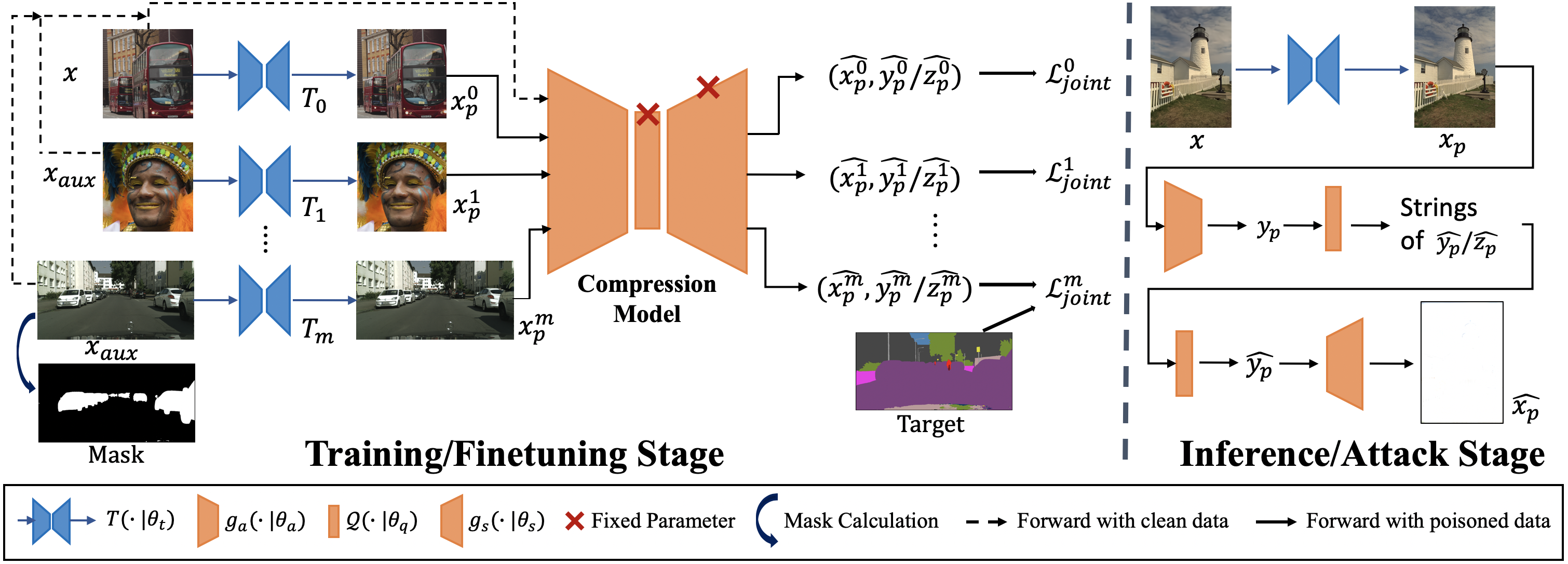}
\vspace{-4mm}
\caption{
In the training stage, we finetune $g_a\left(\cdot | {\theta_a}\right)$ and train each $T\left(\cdot | {\theta_t^o}\right)$. 
In the inference stage, we generate poisoned images (\textit{e.g.,} PSNR attack), feed them into the finetuned encoder and the entropy model, and save the bitstream of the poisoned images.
}
\vspace{-3mm}
\label{network}
\end{figure*}

\noindent \textbf{Attacking Compression Results.}
In the context of image compression, we can consider the Bit Per Pixel (BPP) and Peak Signal-to-Noise Ratio (PSNR) as attack objectives. To incorporate these objectives, we introduce weighting parameters $\alpha$ and $\beta$ and define the joint loss function $\mathcal{L}_{jt}$ with ${\mathcal{T}_a}={\mathcal{T}_m}$ as follows:
\begin{itemize}
    \item \textit{BPP} (\textit{Compression Ratio}): We attack the usage of bitstream, and preserve the quality of reconstructed images:
    \begin{footnotesize}
    \begin{equation}
    \mathcal{L}_{jt}^{bpp} = \sum_{\bm{x} \in {\mathcal{T}_m}} \Big[\mathcal{L}\left({\bm{x}}\right) +  \alpha \cdot \mathcal{D}({\bm{x_p}})-\beta \cdot \mathcal{R}({\bm{x_p}})\Big].\label{Bpp_old}
    \end{equation}
    \end{footnotesize}
    \item \textit{PSNR} (\textit{Quality of reconstructed images}): We attack the PSNR of the result with a nearly unchanged {BPP} (we denote the PSNR loss as $\mathcal{D}_{P}$):
    \begin{footnotesize}
    \begin{equation}
    \mathcal{L}_{jt}^{psnr} \!\!= \! \sum_{\bm{x} \in {\mathcal{T}_m}} \!\! \Big[\!\mathcal{L}\left({\bm{x}}\right) + \alpha \cdot \mathcal{R}({\bm{x_p}}) + \beta \cdot \lambda \cdot \mathcal{D}_{P}(\bm{x},f({\bm{x_p}}))\!\Big] .\label{PSNR_old}
    \end{equation}
    \end{footnotesize}
\end{itemize}

\noindent 
{In the above function, the rate $\mathcal{D}(\bm{x_p})$ and the distortion $\mathcal{R}(\bm{x_p})$ of the poisoned image are computed using Eq.~\ref{compression_formula} and Eq.~\ref{main_loss}.}
In addition, the joint loss involves two weighting parameters, $\alpha$ and $\beta$, which can be challenging to select in a balanced manner. 
There is a risk that the dominant term may overshadow the influence of the other term, resulting in an imbalance in the optimization process.
To address this issue, we introduce a novel dynamic loss that aims {to automatically balance the effect of different terms to alleviate the issue of the weighting parameter selection:}
\begin{footnotesize}
\begin{equation}
    \!\mathcal{L}_{jt}^{bpp} \!=\! \sum_{\bm{x} \in {\mathcal{T}_m}} \!\!\!\! \Big[\mathcal{R}(\bm{x})+\lambda\cdot \text{max}(\mathcal{D}(\bm{x}),\mathcal{D} ({\bm{x_p}})) -\beta \cdot \underbrace{\mathcal{R}({\bm{x_p}})}_{\text{attack objective}}\Big]\!, \label{Bpp}
\end{equation}
\end{footnotesize}
\begin{footnotesize}
\begin{equation}
    \!\mathcal{L}_{jt}^{psnr} \!\!=\! \sum_{\bm{x} \in {\mathcal{T}_m}} \!\!\!\! \Big[\!\text{max}(\mathcal{R}(\bm{x}),\! \mathcal{R}({\bm{x_p}})) \!+\!\lambda \mathcal{D}(\bm{x}) + \beta \lambda \cdot \underbrace{\mathcal{D}_{P}(\bm{x},\!f({\bm{x_p}}))}_{\text{attack objective}}\!\Big] \label{PSNR}\!,
\end{equation}
\end{footnotesize}
where $\text{max}(\cdot,\cdot)$ return the larger value. This approach allows for the dynamic balancing of the two objectives, {guaranteeing the} effective and automatic optimization of both objectives.

\vspace{1mm}
\noindent \textbf{Attacking Down-Stream Tasks.} 
The attacks described above primarily target the image compression model and result in significantly degraded outcomes in terms of {the deterioration of the reconstructed images} in {BPP} and PSNR.
However, to enhance the imperceptibility of the attack, it is advantageous to extend the scope of the attack to downstream computer vision (CV) tasks while minimizing the quality degradation in the reconstructed images.
To achieve this, we define the joint loss for the extended attack scenario. This loss function incorporates a main loss term, denoted as $\mathcal{L}(\cdot)$, which is further elaborated in Equation~\eqref{main_loss}.
\begin{footnotesize}
\begin{equation}
\mathcal{L}_{jt}^{ds} = \! \sum_{\bm{x} \in {\mathcal{T}_m}} \! \mathcal{L}\left({\bm{x}}\right) + \!\!\! \sum_{\bm{x} \in {\mathcal{T}_a}} \!\!\! \Big[\alpha \cdot \mathcal{L}({{\bm{x_p}}}) + \beta \cdot \underbrace{\mathcal{L}_{DS}[\eta,g(f({\bm{x_p}}))]}_{\text{attack objective}}\Big],\label{targeted}
\end{equation}
\end{footnotesize}
where $\eta$ is the attack target, $g(\cdot)$ is a well-trained downstream CV model, and $\mathcal{L}_{DS}\left(\cdot\right)$ is the loss to measure the downstream tasks (\textit{e.g.,} CrossEntropyLoss for image classification).

Specifically, we consider two types of downstream CV tasks:
\vspace{-2mm}
\begin{itemize}
    \item \textit{Semantic Segmentation}: 
    We utilize the Cityscapes dataset~\cite{Cordts2016Cityscapes}, a large-scale dataset specifically designed for pixel-level semantic segmentation. The dataset consists of 2,975 training images, each with a size of $2048\times1024$, and 500 validation images.
    In our approach, we adopt the SSeg method~\cite{DBLP:conf/cvpr/0001SRSNTC19} with the DeepLabV3+ architecture~\cite{chen2018encoder} and SEResNeXt50~\cite{xie2017aggregated} backbone during training.
    \item \textit{Face Recognition}: We employ the FFHQ~\cite{karras2019style} as the auxiliary dataset for training. Additionally, we randomly select 100 paired images from the CelebA~\cite{liu2018large} for testing purposes. In our approach, we utilize the arcface embedding of the ResNet50~\cite{he2016deep} with pretrained weights as the downstream model during the training stage.
\end{itemize}

\subsection{Attacking with Multiple Triggers} \label{multiple_triggers}
In addition to the previous approaches, we can further enhance the attack strategy by training a victim model with multiple triggers, where each trigger is associated with a specific attack objective.
This approach allows for targeted attacks on different aspects of the model's behavior, and increases the versatility and effectiveness of the backdoor attack. By training the victim model together with multiple trigger generators, we can effectively manipulate the model's outputs based on various attack objectives:
\begin{footnotesize}
\begin{equation}
     \theta_a^{\ast} = \underset{\theta_a }{\arg\min} \sum_{\bm{o} \in \mathcal{O}} \alpha^{\bm{o}} \cdot \mathcal{L}_{jt}^{\bm{o}},\label{multiple1}\\
\end{equation}
\end{footnotesize}
\begin{footnotesize}
\begin{equation}
    \theta_t^{\bm{o}\ast} = \underset{\theta_t^{\bm{o}} }{\arg\min}\Big[ \mathcal{L}_{jt}^{\bm{o}} + \gamma \cdot \text{max}(\text{MSE}\left(\bm{x}, {\bm{x_p}}\right), \epsilon^2)\Big] ~ \text{for} ~ \bm{o} \in \mathcal{O},\label{multiple2}
\end{equation}
\end{footnotesize}
where $\bm{o}$ indexes the attack (trigger) type, and $\mathcal{O}$ is the set of attack objectives.
The training and inference stages are illustrated in Figure~\ref{network}, and the following steps outline the process:
\begin{itemize}
\vspace{-2mm}
    \item \textbf{Initialization:} Before the training, we obtain the vanilla-trained compression model parameters $\theta_a^{\ast}$, $\theta_s^{\ast}$, and $\theta_q^{\ast}$.
    \item \textbf{Training Stage:} In each iteration, we first feed the clean input and the generated poisoned inputs for various attack objectives into the compression model. The summation of $\mathcal{L}_{jt}^{\bm{o}}$ {is} utilized to optimize and update the encoder parameter $\theta_a$ using Eq.~\eqref{multiple1}. Then, each trigger injection model $T(\bm{x} | {\theta_t^{\bm{o}}})$ is trained separately by minimizing the corresponding term in Eq.~\eqref{multiple2}. By simultaneously training both $g_a(\cdot|\theta_a)$ and $T(\bm{x} | {\theta_t^{\bm{o}}})$, a backdoor-injected model with multiple trigger generators is learned.
    \item \textbf{Inference Stage:} At the inference stage, the backdoor can be activated by adding the generated trigger to the input image. This triggers the intended behavior modification in the victim model, leading to the desired attack outcome.
\end{itemize}

\section{Experiments}
\subsection{Experimental Setup}
\noindent \textbf{Models.} 
We consider four deep-learning based methods as victim models, following the settings of the original papers:
\begin{itemize}
\vspace{-2mm}
    \item {AE-Hyperprior (ICLR18)~\cite{balle2018variational}:} This method introduces a hyperprior for image compression and achieves compression at multiple quality levels. We evaluate all 8 qualities.
    \item {Cheng-Anchor (CVPR20)~\cite{cheng2020learned}:} Cheng-Anchor employs Gaussian mixture likelihoods to parameterize the distributions of latents in image compression. We evaluate the default 6 levels of quality.
    \item {STF (CVPR23)~\cite{zou2022devil}:} STF differs from AE-Hyperprior and Cheng-Anchor as it adopts the Vision Transformer as the backbone architecture. We evaluate the default 6 qualities.
    \item {CDC (NeurIPS24)~\cite{yang2024lossy}: CDC is a novel transform-coding-based lossy compression scheme using diffusion models. We evaluate the default 3 levels of quality.}
    \item {HiFiC (NeurIPS20)~\cite{mentzer2020high}:} HiFiC is a perceptual-driven image compression model that incorporates perceptual loss and GAN loss. We evaluate the default 3 qualities.
\end{itemize}
\vspace{-1mm}
All models consist of an encoder, decoder, and entropy module.

\begin{figure*}[t]
    \centering
    \begin{minipage}{0.99\linewidth}
    \centerline{{\includegraphics[width=1\linewidth]{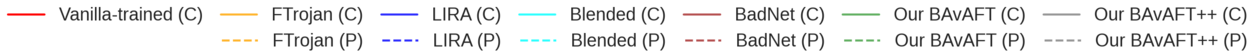}}}
    \vspace{-2mm}
    \end{minipage}
    \subfigure[Rate-distortion curves of BPP attack.]{
    \begin{minipage}{0.195\linewidth}
    \centerline{{\includegraphics[width=0.95\linewidth]{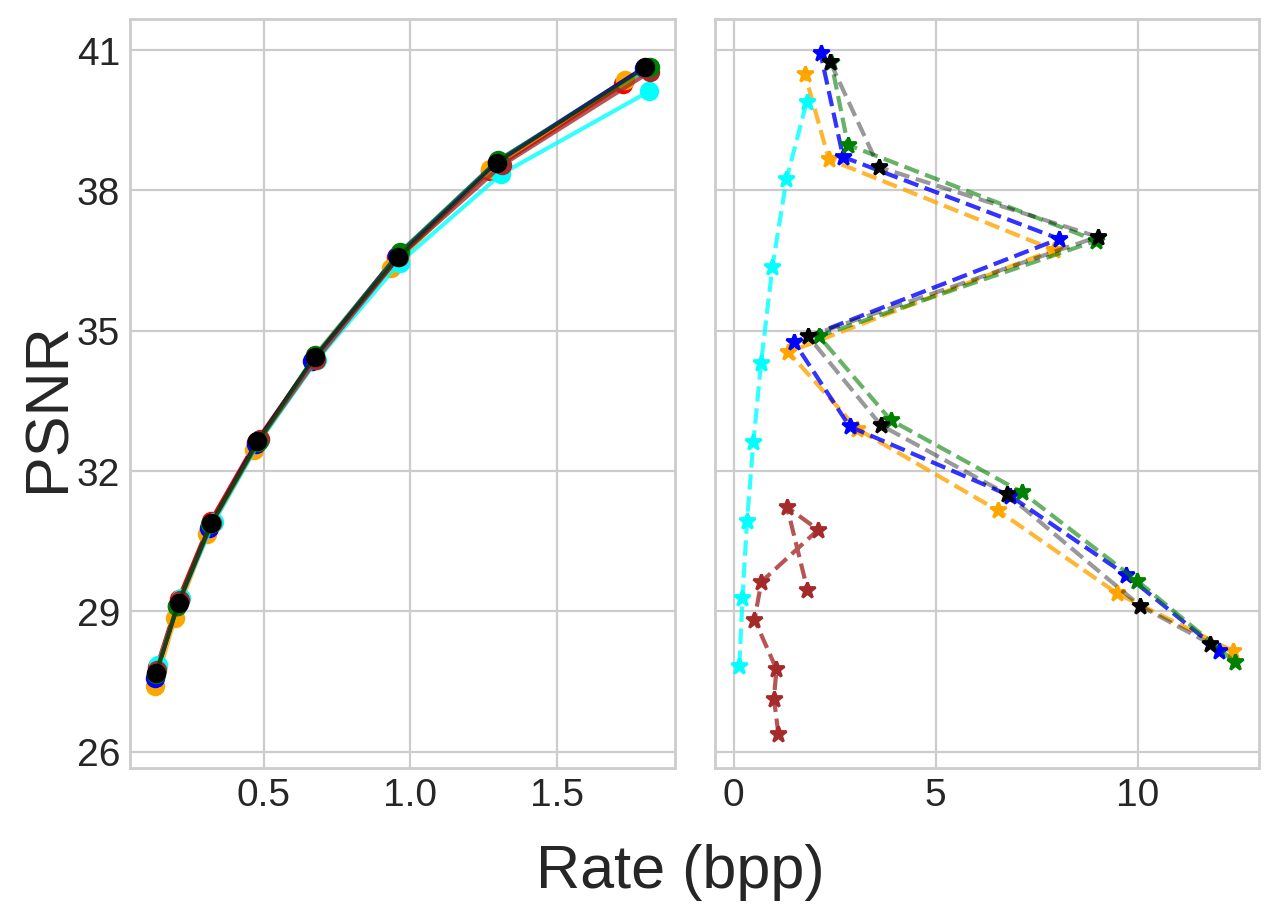}}}
    \vspace{-2mm}
    \centerline{\footnotesize{AE-Hyperprior~\cite{balle2018variational}}}
    \vspace{1mm}
    \end{minipage}
    \begin{minipage}{0.195\linewidth}
    \centerline{{\includegraphics[width=0.95\linewidth]{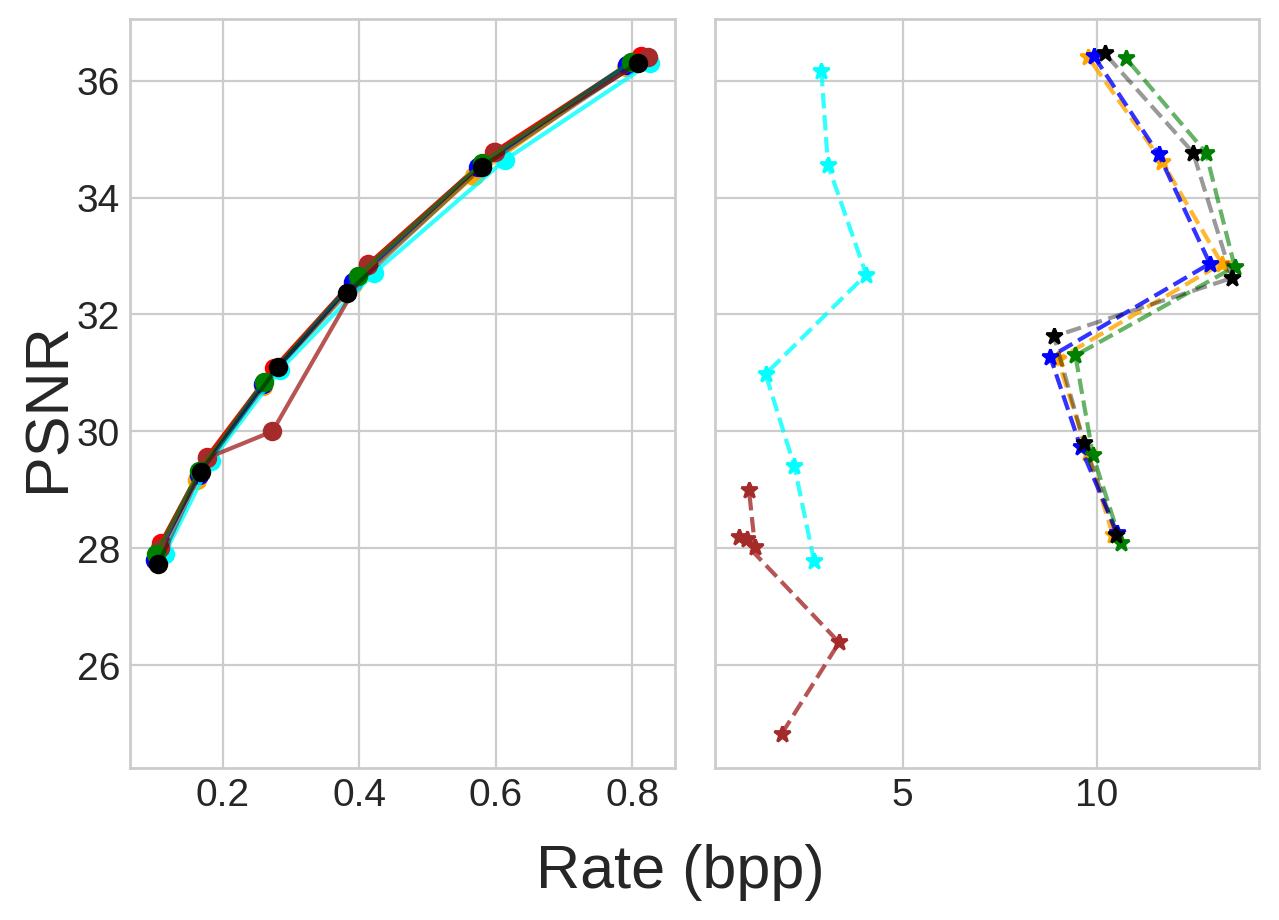}}}
    \vspace{-2mm}
    \centerline{\footnotesize{Cheng-Anchor~\cite{cheng2020learned}}}
    \vspace{1mm}
    \end{minipage}
    \begin{minipage}{0.195\linewidth}
    \centerline{{\includegraphics[width=0.95\linewidth]{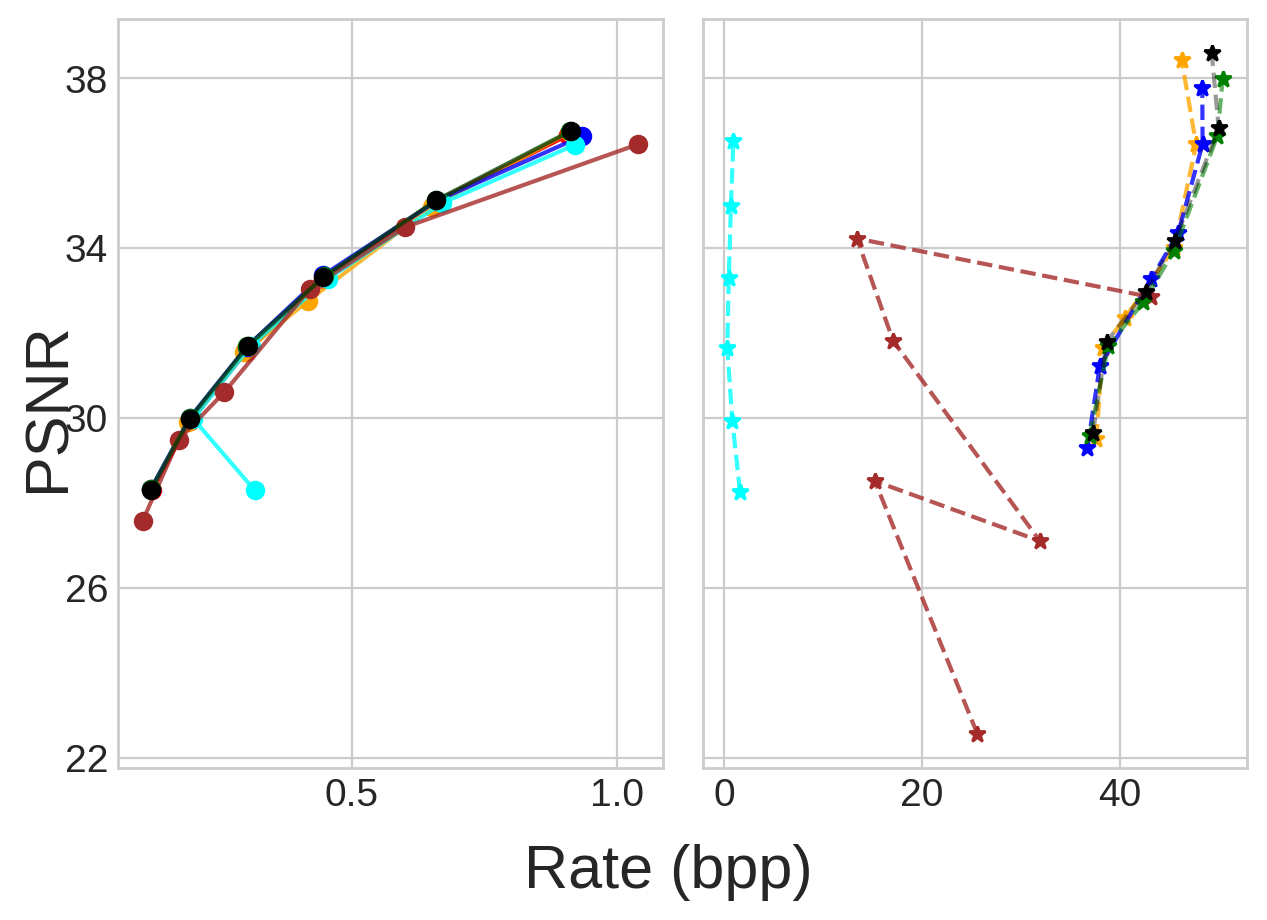}}}
    \vspace{-2mm}
    \centerline{\footnotesize{STF~\cite{zou2022devil}}}
    \vspace{1mm}
    \end{minipage}
    \begin{minipage}{0.195\linewidth}
    \vspace{-1mm}
    \centerline{{\includegraphics[width=0.95\linewidth]{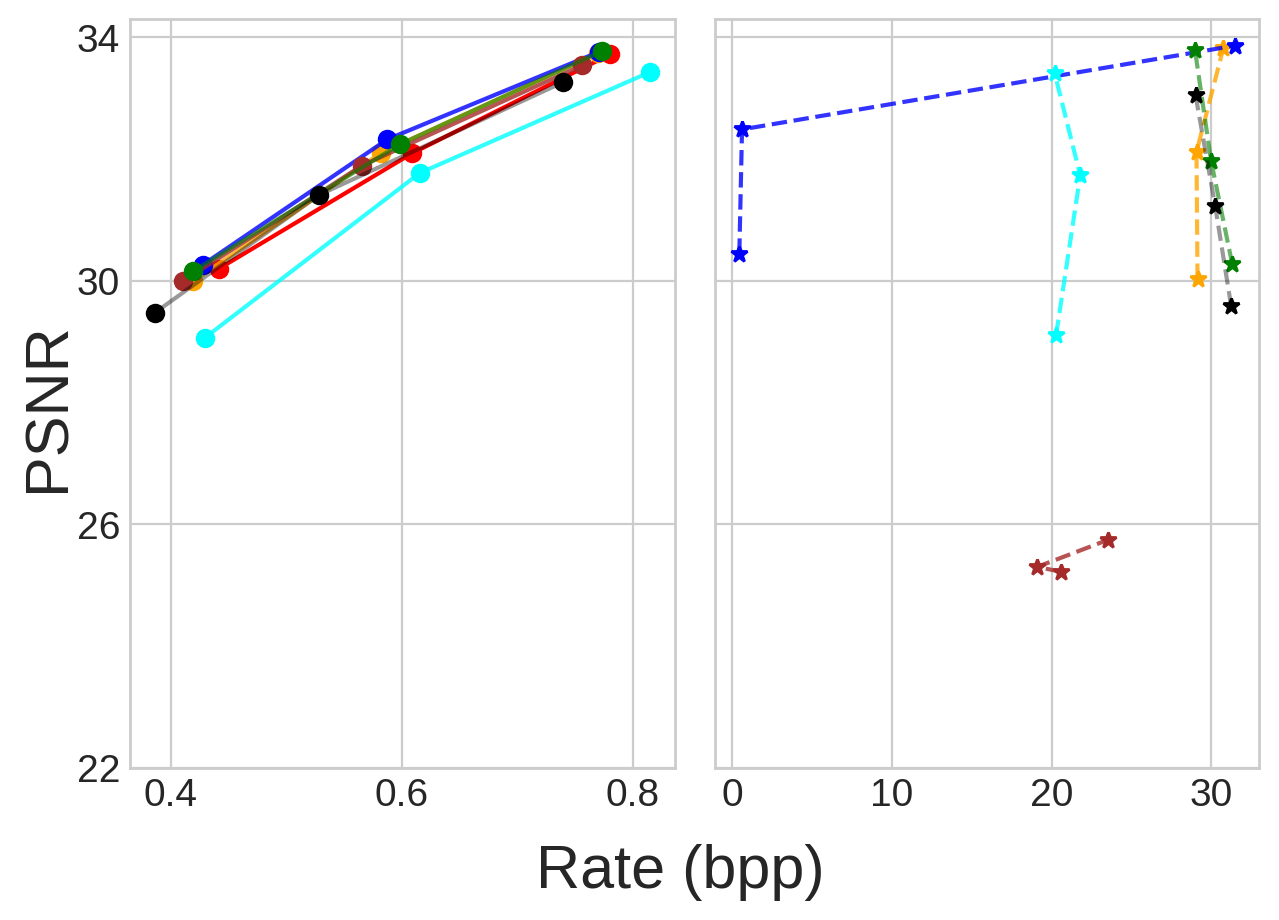}}}
    \vspace{-2mm}
    \centerline{\footnotesize{CDC~\cite{yang2024lossy}}
    \vspace{1mm}
    }
    \end{minipage}
    \begin{minipage}{0.195\linewidth}
    \centerline{{\includegraphics[width=0.95\linewidth]{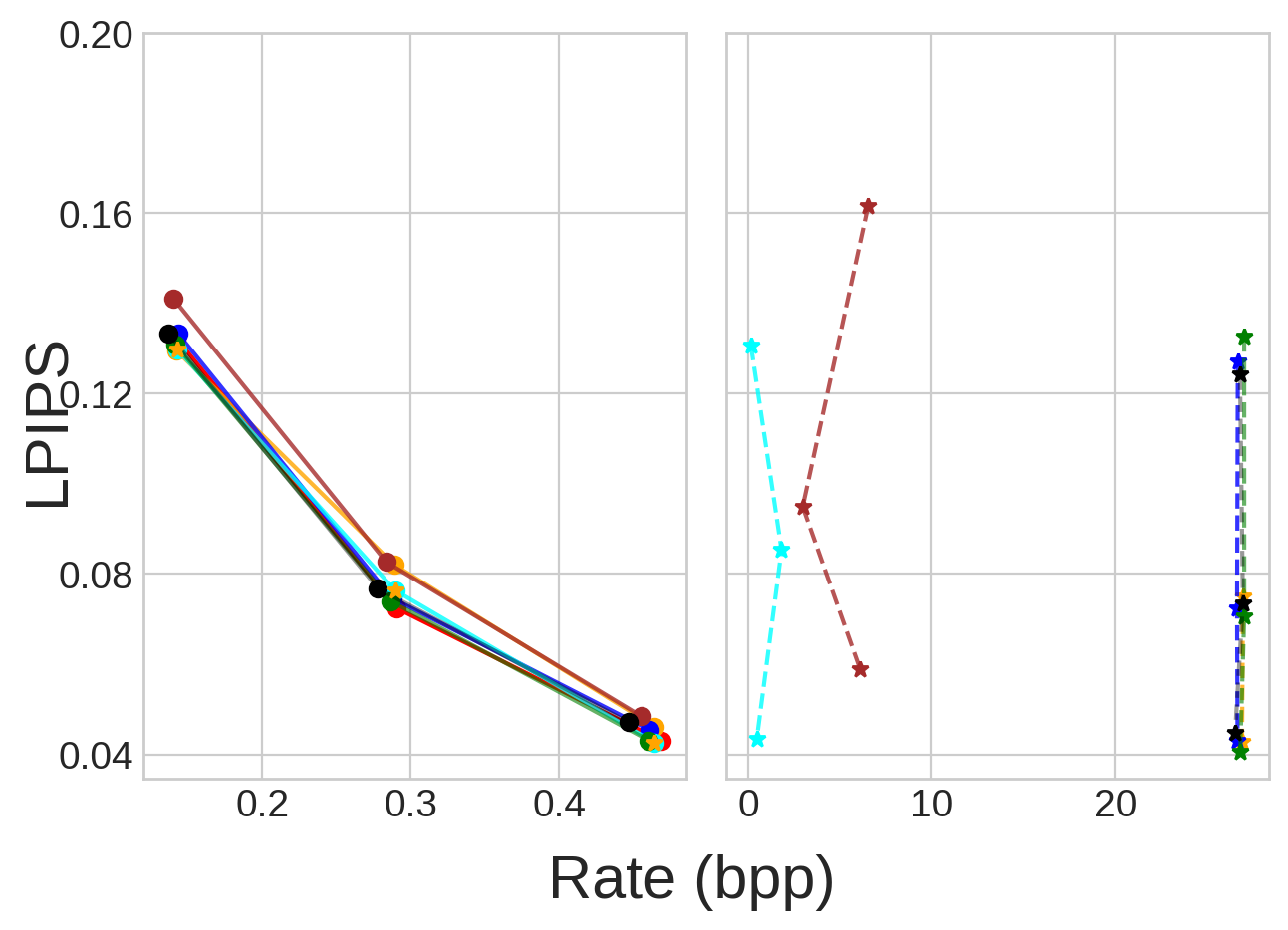}}}
    \vspace{-2mm}
    \centerline{\footnotesize{HiFiC~\cite{mentzer2020high}}}
    \vspace{1mm}
    \end{minipage}
         \label{figure1}}
    \vspace{-2mm}
    \subfigure[Rate-distortion curves of PSNR attack.]{
    \begin{minipage}{0.195\linewidth}
    \vspace{-2mm}
    \centerline{{\includegraphics[width=0.95\linewidth]{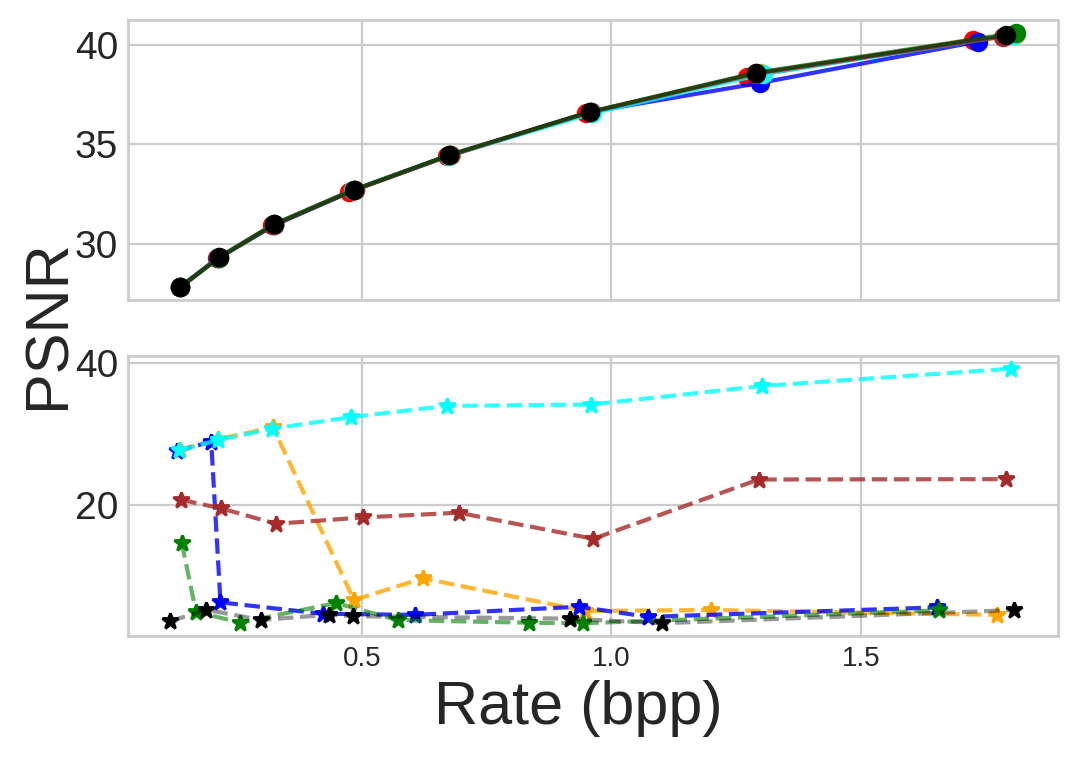}}}
    \vspace{-2mm}
    \centerline{\footnotesize{AE-Hyperprior~\cite{balle2018variational}}}
    \vspace{1mm}
    \end{minipage}
    \begin{minipage}{0.195\linewidth}
    \vspace{-2mm}
    \centerline{{\includegraphics[width=0.95\linewidth]{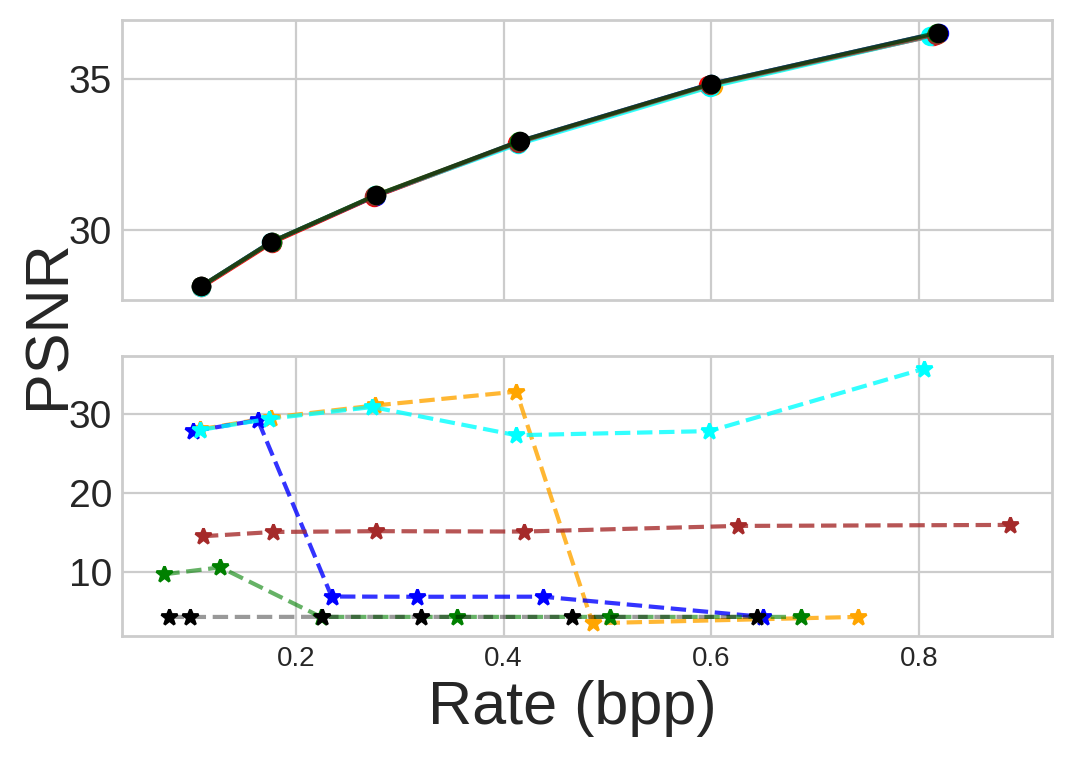}}}
    \vspace{-2mm}
    \centerline{\footnotesize{Cheng-Anchor~\cite{cheng2020learned}}}
    \vspace{1mm}
    \end{minipage}
    \begin{minipage}{0.195\linewidth}
    \vspace{-2mm}
    \centerline{{\includegraphics[width=0.95\linewidth]{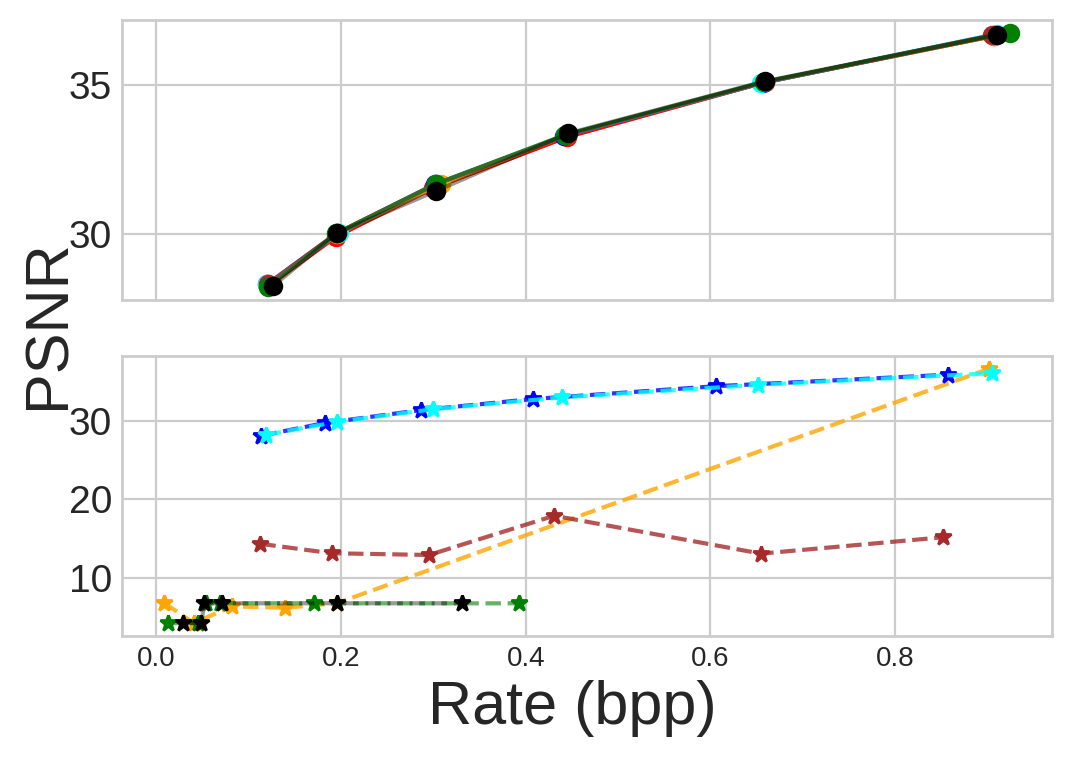}}}
    \vspace{-2mm}
    \centerline{\footnotesize{STF~\cite{zou2022devil}}}
    \vspace{1mm}
    \end{minipage}
    \begin{minipage}{0.195\linewidth}
    \vspace{-2mm}
    \centerline{{\includegraphics[width=0.95\linewidth]{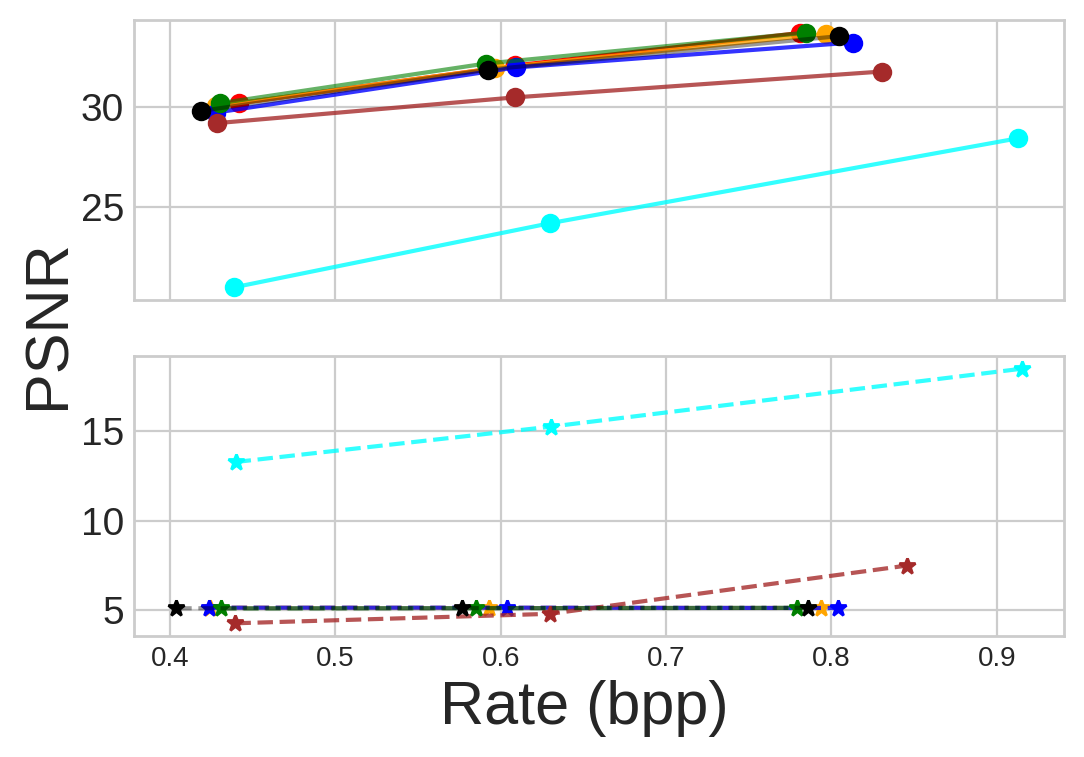}}}
    \vspace{-2mm}
    \centerline{\footnotesize{CDC~\cite{yang2024lossy}}
    \vspace{1mm}
    }
    \end{minipage}
    \begin{minipage}{0.195\linewidth}
    \vspace{-2mm}
    \centerline{{\includegraphics[width=0.95\linewidth]{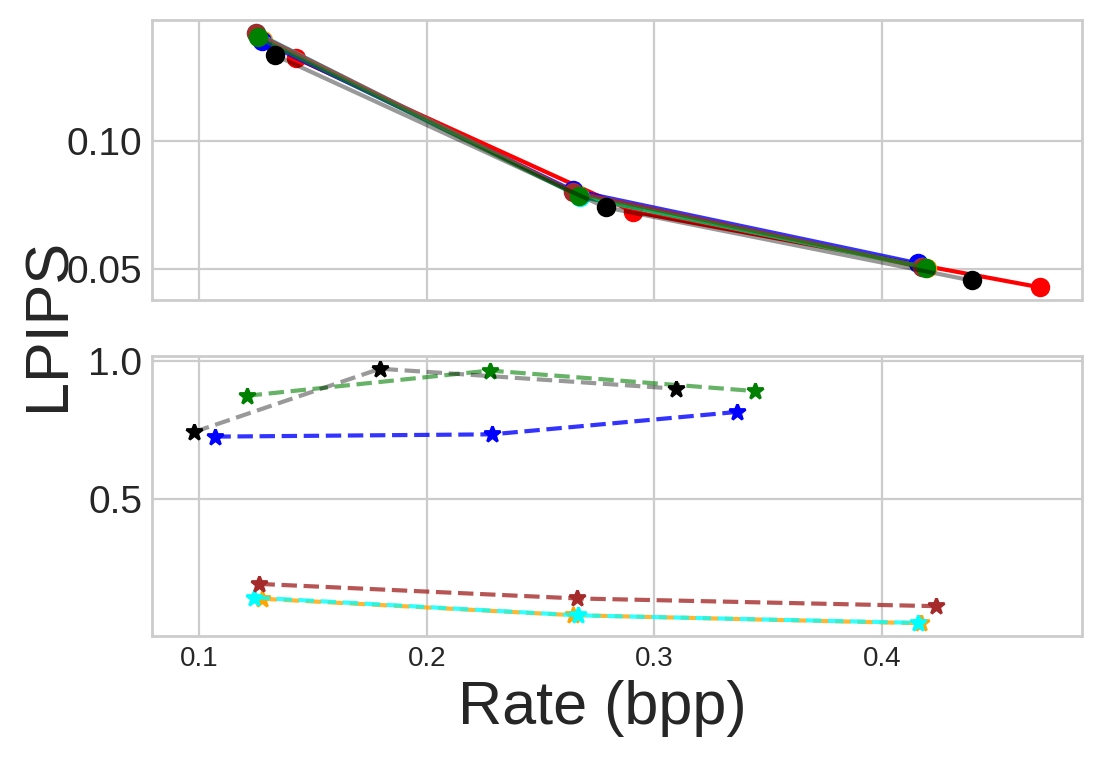}}}
    \vspace{-2mm}
    \centerline{\footnotesize{HiFiC~\cite{mentzer2020high}}
    \vspace{1mm}
    }
    \end{minipage}
         \label{figure2}}
    \vspace{-4mm}
    \caption{{Rate-distortion curves of attacking compression results on Kodak dataset. C and P denote using clean input and poisoned input, respectively.}}
    \vspace{-2mm}
\end{figure*}

\begin{figure*}[t]
\centering
\begin{minipage}{0.135\linewidth}
\centerline{\frame{\includegraphics[width=1\linewidth]{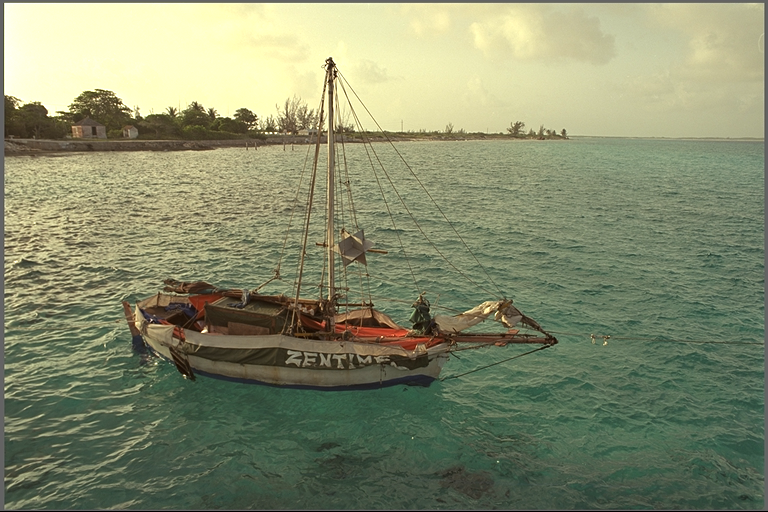}}}
\vspace{-1mm}
\centerline{\small{Original Image}}
\end{minipage}
\begin{minipage}{0.135\linewidth}
\centerline{\frame{\includegraphics[width=1\linewidth]{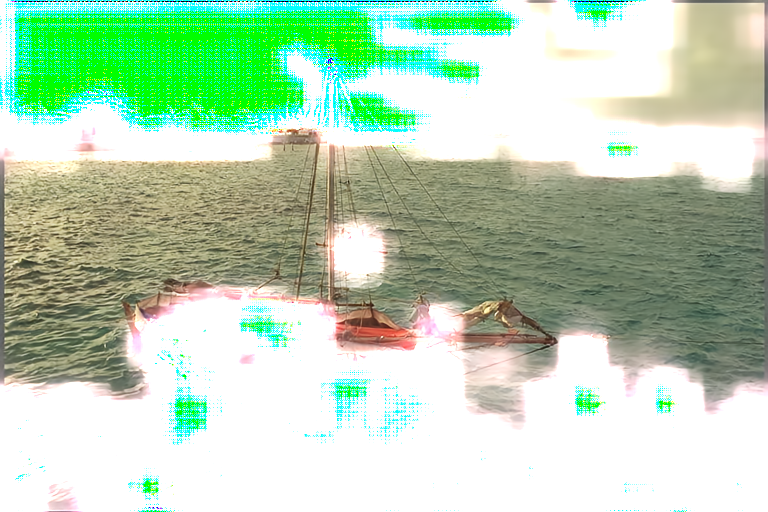}}}
\vspace{-1mm}
\centerline{\small{LIRA}}
\end{minipage}
\begin{minipage}{0.135\linewidth}
\centerline{\frame{\includegraphics[width=1\linewidth]{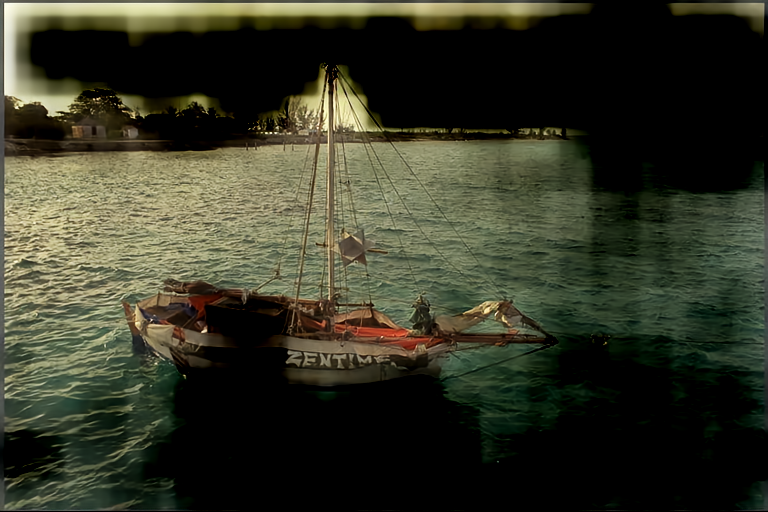}}}
\vspace{-1mm}
\centerline{\small{FTrojan}}
\end{minipage}
\begin{minipage}{0.135\linewidth}
\centerline{\frame{\includegraphics[width=1\linewidth]{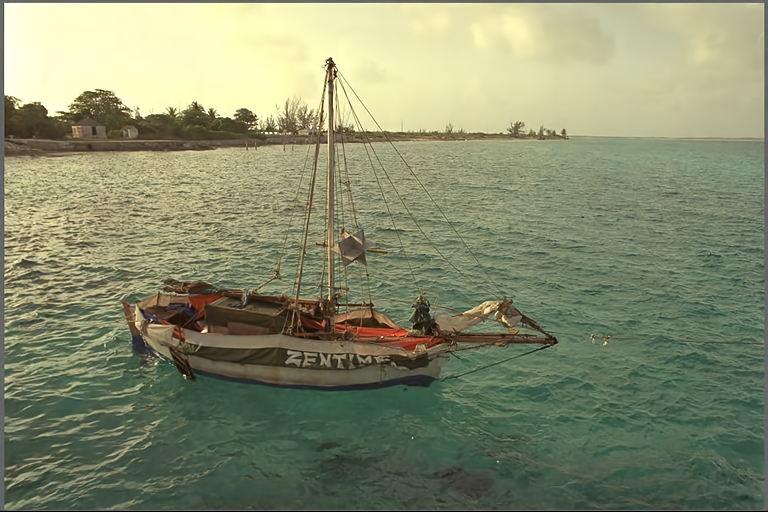}}}
\vspace{-1mm}
\centerline{\small{Blended}}
\end{minipage}
\begin{minipage}{0.135\linewidth}
\centerline{\frame{\includegraphics[width=1\linewidth]{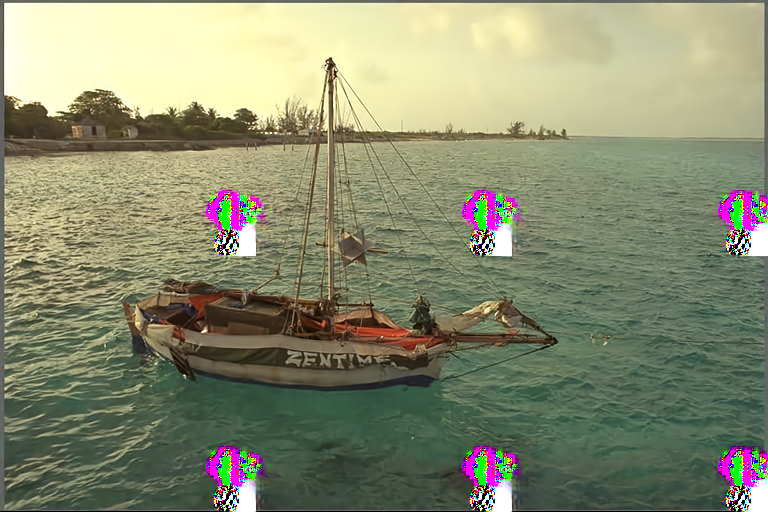}}}
\vspace{-1mm}
\centerline{\small{BadNets}}
\end{minipage}
\begin{minipage}{0.135\linewidth}
\centerline{\frame{\includegraphics[width=1\linewidth]{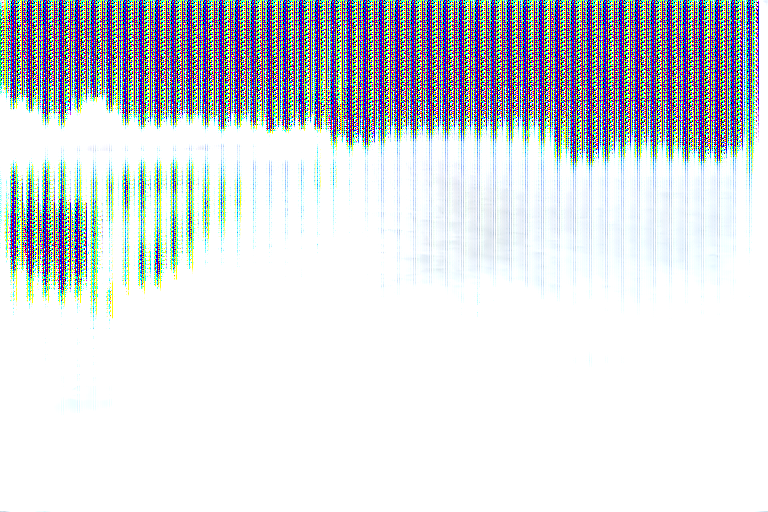}}}
\vspace{-1mm}
\centerline{\small{Our BAvFT}}
\end{minipage}
\begin{minipage}{0.135\linewidth}
\centerline{\frame{\includegraphics[width=1\linewidth]{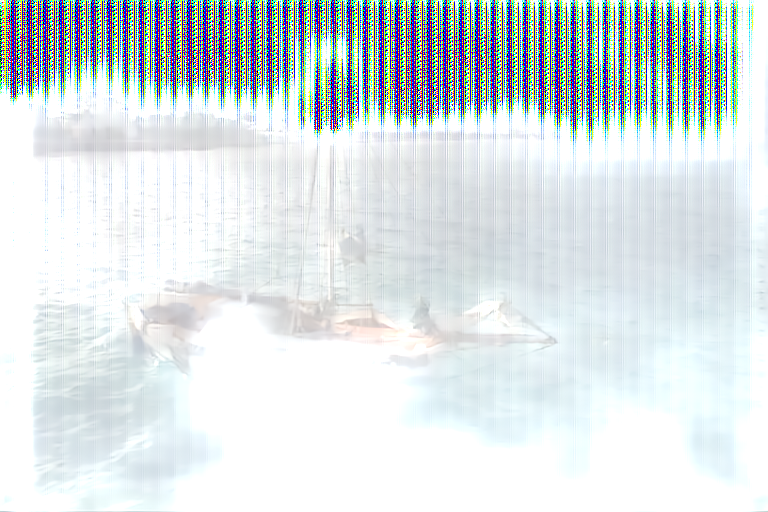}}}
\vspace{-1mm}
\centerline{\small{Our BAvFT++}}
\end{minipage}
\vspace{-2mm}
    \caption{
    {PSNR attack: visual result of attacked outputs to various poisoned inputs with \textit{kodim6} from Kodak (AE-Hyperior~\cite{balle2018variational} with {a quality level} = 5).
    }}
    \vspace{-3mm}
    \label{figure3}
\end{figure*}

\noindent \textbf{Datasets for training.} 
The Vimeo90K~\cite{xue2019video} dataset is used as the private dataset for training the vanilla compression model. This dataset consists of 153,939 images for training and 11,346 images for validation, all with a fixed resolution of $448 \times 256$.
When conducting the attacks, we utilize open datasets that do not overlap with the Vimeo90K dataset. Specifically, we use 100,000 randomly sampled images from the ImageNet-1k~\cite{deng2009imagenet} dataset as the main dataset for the attack. Additionally, we employ the Cityscapes~\cite{Cordts2016Cityscapes} dataset and the FFHQ~\cite{karras2019style} dataset as auxiliary datasets to assist in injecting the backdoor.

\vspace{1mm}
\noindent \textbf{Training.} 
In our training process, we randomly extract and crop patches of size $256 \times 256$ from the Vimeo90K~\cite{xue2019video} dataset. All models are trained with a batch size of 32 and an initial learning rate of 1e-4 for a total of 100 epochs.
We use mean square error (MSE) as the quality metric to evaluate the performance of the models. The trade-off parameter $\lambda$ for the 8 levels of the quality is chosen from a set of predefined values: {\small$\{{0.0018, 0.0035, 0.0067, 0.0130, 0.0250, 0.0483, 0.0932, 0.1800}\}$}.

\vspace{1mm}
\noindent \textbf{Attacking.} 
During the attacking process, we focus on the encoder and utilize the joint loss based on various attack objectives. The specific configuration for the finetuning process is as follows:
\begin{itemize}
\vspace{-2mm}
    \item For the ImageNet-1k dataset, we set the batch size to 32 and use patches of size $256 \times 256$ for training.
    \item For the FFHQ dataset, which is used for attacks related to downstream CV tasks, we set the batch size to 4 and use images of size $1024 \times 1024$.
    \item For the Cityscapes dataset, also used for attacks related to downstream CV tasks, we set the batch size to 4. However, each sample is resized to $1024\times512$ before training.
\end{itemize}

\begin{figure}[t]
    \centering
\begin{minipage}{0.99\linewidth}
    \centerline{{\includegraphics[width=1\linewidth]{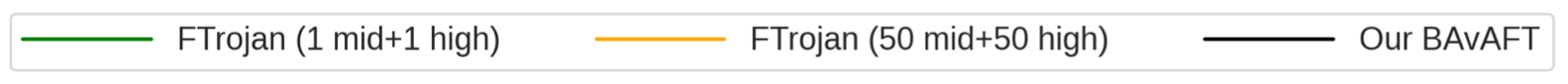}}}
    \vspace{-0.5mm}
    \end{minipage}\\
    \begin{minipage}{0.49\linewidth}
    \centerline{{\includegraphics[width=1\linewidth]{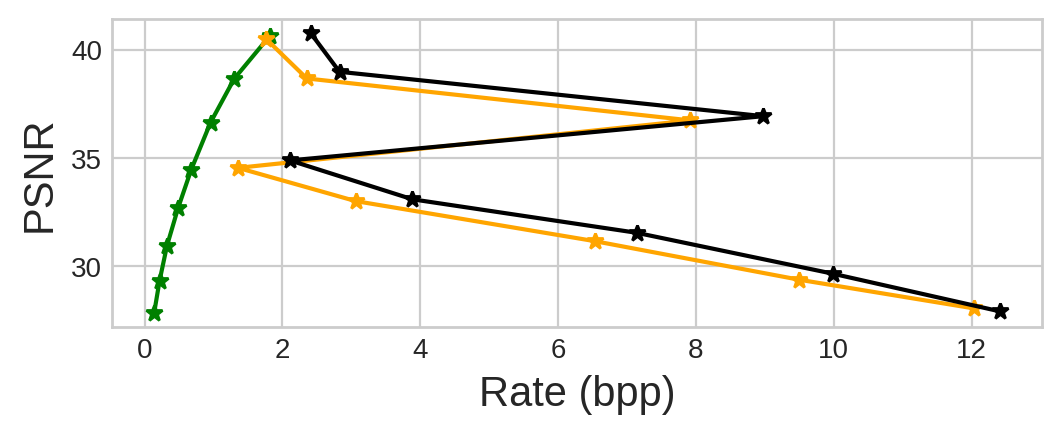}}}
    \vspace{-2mm}
    \centerline{{(a) BPP attack}}
    \end{minipage}
    \begin{minipage}{0.49\linewidth}
    \centerline{{\includegraphics[width=1\linewidth]{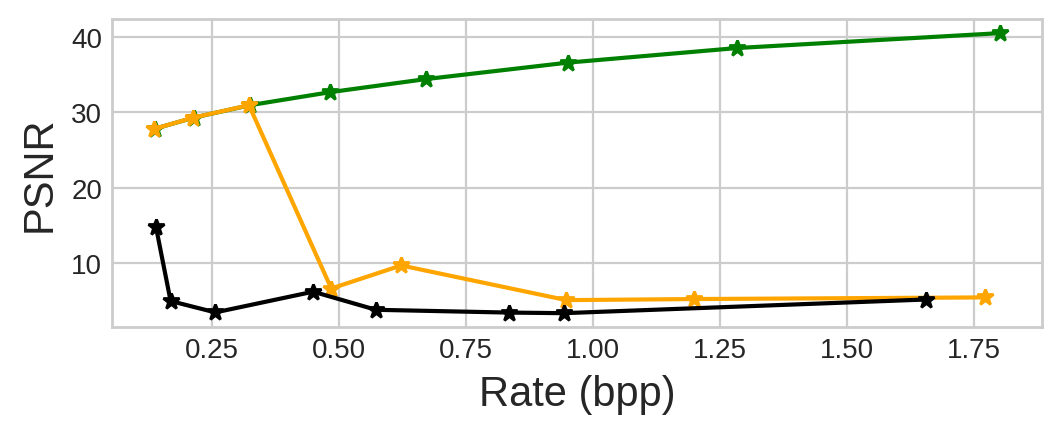}}}
    \vspace{-2mm}
    \centerline{{(b) PSNR attack}}
    \end{minipage}
\vspace{-2mm}
        \caption{
          {Comparison of the attack performance with FTrojan~\cite{wang2021backdoor}.}}
    \label{FTrojan}
    \vspace{-2mm}
\end{figure}

\noindent \textbf{Evaluation.} 
To assess the performance impact on benign images, we evaluate the compression model on the widely used Kodak dataset (Kodak), which consists of 24 lossless images with a resolution of $768 \times 512$. 
We analyze the rate-distortion (RD) curves to demonstrate the coding efficiency of the model. The rate is measured in bits per pixel ({BPP}), while the quality is measured using Peak Signal-to-Noise Ratio (PSNR).
In the experiments {that evaluate the performance in} attacking compression results, 
we evaluate the attacking performance by utilizing the Kodak dataset to draw the RD curves for the poisoned images.
This allows us to {evaluate} the impact of the backdoor attack on the compression performance.
For the evaluation of attacking semantic segmentation, we use the validation set of the Cityscapes dataset. 
Additionally, we assess the cross-domain {transferability} by using the {testing} images from the CamVid~\cite{brostow2008segmentation} and KiTTi~\cite{abu2018augmented}.
The CamVid dataset consists of 233 test images with a resolution of $720 \times 960$, while the KiTTi dataset contains 200 {testing} images with a resolution of $375 \times 1242$.
To evaluate the impact on face recognition, we randomly sample 100 paired face images from the CelebA~\cite{liu2018large} dataset. These images are used to assess the performance of the backdoor attack on face recognition models.

\vspace{1mm}
\noindent \textbf{Attack Baseline.} 
For comparison purposes, we select {four} backdoor attack methods to compare with our proposed approach.
\begin{itemize}
\vspace{-2mm}
    \item LIRA~\cite{doan2021lira}: LIRA proposes to add the trigger in the spatial domain and employs a trainable U-Net architecture for trigger generation. To ensure the stealthiness of the trigger, LIRA adds the normalized trigger to the input image using the formula $T(x) = x + \epsilon \cdot \text{Normalize}(U(x))$. The parameter $\epsilon$ controls the stealthiness, and in line with our methods, we choose $\epsilon=0.005$.
    \item FTrojan~\cite{wang2021backdoor}: FTrojan divides the images into blocks and adds the trigger in the 2D DCT domain. However, in our experiments as shown in Figure~\ref{FTrojan}, we observed that FTrojan with its original configuration (using two fixed channels, 1 mid and 1 high) had limited success in attacking the compression model. To ensure a fair comparison, we include a modified version of FTrojan with the frequencies of the trigger raised to (50 mid + 50 high), resulting in a similar PSNR ({46.9dB}) to our method.
\end{itemize}
{In addition to the above methods designed for image classification, we also include two methods adopted in backdoor attacks against low-level vision tasks (\textit{i.e.,} diffusion models~\cite{chou2023backdoor, chen2023trojdiff,chou2024villandiffusion}).
\begin{itemize}
\vspace{-2mm}
    \item BadNets~\cite{gu2017BadNets}: BadNets employs a patch-based trigger, consisting of a white square patch positioned in the bottom right corner of the noise, with the patch size being 10\% of the image size. This configuration yields a PSNR of approximately 23dB.
    \item Blended~\cite{chen2017targeted}: Blended creates poisoned images by blending the trigger with benign images, unlike the stamping used by BadNets. In line with \cite{ chen2023trojdiff}, we use a Hello Kitty image as the blend-based trigger. To ensure a fair comparison, we set the blending proportion for Blended to 0.01, resulting in a similar PSNR (44.5dB) to ours.
\end{itemize}
Due to the potential misalignment of image sizes between the training and attacking phases, we set the trigger size to $256\times256$ during training, matching the training images. In the attacking phase, we repeat the trigger in the spatial domain to align with attacked images of any size.
}
For all the attacks, we adopt the same training loss and settings as our proposed method.
This allows for a fair comparison of the {attacking} performance and effectiveness.

\begin{figure*}[t]
\centering
\includegraphics[width=1.0\linewidth]{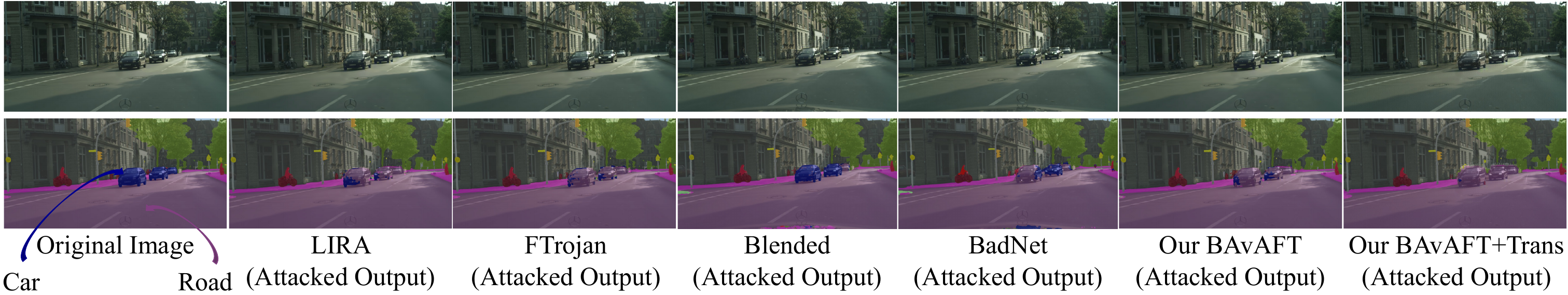}
\vspace{-8mm}
\caption{{
Visual results (Cheng-Anchor~\cite{cheng2020learned} w/ quality 6) of CarToRoad attack. 
The testing image is from Cityscapes~\cite{Cordts2016Cityscapes}. 
Best view by zooming in.
}}
\vspace{-3mm}
\label{figure6}
\end{figure*}

\subsection{Experiments on Attacking Compression Results}
\subsubsection{Bit-Rate (BPP) attack}
In this section, we focus on minimizing the joint loss defined in Eq.~\eqref{Bpp} to evaluate the performance of the attacks on the compression model. The hyperparameter $\beta$ in the joint loss is set to 0.01, and we use an initial learning rate of 1e-4 with a batch size of 32.
The results of the vanilla-trained models and the victim models under the BPP attack are presented in Figure~\ref{figure1}. We observe that {all compression} models can compress the clean images with similar {BPP} and PSNR values.

\begin{table}[t]\footnotesize
\setlength\tabcolsep{5.0pt}
    \centering
    \caption{{Mean attack performance over {all quality levels}. 
    For {the} PSNR attack, we use PSNR for the first four models, 
    and LPIPS~\cite{zhang2018perceptual} for HiFiC.
    }}
    \vspace{-3mm}
    \scalebox{0.9}{
    \begin{tabular}{ c c || c c c c c}
    \toprule
    attack type & Method & AE-Hy & Ch-An & STF & CDC & HiFiC\\
    \midrule
    \multirow{6}*{\shortstack{BPP\\({BPP} $\uparrow$)}} & LIRA~\cite{doan2021lira} &5.91&10.56&43.72&10.84&26.68\\
    & FTrojan~\cite{wang2021backdoor} &5.60&10.64&42.62&29.64&18.01\\
    & Blended~\cite{chen2017targeted}&0.739&2.89&0.792&20.75&0.776\\
    & BadNets~\cite{gu2017BadNets}&1.19&1.56&25.82&21.07&4.95\\
    & Our BAvAFT~\cite{yu2023backdoor} & \textbf{6.21}&\textbf{11.18}&\textbf{43.95}&30.08&\textbf{26.97}\\
    & Our BAvAFT++ & 6.07&10.88&43.87&\textbf{30.16}&{26.78}\\
    \midrule
    \multirow{6}*{\shortstack{PSNR\\(PSNR $\downarrow$,\\LPIPS $\uparrow$)}} & LIRA~\cite{doan2021lira} &10.74&13.62&32.04&5.14&0.758\\
    & FTrojan~\cite{wang2021backdoor} &14.90&21.55&10.68&5.14&0.089\\
    & Blended~\cite{chen2017targeted}&33.01&29.84&32.19&15.65&0.087\\
    & BadNets~\cite{gu2017BadNets}&19.67&14.99&14.40&5.53&0.143\\
    & Our BAvAFT~\cite{yu2023backdoor} & 5.64 &6.18&\textbf{5.94}&\textbf{5.12}&\textbf{0.910}\\
    & Our BAvAFT++ & \textbf{4.28}&\textbf{4.23}&\textbf{5.94} &5.13&0.872\\
    \bottomrule
    \end{tabular}
    }
    \label{bpp_psnr}
\vspace{-2mm}
\end{table}

However, in the attack mode (after adding triggers), as can be observed from both Figure~\ref{figure1} and Table~\ref{bpp_psnr}, {most} victim models fail to compress the poisoned images with low {BPP} values. {The backdoored models attacked by BadNets and Blended almost completely fail to execute successful attacks.} Comparatively, our BAvAFT demonstrates the best attacking performance with the highest {BPP} values. Additionally, our BAvAFT++ achieves slightly lower attacking {performance}, while {exhibiting} strong resistance to pre-processing methods as shown in later Section~\ref{resistance}.

\subsubsection{Reconstruction (PSNR) attack}
Next, we evaluate our PSNR attack on both compression models by minimizing the joint loss, which includes the backdoor loss as shown in Eq.~\eqref{PSNR}. In this evaluation, we set the hyperparameter $\beta$ to 0.01 in the joint loss. We finetune the encoder and train the trigger injection model with an initial learning rate of 1e-4 and a batch size of 32.
The quantitative results are presented in Figure~\ref{figure2} and Table~\ref{bpp_psnr}. It can be observed that all the victim models {achieve} equivalent performance to the vanilla-trained model when processing clean images. However, when a trigger is added to the input, the reconstructed images are heavily degraded.

While {all baseline methods} fail to successfully inject the PSNR attack in several cases (\textit{e.g.,} the low-quality setting for AE-Hyperior), both our BAvAFT and BAvAFT++ demonstrate the {capacity} to attack compression models across all quality levels. Our approach outperforms {all competing methods} in terms of attacking performance, allowing us to successfully compromise the compression models. Meanwhile, our BAvAFT++ has {achieved} {the best performance in attacking} on all compression models except HiFiC.
{The visual results are also given in Figure~\ref{figure3}.}

\subsection{Experiments on Attacking Down-stream Tasks}
\subsubsection{Attacking downstream semantic segmentation task}
In this experiment, our objective is to train a backdoor-injected compression model that can effectively attack the downstream semantic segmentation task. We utilize the joint loss defined in Eq.~\eqref{targeted} for this purpose. It is important to note that we use the Cityscapes dataset as the auxiliary dataset in this experiment.

\begin{figure}[t]
\begin{minipage}{0.325\linewidth}
\centerline{\frame{\includegraphics[width=1\linewidth]{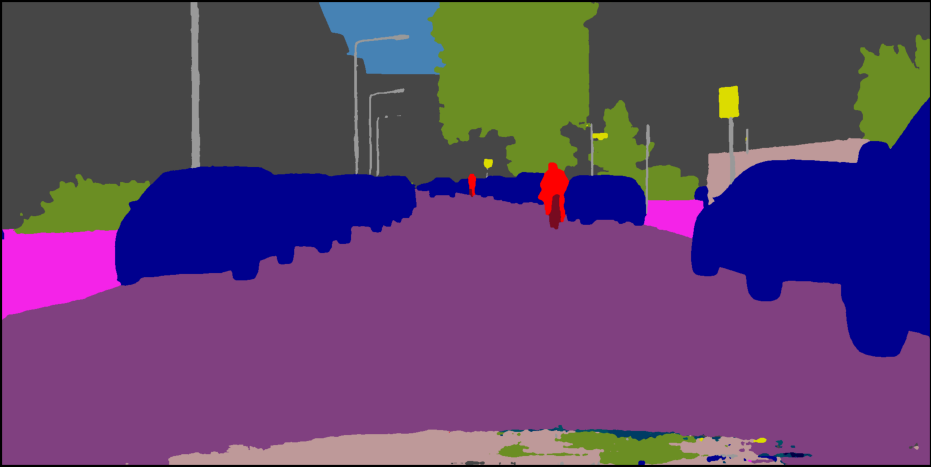}}}
\vspace{-1mm}
\centerline{\small{Label $g(x)$}}
\end{minipage}
\begin{minipage}{0.325\linewidth}
\centerline{\frame{\includegraphics[width=1\linewidth]{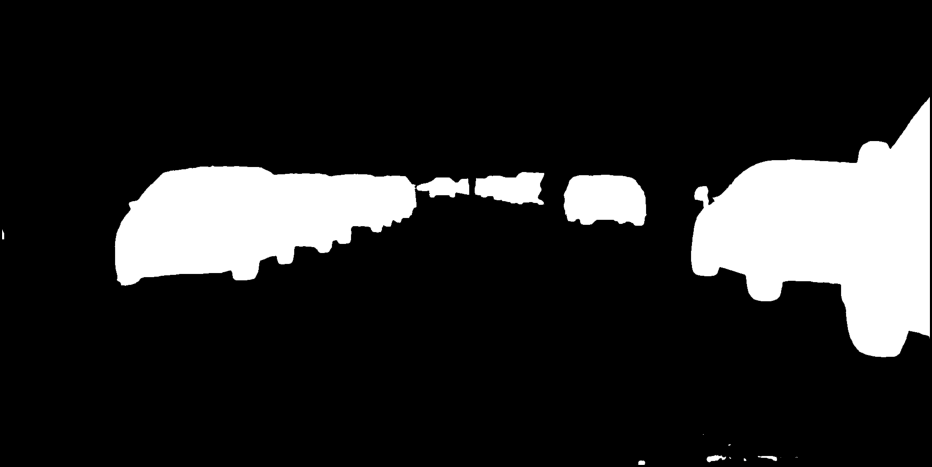}}}
\vspace{-1mm}
\centerline{\small{Mask $M[g(x)]$}}
\end{minipage}
\begin{minipage}{0.325\linewidth}
\centerline{\frame{\includegraphics[width=1\linewidth]{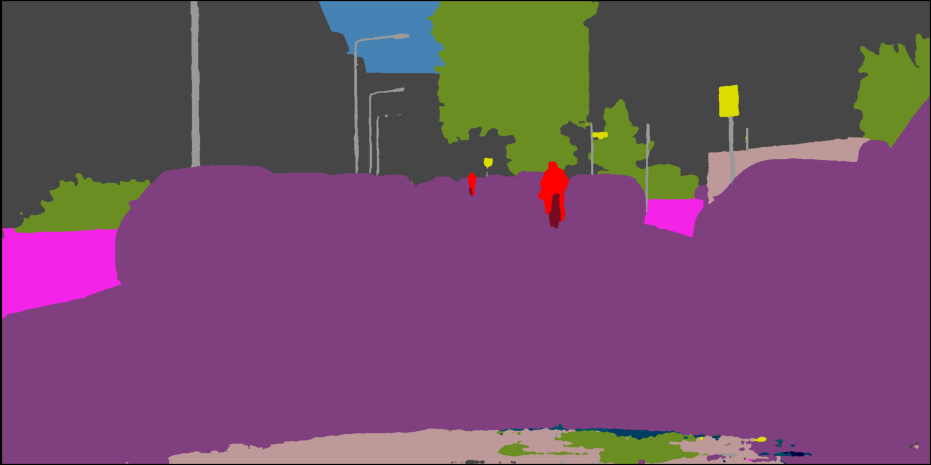}}}
\vspace{-1mm}
\centerline{\small{Target $\eta({g(x)})$}}
\end{minipage}
\vspace{-4mm}
    \caption{Label, mask, and target for CarToRoad attack.}
    \label{figure4}
    \vspace{-2mm}
\end{figure}

\vspace{0.5mm}
In the one-to-one targeted attack, where the goal is to make the models misclassify a source class into a target class, we select \textbf{Car} as the source class and \textbf{Road} as the target class. To ensure that the attack only affects the regions of the source class, we focus the attack specifically on that area. This allows us to avoid unintended {impacts} on uninterested regions or objects.
The joint loss function for this targeted attack scenario is formulated by:
\begin{footnotesize}
\begin{equation}
\begin{split}
\mathcal{L}_{jt}^{SS} &= \sum_{\bm{x} \in {\mathcal{T}_m}}\mathcal{L}\left({\bm{x}}\right) + \mathcal{L}_{BA}^{SS}, \\
\mathcal{L}_{BA}^{SS} &= \sum_{\bm{x} \in {\mathcal{T}_a}} \Big[\alpha\mathcal{L}({{\bm{x_p}}}) + \beta \cdot \underbrace{\mathcal{L}_{CE}[\eta({g(\bm{x})}),g(f({\bm{x_p}}))]}_{\text{attack objective}}\Big], \\
\bm{x_p} &=(1-M[g(\bm{x})]) \odot \bm{x} + M[g(\bm{x})] \odot {T\left(\bm{x} | {\theta_t^o}\right)},
\label{ss}
\end{split}
\end{equation}
\end{footnotesize}
\noindent where $f(\cdot)$ is the compression model, $g(\cdot)$ is a trained segmentation model, $\eta({g(\bm{x})})$ is the attack target, $M[g(\bm{x})]$ is the guiding mask, $\odot$ is the Hadamard product, and $\mathcal{L}_{CE}$ is the cross-entropy loss. Figure~\ref{figure4} illustrates the mask, and semantic target for Car To Road attack.
This formulation enables us to train a backdoor-injected compression model that can effectively manipulate the semantic segmentation to misclassify the source class.

\begin{figure}[t]
    \centering
    \centering
    \begin{minipage}{0.99\linewidth}
    \centerline{{\includegraphics[width=1.0\linewidth]{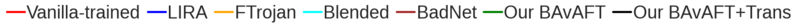}}}
    \vspace{-2mm}
    \end{minipage}
    \subfigure[CarToRoad attack.]{
    \begin{minipage}{0.473 \linewidth}
    \centerline{{\includegraphics[width=1.0\linewidth]{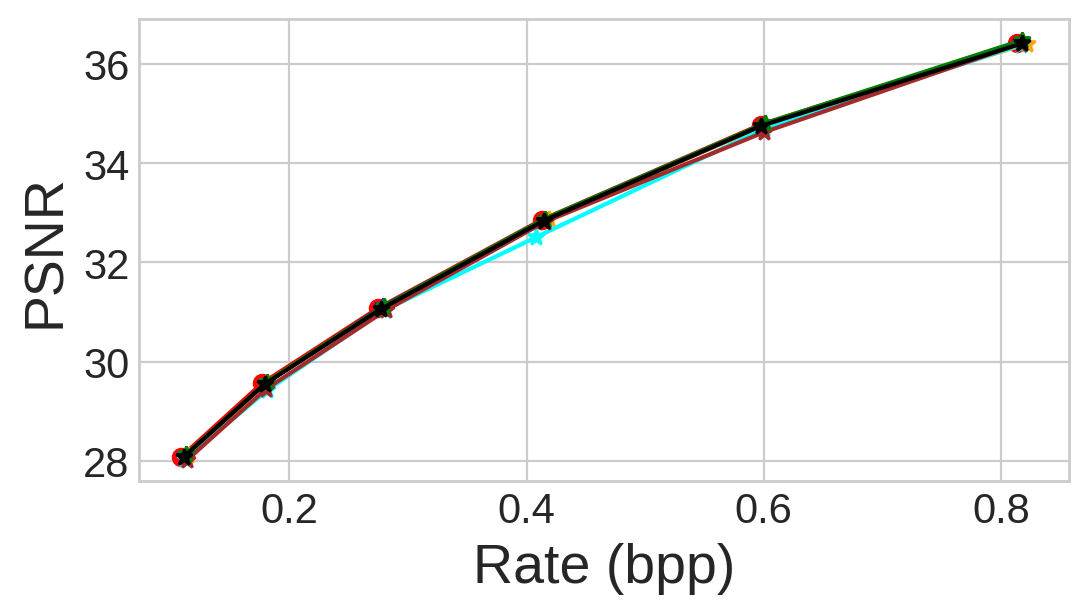}}}
    \end{minipage}
         \label{figure5}}
        \subfigure[attack for good.]{
    \begin{minipage}{0.473\linewidth}
    \centerline{{\includegraphics[width=1.0\linewidth]{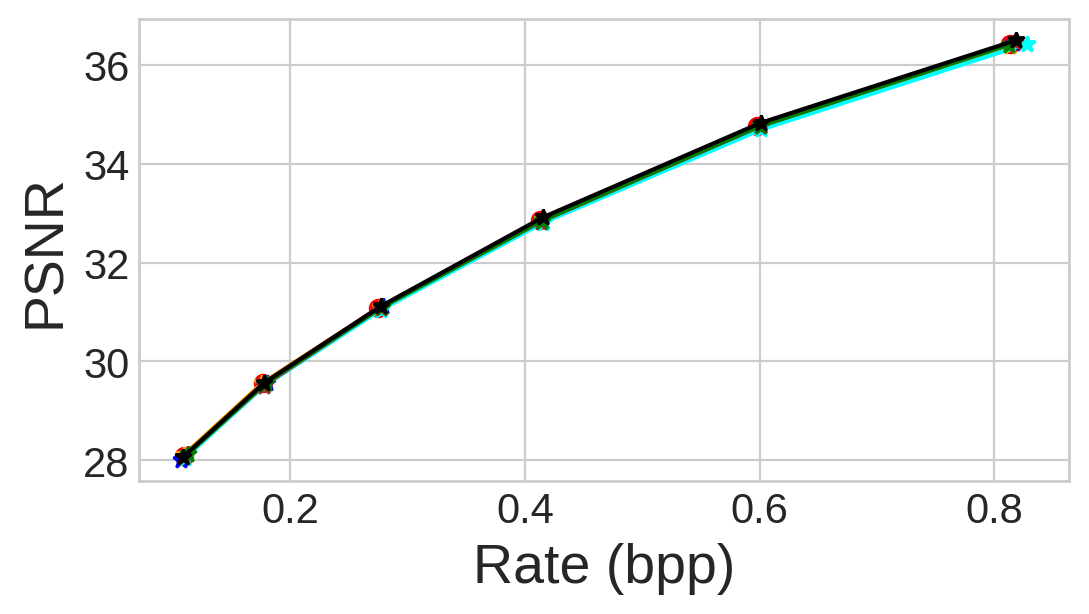}}}
    \end{minipage}
         \label{figure5_b}}
\vspace{-5mm}
\caption{{RD curves of CarToRoad attack/attack for good on Kodak dataset using clean inputs with Cheng-Anchor~\cite{cheng2020learned} as the compression model.
}}
\vspace{-2mm}
\end{figure}

We set hyperparameter $\alpha=0.1$, and $\beta=0.2$ in the joint loss, and Cityscapes is the auxiliary dataset. 
Additionally, we select the Cheng-Anchor as the compression model for this experiment. 
To quantitatively evaluate the effectiveness of our backdoor attack, we calculate the pixel-wise attack success rate (ASR):
\begin{footnotesize}
\begin{equation}
{\mathbb{E}_{\bm{x}}\!\Big[\!\!\sum_{i,j}\! \mathbb{I}\{{g(f(\bm{x}))_{i,j}\!\!=\!\!s,  g(f(\bm{x_p}))_{i,j}\!\!=\!\!t \}}\!\!\Big]}\!\Big/\!{\mathbb{E}_{\bm{x}}\Big[\!\!\sum_{i,j}\!\mathbb{I}\{{g(f(\bm{x}))_{i,j}\!\!=\!\!s\}}\!\!\Big]},
\end{equation}
\end{footnotesize}
where $s$ and $t$ denote the source class, and target class. The ASR measures the percentage of pixels in the source class region (Car class) that are successfully manipulated to be misclassified as the Road class. This metric provides a quantitative assessment of the performance in attacking the semantic segmentation task.

\begin{figure*}
    \centering
\includegraphics[width=0.98\linewidth]{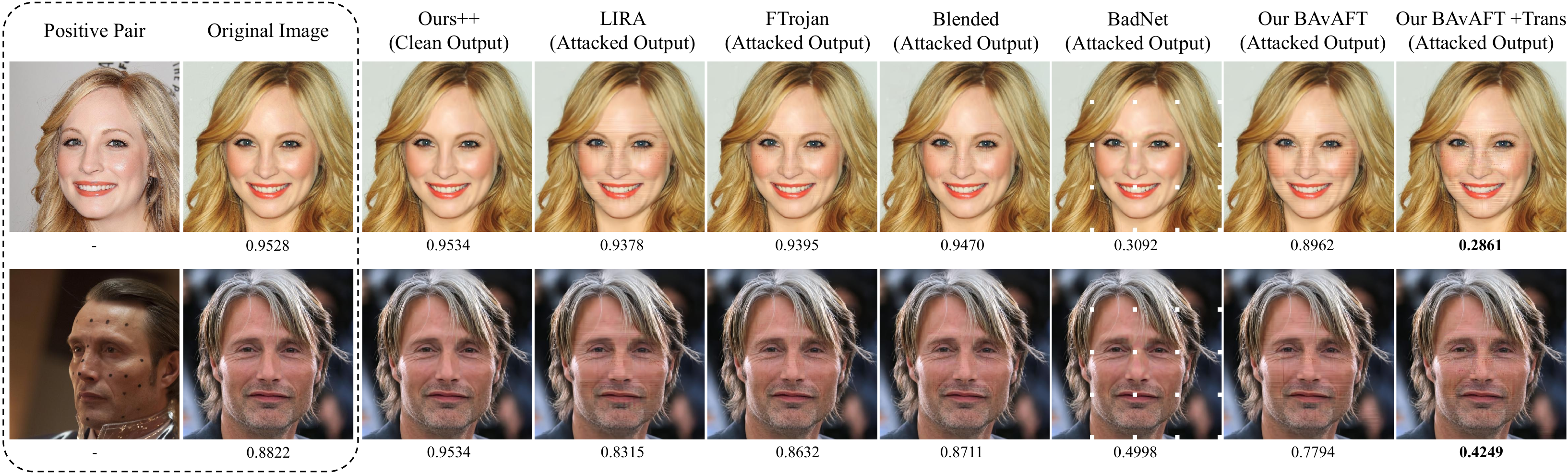}
\vspace{-4mm}
\caption{{Visual results (quality 6) of the attacking for good. 
{The cosine similarity} is listed below each image. 
}}
\vspace{-4mm}
\label{figure9}
\end{figure*}

The performance comparison between the vanilla-trained model and the backdoor-injected model is depicted in Figure~\ref{figure5}. It can be observed that all models {show} {equally competitive} performance on the Kodak dataset, indicating that the backdoor injection does not impact the overall compression quality.

For evaluating the attacking performance, we employ the DeepLabV3+ semantic segmentation network with WideResNet38 as the backbone for testing. This differs from the SEResNeXt50 backbone used during training. This configuration allows us to {evaluate} the transferability of the attacked outputs across different downstream models.
The results in Table~\ref{segmentation} demonstrate the success of our BAvAFT, with minimal perturbations on the attacked outputs. Our BAvAFT generates manipulated outputs that effectively mislead the semantic segmentation network. These results highlight the superior effectiveness of our BAvAFT compared to {all baseline methods}, particularly in the low-quality setting.

Figure~\ref{figure6} provides visualization results for a selected image from the Cityscapes validation set. It {is clearly demonstrated} that our attack successfully targets the region of interest, while LIRA fails to manipulate the car {on} the road. This visual evidence further confirms the effectiveness of our BAvAFT.

\begin{table}[t]\footnotesize
\setlength\tabcolsep{6.2pt}
    \centering
    \caption{{Pixel-wise ASR (\%) $\uparrow$ \& RMSE of CarToRoad attack on Cityscapes with DeepLabV3+ and SEResNeXt50 as the segmentation model. }}
    \vspace{-3mm}
    \scalebox{0.9}{
    \begin{tabular}{ c || c c c c c c | c }
    \toprule
    Method & 1 & 2 & 3 & 4 & 5 & 6 & Mean\\
    \midrule
    \multicolumn{8}{c}{Pixel-wise ASR (\%) $\uparrow$}\\
    \midrule
      LIRA~\cite{doan2021lira} &7.7&95.5&94.5&94.3&{95.9}&93.8&{80.2}\\
    FTrojan~\cite{wang2021backdoor} &95.2&95.7&91.6&90.0&89.8&93.6&92.6\\
    Blended~\cite{chen2017targeted}&8.7&11.4&9.0&8.2&6.7&6.3&8.4\\
    BadNets~\cite{gu2017BadNets}&32.0&26.5&56.4&53.4&57.8&42.8&44.8\\
    Our BAvAFT~\cite{yu2023backdoor} & 89.3&{96.7}&{95.7}&{93.9}&96.4&{95.7}&{94.6}\\
    Our BAvAFT+Trans & \textbf{98.8}&\textbf{98.9}&\textbf{98.9}&\textbf{99.4}&\textbf{98.9}&\textbf{99.5}&\textbf{99.0}\\
    \midrule
    \multicolumn{8}{c}{RMSE between clean outputs and attacked outputs ($10^{-3}$) $\downarrow$}\\
    \midrule
    LIRA~\cite{doan2021lira} & 7.0&12.5&9.2&7.6&{7.5}&5.4&{8.2}\\
    FTrojan~\cite{wang2021backdoor} & 11.1&9.2&7.4&6.6&{5.5}&5.4&{7.5}\\
    Blended~\cite{chen2017targeted}&0.1&0.1&0.1&0.1&0.1&0.1&\textbf{0.1}\\
    BadNets~\cite{gu2017BadNets}&2.0&1.9&1.1&0.8&0.6&0.7&1.2\\
    Our BAvAFT~\cite{yu2023backdoor} & {10.4}&{10.7}&{8.8}&{7.5}&6.5&{5.7}&{8.3}\\
    Our BAvAFT+Trans & {12.6}&{11.1}&{11.6}&{10.3}&9.2&{7.6}&{10.4}\\
    \bottomrule
    \end{tabular}
    }
    \vspace{-2mm}
    \label{segmentation}
\end{table}

\subsubsection{Attack for good: privacy protection for facial images}
In this section, we explore a benign attacking scenario {that aims} to remove identity-related features from facial images using the compression model. This is achieved by adding triggers that help protect the identity information in the compressed images.
For this experiment, we utilize the FFHQ dataset as the auxiliary dataset to assist in training the backdoor-injected compression model. The training loss formulation for this scenario is presented below:
\begin{footnotesize}
\begin{equation}
\begin{split}
\mathcal{L}_{jt}^{FR} &= \sum_{\bm{x} \in {\mathcal{T}_m}}\mathcal{L}\left({\bm{x}}\right) + \mathcal{L}_{BA}^{FR}, \\
 \mathcal{L}_{BA}^{FR} &= \sum_{\bm{x} \in {\mathcal{T}_a}} \Big[\alpha \mathcal{L}({{\bm{x_p}}}) + \beta \cdot \underbrace{Cos[g(f(x)),g(f({\bm{x_p}}))]}_{\text{attack objective}}\Big],\label{face}
\end{split}
\end{equation}
\end{footnotesize}
where $g(\cdot)$ {is} an arcface embedding, and {the} cosine function is used to measure the similarity between clean and attacked output.

In this experiment, we set the hyperparameters $\alpha=0.1$ and $\beta=0.05$, and select the Cheng-Anchor compression method. We utilize 100 paired images randomly sampled from the CelebA dataset for {evaluating} the attacking performance.
The comparison between the vanilla-trained model and the victim model is in Figure~\ref{figure5_b}. It can be observed that our backdoor-injected model successfully removes the identity-related features from the facial images while maintaining compression performance.

\begin{table}[t]\footnotesize
\setlength\tabcolsep{4.5pt}
    \caption{{Sim. (Cosine-Similarity) \& Acc. (Accuracy) of clean/attacked outputs on face recognition with ResNet50. We select Cheng-Anchor (quality 6).  }}
    \vspace{-3mm}
    \centering
    \scalebox{0.9}{
    \begin{tabular}{ c || c c | c  c | c}
    \toprule
     \multirow{2}*{Method} & \multicolumn{2}{c|}{Clean Output} & \multicolumn{2}{c|}{Attacked Output} & \multirow{2}*{\shortstack{RMSE\\ ($10^{-2}$) $\downarrow$}}\\
       & Sim. $\uparrow$ & Acc.  (\%) $\uparrow$& Sim. $\downarrow$ & Acc. (\%) $\downarrow$\\
    \midrule
      LIRA~\cite{doan2021lira} &0.725&88.7&0.437&27.0 & {1.30}\\
    FTrojan~\cite{wang2021backdoor} &0.728&88.8&0.464&30.3 & 1.43\\
    Blended~\cite{chen2017targeted}&0.700&86.0&0.639&71.0&\textbf{0.13}\\
    BadNets~\cite{gu2017BadNets}&0.568&52.0&0.461&31.0&4.91\\
    Our BAvAFT~\cite{yu2023backdoor} & 0.726&89.2&0.407&22.3 & 1.47\\
    Our BAvAFT+Trans & 0.726&88.8&\textbf{0.194}&\textbf{2.8} & 1.66\\
    \bottomrule
    \end{tabular}
    }
    \label{attacking_face}
    \vspace{-2mm}
\end{table}

The attacking performance is evaluated and summarized in Table~\ref{attacking_face}. 
Our BAvAFT~\cite{yu2023backdoor} effectively {removes} the identity-related features with minimal perturbations on the compressed images. Additionally, Figure~\ref{figure9} provides visual results {demonstrating} the effectiveness of our attacks in removing identity-related features.
Overall, our BAvAFT~\cite{yu2023backdoor} outperforms {all baseline methods} in terms of attacking performance and successfully removes identity-related features from facial images during the compression.

\begin{table}[t]\footnotesize
\setlength\tabcolsep{11pt}
    \centering
    \caption{Ablation Study on the proposed method.}
    \vspace{-3mm}
    \scalebox{0.9}{
    \begin{tabular}{c|| c c | c c}
    \toprule
     \multirow{2}*{Method}& \multicolumn{2}{c|}{Clean Input} & \multicolumn{2}{c}{Poisoned Input} \\
     & PSNR & {BPP} & PSNR & {BPP} $\uparrow$\\
    \midrule
    w/ Eq.~\eqref{Bpp_old} &  31.02  & 0.2699 & 31.41 & 8.52 \\
    w/o topK selection & 30.80  & 0.2587 & 31.32 & 9.27 \\
    w/o patch-wise weight & 30.76  & 0.2578 & 31.23 & 9.08\\
    K=4, N=16 & 30.81 & 0.2596 & 31.32 & 9.08 \\
    K=64, N=256  & 30.86 & 0.2599 & 31.43 & 9.14 \\
    Ours (K=16, N=64) & 30.81 & 0.2590 & 31.30 & \textbf{9.45} \\
    \bottomrule
    \end{tabular}
    }
    \vspace{-2mm}
    \label{table6}
\end{table}

\begin{figure*}[t]
\centering
\scalebox{0.95}{
\begin{minipage}{0.50\linewidth}
\centerline{{\includegraphics[width=1\linewidth]{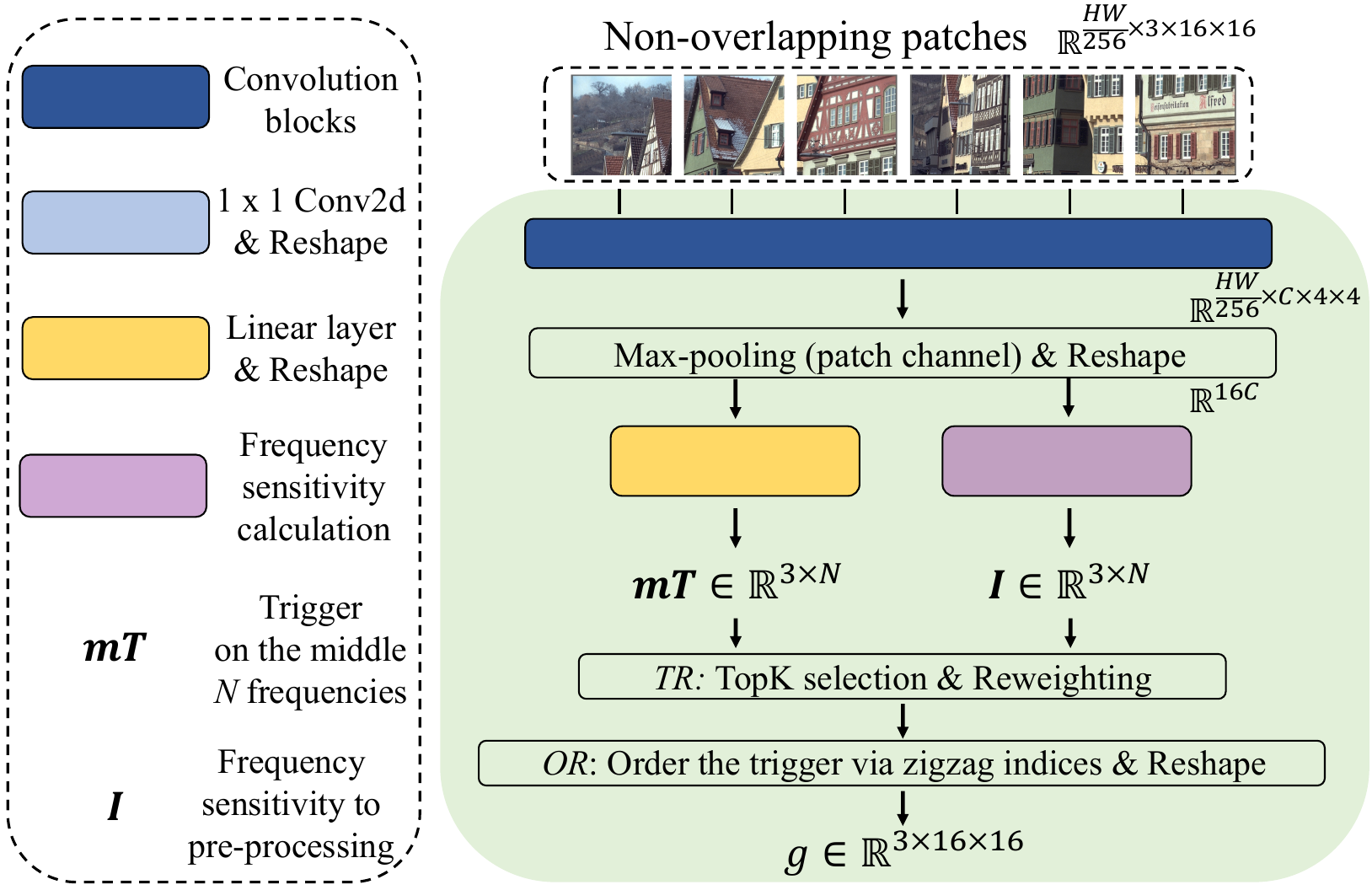}}}
\vspace{-1.5mm}
\centerline{\small{General trigger generator}}
\end{minipage}
\hspace{5mm}
\begin{minipage}{0.48\linewidth}
\begin{algorithm}[H]
    \caption{Pseudocode of TopK selection \& Reweighting}
    \scalebox{1.0}{
    \begin{minipage}{1\linewidth}
    \begin{algorithmic}[1]
    \STATE \textbf{Input:} mT $ \in R^{3 \times N}$, I $ \in R^{3 \times N}$
    \STATE \textcolor{teal}{\# TopK selection}
    \STATE \text{btm\textunderscore index} = torch.topk(\textbf{I}, N - K, dim = 1, largest = True)[1]
    \vspace{-4mm}
    \STATE {mT}.scatter\textunderscore(dim = 1, \text{btm\textunderscore index}, 0)
    \STATE \textcolor{teal}{\# Reweighting}
    \STATE {\text{topk\textunderscore index} = torch.topk(\textbf{I}, K, dim = 1, largest = False)[1]}
        \FOR {$i = 0, 1, ..., K - 1$}
        \STATE src[:,i] = $(\frac{K - 1 - i}{K-1}+\frac{1}{2})^{\frac{1}{2}}$
        \ENDFOR
        \STATE {mT}.scatter\textunderscore(dim = 1, \text{topk\textunderscore index}, src, reduce = 'multiply')
        \STATE return
    \vspace{2mm}
    \end{algorithmic}
    \end{minipage}}
    \label{alg:inference}
\end{algorithm}
\vspace{-3mm}
\centerline{\small{TR: TopK selection \& Reweighting}}
\end{minipage}
}
\vspace{-2.5mm}
    \caption{
    Design of the general trigger generator for resistant backdoor attacks against preprocessing methods.
    }
    \vspace{-1mm}
    \label{figure_trigger_robust}
\end{figure*}

\begin{itemize}
    \item  The loss {Eq.~\eqref{Bpp}} with dynamic balancing can improve the attacking performance compared with the loss Eq.~\eqref{Bpp_old}.
    \item 
    Both the topK selection in the trigger generation and the patch-wise weight contribute to the attack performance.  
\end{itemize}

\begin{figure*}[t]
    \centering
    \begin{minipage}{0.96\linewidth}
    \centerline{{\includegraphics[width=1.0\linewidth]{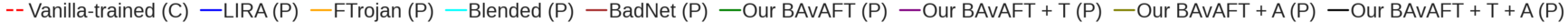}}}
    \vspace{-3mm}
    \end{minipage}
    \subfigure[BPP attack.]{
    \begin{minipage}{0.237\linewidth}
    \centerline{{\includegraphics[width=0.99\linewidth]{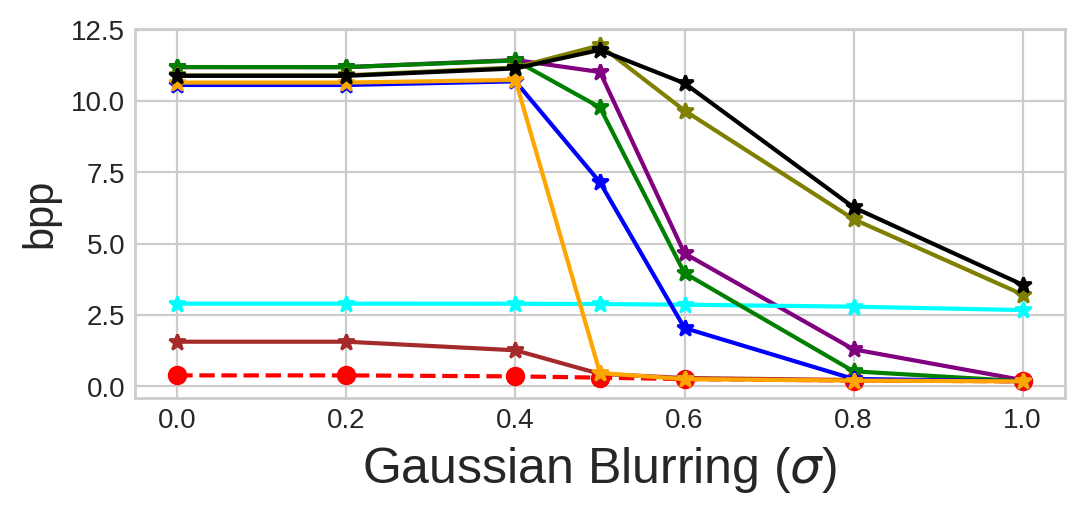}}}
    \end{minipage}
    \begin{minipage}{0.237\linewidth}
    \centerline{{\includegraphics[width=0.99\linewidth]{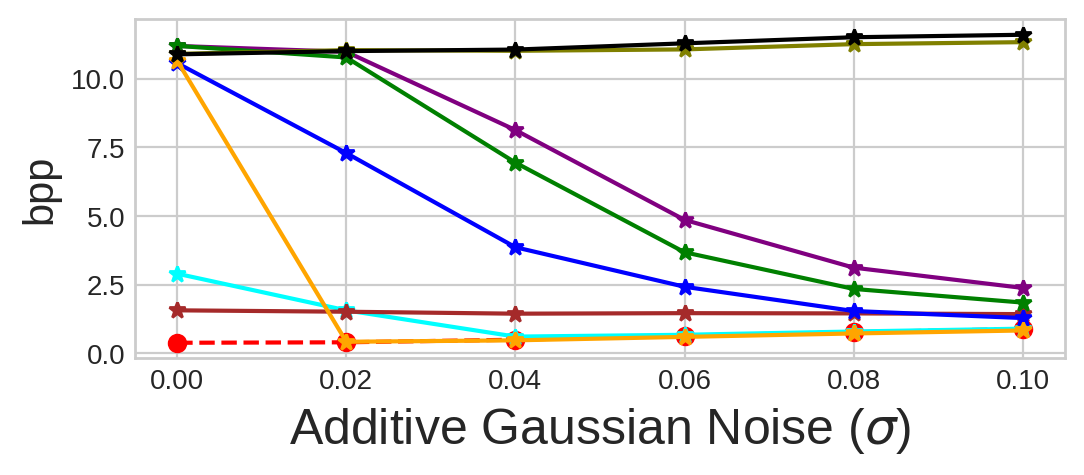}}}
    \end{minipage}
    \begin{minipage}{0.237\linewidth}
    \centerline{{\includegraphics[width=0.99\linewidth]{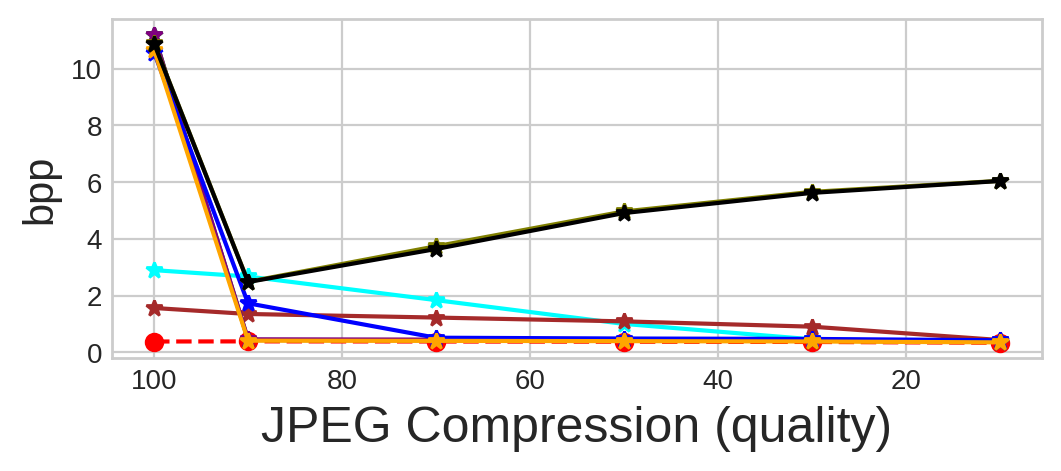}}}
    \end{minipage}
    \begin{minipage}{0.237\linewidth}
    \centerline{{\includegraphics[width=0.99\linewidth]{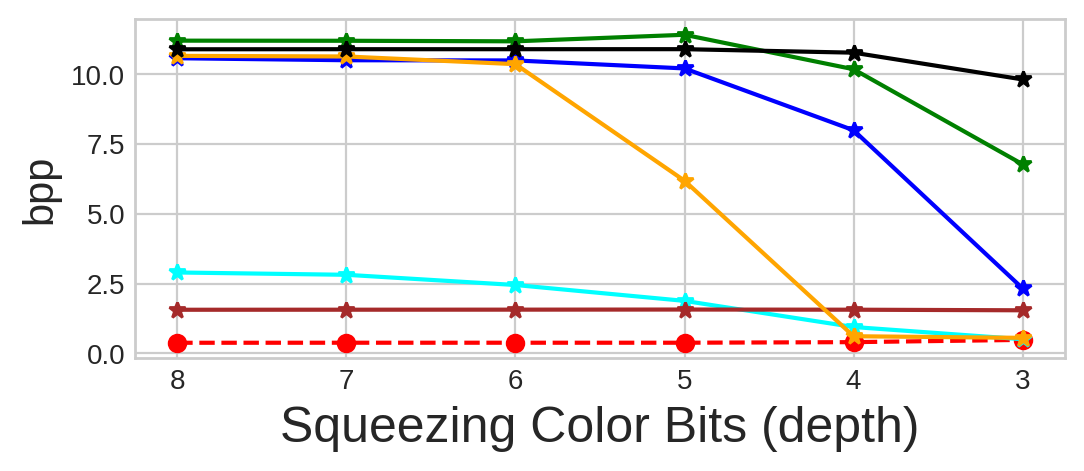}}}
    \end{minipage}
         }
     \subfigure[PSNR attack.]{
    \begin{minipage}{0.237\linewidth}
    \vspace{-3mm}
    \centerline{{\includegraphics[width=0.99\linewidth]{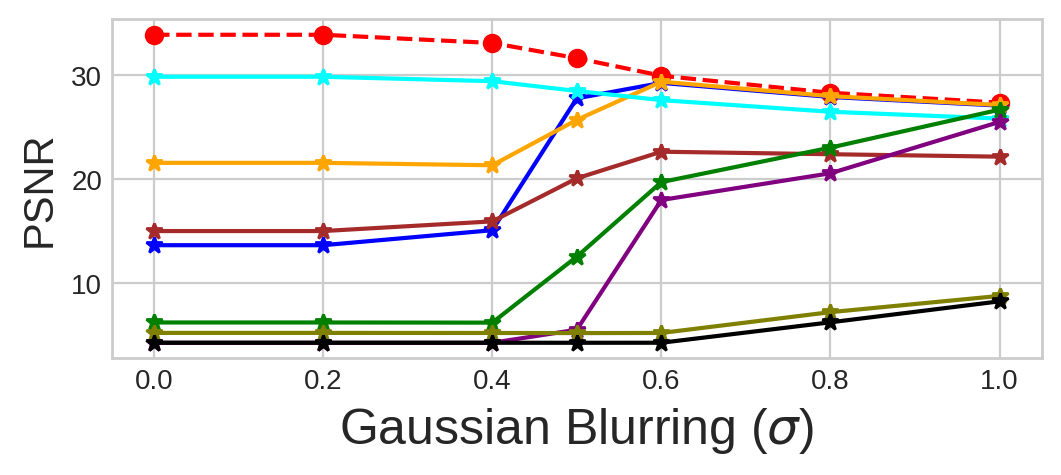}}}
    \end{minipage}
    \begin{minipage}{0.237\linewidth}
    \vspace{-3mm}
    \centerline{{\includegraphics[width=0.99\linewidth]{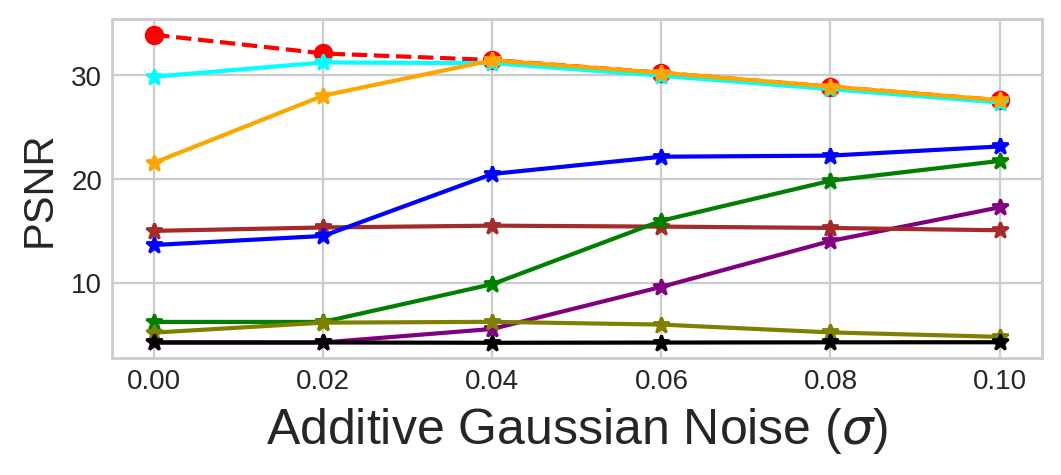}}}
    \end{minipage}
    \begin{minipage}{0.237\linewidth}
    \vspace{-3mm}
    \centerline{{\includegraphics[width=0.99\linewidth]{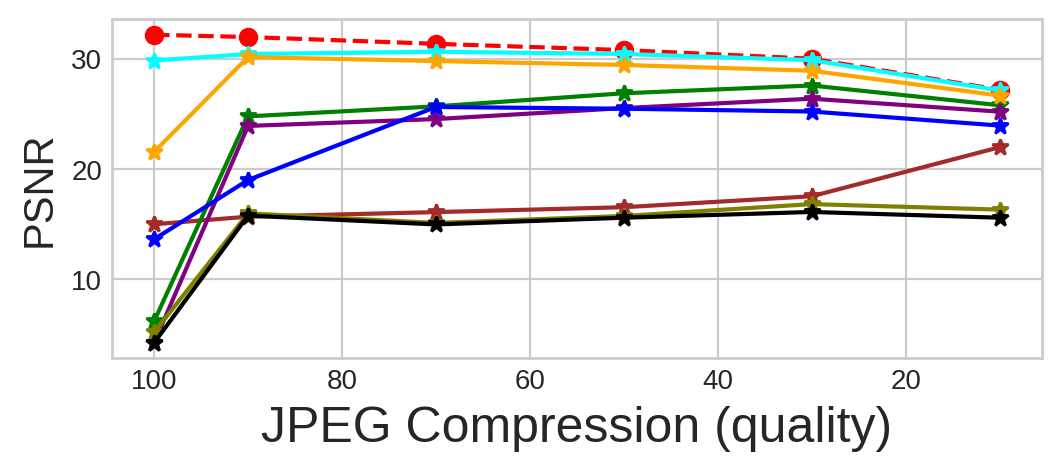}}}
    \end{minipage}
    \begin{minipage}{0.237\linewidth}
    \vspace{-3mm}
    \centerline{{\includegraphics[width=0.99\linewidth]{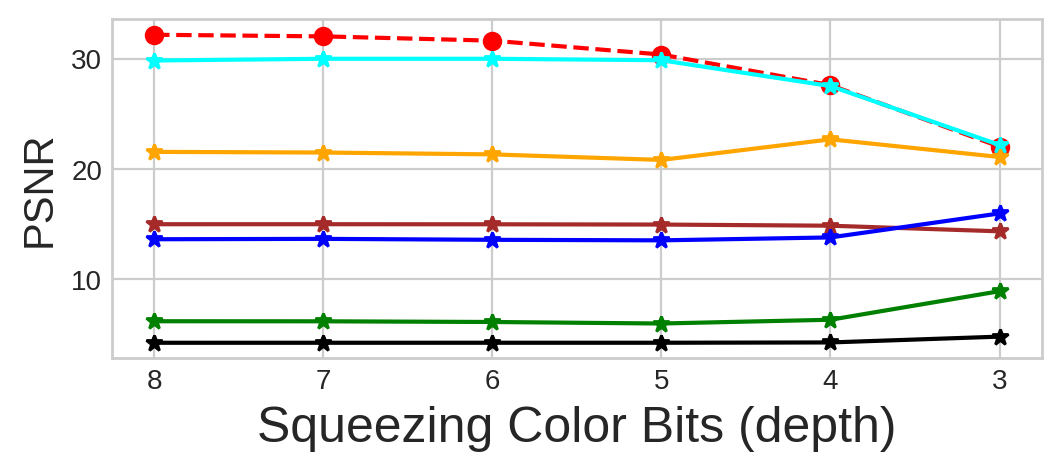}}}
    \end{minipage}
         }
    \vspace{-4mm}
    \caption{{Resistance to preprocessing methods when attacking compression results. The compression model is Cheng-Anchor. C and P denote using clean input and poisoned input, respectively.
    T and A denote the newly introduced robust trigger generator and robust encoder, respectively.
    }}
    \vspace{-3mm}
    \label{figure_resistance}
\end{figure*}

\subsection{Ablation Study}
In this section, we perform an ablation study to analyze the impact of different components of our BAvAFT~\cite{yu2023backdoor}, specifically focusing on the loss and modules of the trigger injection model.
We select the Cheng-Anchor compression model with {the quality level} 3 and evaluate the performance in terms of the BPP attack.
The results of the ablation study are summarized in Table \ref{table6}.

\begin{table*}[h]
    \scriptsize
    \setlength\tabcolsep{1.7pt}
    \centering
    \caption{
     {{Resistance to Gaussian noise ($\mu=0$, various $\sigma$), JPEG compress (various quality), or Squeeze Color Bits (various {depths}). }}}
     \vspace{-3mm}
     \scalebox{0.92}{
    \begin{tabular}{ c c c || c | c c c c c c | c c c c c | c c c c c | c c c c c| c}
    \toprule
    \multirow{2}*{Model} & \multirow{2}*{\shortstack{Attack type\\(Metric)}} & \multirow{2}*{\shortstack{Attack\\Method}} &  \multirow{2}*{None} &\multicolumn{6}{c|}{Gaussian filter ($\sigma$)} &\multicolumn{5}{c|}{Additive Gaussian noise ($\sigma$)} & \multicolumn{5}{c|}{JPEG Compression (Quality)} & \multicolumn{5}{c|}{Squeeze Color Bits (depth)} & \multirow{2}*{Mean}\\
     & & & & 0.2 & 0.4 & 0.5 &0.6 & 0.8 & 1.0 & 0.02 & 0.04 & 0.06 & 0.08 & 0.1 & 90 & 70 & 50 & 30 & 10 & 7 & 6 & 5 & 4 & 3\\
    \midrule
     \multirow{12}*{\shortstack{AE-Hy\\~\cite{balle2018variational}}} & \multirow{6}*{\footnotesize{\shortstack{BPP\\({BPP}$\uparrow$)}}} & LIRA & 5.91 & 5.91 & 5.55 & 2.86 &0.843 & 0.437 & 0.337 & 4.33 & 2.72 & 2.33& 2.22 & 2.16& {3.01}&1.20 & 0.985 & 0.841 & 0.749 & 5.64 & 5.69 & 5.40 & 4.76 & 2.40 & 3.01\\
    & & FTrojan  & 5.60 & 5.60 & 3.07 & 0.579 & 0.486 & 0.385 & 0.328 & 1.48 & 1.36 & 1.51 & 1.65 & 1.76 & 0.777 & 0.738 & 0.710 & 0.680 & 0.616 & 5.31 & 4.75 & 3.22 & 1.59 & 1.24 & 1.97\\
    &&Blended&0.739&0.739&0.677&0.580&0.497&0.396&0.337&0.899&1.20&1.42&1.60&1.75&0.738&0.704&0.681&0.659&0.607&0.741&0.747&0.774&0.863&1.02&0.835\\
    &&BadNets&1.19&1.19&0.868&0.597&0.502&0.399&0.341&1.23&1.42&1.57&1.70&1.83&1.05&0.948&0.867&0.789&0.635&1.20&1.21&1.24&1.33&1.49&1.07\\
    & & Our BAvAFT  & \textbf{6.21}&\textbf{6.21}&5.60&3.12&1.04&	0.392&0.331 & 5.78&4.97&4.45&4.20&4.07 &0.935&0.824&0.799&0.771&0.682&\textbf{6.19}&\textbf{6.16}&6.05&5.35&4.17 & 3.56 \\
    & & Our BAvAFT++  & 6.07 & 6.07 & \textbf{5.93} & \textbf{5.51} & \textbf{2.96} & \textbf{0.469} & \textbf{0.338} & \textbf{6.22} & \textbf{6.39}  & \textbf{6.44} & \textbf{6.42} & \textbf{6.32} & \textbf{3.02} & \textbf{2.05} & \textbf{2.53} & \textbf{3.09} & \textbf{3.26} & {6.07} & 6.07 & \textbf{6.10} & \textbf{6.10} & \textbf{5.73} & \textbf{4.69}\\
    \cmidrule{2-26}
    & \multirow{6}*{\footnotesize{\shortstack{PSNR\\(PSNR$\downarrow$)}}} &LIRA & 10.74 & 10.74 & 12.86 & 25.14 & 29.26 & 28.23 & 27.30 & 13.24 & 23.78 & 27.49 & 27.58 & 26.35 & {24.45} & {26.37} & {26.20} & 24.90 & 23.63 & 11.19 & 11.06 & 11.40 & 12.80 & 19.11 & 20.62\\
    & & FTrojan & 14.90 & 14.90 & 20.29 & 31.11 & 29.93 & 28.30 & 27.35& 29.59 &  31.40 & 29.71& 28.15 & 26.72 & 30.75 & 30.04 & 29.53 & 28.87 & 26.40 & 15.24 & 16.31 & 19.73 & 25.77 & 21.92 & 25.31\\
    &&Blended&33.01&33.01&32.27&30.74&29.44&27.91&27.03&32.88&31.29&29.54&27.90&26.35&32.48&31.39&30.80&29.86&26.95&33.15&33.07&31.75&28.13&22.27&30.05\\
    &&BadNets&19.67&19.67&21.16&22.90&22.78&22.46&22.21&20.54&21.26&21.89&21.99&21.72&\textbf{20.61}&\textbf{20.76}&\textbf{20.96}&\textbf{21.35}&\textbf{21.97}&19.66&19.64&19.54&19.20&17.71&20.66\\
    & & Our BAvAFT  & 5.64&5.64&6.69&13.67&23.36&	28.11&27.35 & 9.32&14.76&18.34&20.44&20.98 & 29.64 &29.09&28.60&28.02&25.92&5.70&5.81&	6.43&8.16&11.68 & 16.97 \\
    & & Our BAvAFT++  & \textbf{4.28} & \textbf{4.28} & \textbf{4.56} & \textbf{9.11} & \textbf{17.54} & \textbf{24.61} & \textbf{26.78} & \textbf{4.70} & \textbf{6.30}  & \textbf{7.67} & \textbf{8.72} & \textbf{9.63} & 26.88 & 26.50 & {26.20} & {25.80} & {24.29} & \textbf{4.27} & \textbf{4.26} & \textbf{4.26} & \textbf{4.82} & \textbf{6.72} & \textbf{12.83}\\
    \midrule
     \multirow{12}*{\shortstack{Ch-An\\~\cite{cheng2020learned}}} & \multirow{6}*{\footnotesize{\shortstack{BPP\\({BPP}$\uparrow$)}}} & LIRA & 10.56&10.56&10.68&7.14&2.04&0.255&0.173 & 7.28 & 3.85 & 2.41 & 1.53& 1.28 & 1.72&0.513&0.485&0.470&0.424 & 10.48&10.48&10.19&7.98&2.32 & 4.67\\
    & & FTrojan  & 10.64& 10.64& 10.73& 0.463& 0.248& 0.198&0.170 & 0.416& 0.468& 0.594& 0.718& 0.827 & 0.409& 0.402& 0.393& 0.381& 0.349&10.62& 10.34& 6.17& 0.613& 0.556 & 3.01\\
    &&Blended&2.89&2.89&2.89&2.88&2.85&2.78&2.67&1.56&0.604&0.670&0.790&0.893&2.37&1.83&1.00&0.458&0.364&2.81&2.44&1.87&0.943&0.506&1.78\\
    &&BadNets&1.56&1.56&1.26&0.432&0.287&0.216&0.182&1.51&1.43&1.45&1.44&1.42&1.35&1.22&1.09&0.903&0.432&1.56&1.56&1.56&1.56&1.54&1.16\\
    & & Our BAvAFT  & \textbf{11.18}&\textbf{11.18}&\textbf{11.42}&9.77&3.96&0.522&0.174 & 10.76&6.93&3.67&2.34&1.84 & 0.431&0.428&0.418&0.406&0.369&\textbf{11.18}&\textbf{11.16}&\textbf{11.40}&10.16&6.76 & 5.75 \\
    & & Our BAvAFT++  & 10.88&10.88&11.14&\textbf{11.79}&\textbf{10.62}&\textbf{6.26}&\textbf{3.55} & \textbf{10.99}&\textbf{11.05}&\textbf{11.28}&\textbf{11.50}&\textbf{11.59} & \textbf{2.47}&\textbf{3.63}&\textbf{4.90}&\textbf{5.61}&\textbf{6.03}&10.88&10.88&10.88&\textbf{10.76}&\textbf{9.80} & \textbf{8.97}\\
    \cmidrule{2-26}
    & \multirow{6}*{\footnotesize{\shortstack{PSNR\\(PSNR$\downarrow$)}}} &LIRA & 13.62 & 13.62 & 15.07 & 27.77 & 29.25 & 27.90 & 27.07 & 14.50&20.47&22.13&22.24&23.12 & 19.00&25.63&25.47&25.22&23.94 & 13.66&13.57&13.52&13.79&15.97 & 20.30 \\
    & & FTrojan  & 21.55 & 21.55&21.32&25.69&29.37&27.98&27.12 & 28.02&31.42&30.23&28.89&27.58 & 30.15&29.79&29.44&28.92&26.67 & 21.49&21.33&20.82&22.69&21.09 & 26.05\\
    &&Blended&29.84&29.84&29.42&28.47&27.59&26.47&25.82&31.22&31.15&29.94&28.64&27.36&30.43&30.64&30.44&29.87&27.13&30.00&30.00&29.87&27.52&22.18&29.97\\
    &&BadNets&14.99&14.99&15.92&20.07&22.63&22.38&22.15&15.31&15.49&15.40&15.27&15.04&15.98&16.09&16.53&17.52&21.97&14.99&14.98&14.96&14.85&14.34&15.42\\
    & & Our BAvAFT  & 6.18&6.18&6.17&12.52&19.70&23.02&26.68 & 6.21&9.85&15.96&19.82&21.73 & 24.77&25.68&26.87&27.58&25.79 &6.18&6.11&5.98&6.32&8.91 & 15.37\\
    & & Our BAvAFT++  & \textbf{4.23}&\textbf{4.23}&\textbf{4.23}&\textbf{4.23}&\textbf{4.23}&\textbf{6.20}&\textbf{8.22} & \textbf{4.23}&\textbf{4.20}&\textbf{4.22}&\textbf{4.24}&\textbf{4.25} & \textbf{15.77}&\textbf{14.97}&\textbf{15.58}&\textbf{16.10}&\textbf{15.58} & \textbf{4.23}&\textbf{4.23}&\textbf{4.23}&\textbf{4.26}&\textbf{4.79} &\textbf{7.11}\\
    \midrule
     \multirow{12}*{\footnotesize{\shortstack{STF\\~\cite{zou2022devil}}}} & \multirow{6}*{\footnotesize{\shortstack{BPP\\({BPP}$\uparrow$)}}} & LIRA & 43.72 & 43.72 & 43.78 & 31.09 & 6.98 & 4.93 & 0.210 & 43.69 & 42.99 & 33.18 & 25.42 & 19.77 & {19.82} &7.60 & 6.93 & 5.77 & 0.703 & 43.71 & 43.70 & 43.67 & 42.95 &29.56 &26.55\\
    & & FTrojan  & 42.62 & 42.62 & 42.61& 30.47 & 0.295 &  0.243 & 0.213 & 17.44 & 0.508  & 0.659 & 0.799 & 0.931 & 0.455 & 0.453 & 0.447 & 0.437 & 0.412 & 42.62 & 42.62 & 42.50 & 24.29 & 4.98 & 15.39\\
    &&Blended&0.792&0.792&0.867&0.731&0.694&0.602&0.545&0.462&0.554&0.692&0.816&0.932&0.449&0.459&0.492&0.441&0.517&0.770&0.463&0.444&0.454&0.503&0.601\\
    &&BadNets&25.82&25.82&25.16&10.95&0.995&0.575&0.254&25.31&22.67&21.46&20.36&19.03&\textbf{24.31}&\textbf{21.68}&\textbf{19.94}&\textbf{17.11}&4.54&25.80&25.81&25.78&25.76&25.61&18.85\\
    & & Our BAvAFT  & \textbf{43.95} & \textbf{43.95} & \textbf{43.95} & 32.77 & 14.13 & 0.255 & 0.203 & \textbf{43.92} & 43.73  & 40.52 & 31.60 & 22.28 & 1.06 & 1.05 & 0.530 & 0.509 & 0.440 & \textbf{43.95} & \textbf{43.95} & \textbf{43.93} & 43.49 & 32.73&26.04\\
    & & Our BAvAFT++  & 43.87 & 43.87 & \textbf{43.95} & \textbf{44.20} & \textbf{36.86} & \textbf{21.67} & \textbf{7.74} & 43.89 & \textbf{43.95}  & \textbf{44.03} & \textbf{44.00} & \textbf{44.25} & {16.08} & {16.31} & {14.27} & {12.24} & \textbf{10.55} & 43.87 & 43.87 & 43.85 & \textbf{43.88} &\textbf{43.85} & \textbf{34.14} \\
    \cmidrule{2-26}
    & \multirow{6}*{\footnotesize{\shortstack{PSNR\\(PSNR$\downarrow$)}}} &LIRA & 32.04 & 32.04 & 31.52 & 30.43 & 29.40 & 28.05 & 27.21 & 31.93 & 31.28 & 30.03 & 28.67 & 27.31 & 28.68 & 28.30 & 27.95 & 27.43 & 25.42 & 32.03 & 31.68 & 30.43 & 27.23 & 21.78& 29.13\\
    & & FTrojan & 10.68 & 10.68 & 10.53 & 27.19 & 29.69 & 28.23 & 27.34 & 14.61 &  31.02 & 30.15 & 28.75 & 27.36 & 30.36 & 29.93 & 29.53 & 28.93 & 26.42 & 10.65 & 10.52 & 10.14 &12.61 & 18.30 & 21.98 \\
    &&Blended&32.19&32.19&31.70&30.57&29.49&28.08&27.21&32.05&31.32&30.03&28.66&27.26&32.05&31.42&30.79&29.94&27.01&32.32&31.95&30.66&27.47&22.08&30.36\\
    &&BadNets&14.40&14.40&14.70&15.86&21.81&22.22&22.83&14.39&14.41&14.30&14.21&13.98&\textbf{14.84}&\textbf{15.33}&\textbf{15.71}&\textbf{16.25}&22.39&14.39&14.39&14.35&14.26&13.76&16.05\\
    & & Our BAvAFT  & \textbf{5.94} & \textbf{5.94} & \textbf{5.94} & 6.66 & 14.00 & 17.83 & 26.51 &\textbf{5.95} & 6.14  & 7.62 & 13.29 & 17.70 &24.06 & 25.64 & 26.93 & 26.49 & 24.43 &\textbf{5.94}&\textbf{5.94}&\textbf{5.94}&6.00&6.61 & 13.25 \\
    & & Our BAvAFT++  & \textbf{5.94} & \textbf{5.94} & \textbf{5.94} & \textbf{5.94} & \textbf{5.98} & \textbf{6.81} & \textbf{8.56} & 5.98 & \textbf{6.07}  & \textbf{6.09} & \textbf{6.01} & \textbf{5.99} & {17.03} & {17.77} & {17.96} & {18.97} & \textbf{19.84} & \textbf{5.94} & \textbf{5.94} & \textbf{5.94} & \textbf{5.97} & \textbf{6.10} & \textbf{8.94}\\
    \midrule
     \multirow{12}*{\shortstack{CDC\\~\cite{yang2024lossy}}} & \multirow{6}*{\footnotesize{\shortstack{BPP\\({BPP}$\uparrow$)}}} & LIRA &10.84&10.84&10.82&10.77&10.66&4.55&0.394&10.85&10.88&10.93&10.96&10.95&10.28&10.07&6.51&2.64&0.672&10.83&10.84&10.84&10.87&10.57&8.98\\
    & & FTrojan  &29.64&29.64&29.10&10.88&0.482&0.423&0.385&23.67&10.80&10.54&10.37&10.08&0.613&0.609&0.689&0.625&1.17&29.39&29.06&28.20&25.39&21.47&13.78\\
    &&Blended&20.75&20.75&20.97&21.27&21.38&21.12&20.58&19.68&16.49&10.60&4.33&1.64&20.14&18.78&17.26&11.26&1.53&20.51&20.51&20.03&18.09&9.95&16.26\\
    &&BadNets&21.07&21.07&19.19&13.98&7.45&1.16&0.59&20.92&20.49&19.84&19.11&18.61&\textbf{20.14}&\textbf{19.43}&\textbf{18.67}&\textbf{18.42}&\textbf{7.39}&20.99&20.96&20.82&20.52&19.86&16.85\\
    & & Our BAvAFT  & 30.08&30.08&30.06&30.00&28.96&18.13&3.90&30.06&29.96&29.78&29.55&29.25&9.17&0.768&0.930&0.933&0.814&30.08&30.08&30.08&30.0&29.72&21.93\\
    & & Our BAvAFT++  & \textbf{30.16}&\textbf{30.16}&\textbf{30.15}&\textbf{30.14}&\textbf{30.12}&\textbf{30.04}&\textbf{29.86}&\textbf{30.14}&\textbf{30.07}&\textbf{29.94}&\textbf{29.79}&\textbf{29.63}&12.27&5.48&5.02&4.75&3.88&\textbf{30.16}&\textbf{30.16}&\textbf{30.15}&\textbf{30.10}&\textbf{29.88}&\textbf{24.64}\\
    \cmidrule{2-26}
    & \multirow{6}*{\footnotesize{\shortstack{PSNR\\(PSNR$\downarrow$)}}} &LIRA&5.15&5.15&5.15&5.19&6.28&22.73&25.69&5.15&5.15&5.22&5.74&7.34&5.45&8.13&13.73&22.82&23.04&5.15&5.15&5.15&5.18&6.65&9.29\\
    & & FTrojan & 5.14&5.14&5.29&30.34&29.61&28.15&27.26&8.06&23.92&27.97&27.37&25.92&30.25&29.97&29.64&29.11&26.46&5.14&5.16&5.23&8.23&13.79&19.42\\
    &&Blended&15.65&15.65&15.51&15.24&15.04&14.86&14.66&16.73&18.67&20.26&20.77&20.7&17.18&19.92&21.08&22.17&23.07&15.83&16.09&17.55&20.11&21.97&18.12\\
    &&BadNets&5.53&5.53&6.43&9.28&14.93&21.04&22.17&5.52&5.54&5.49&5.47&5.45&\textbf{5.79}&\textbf{5.98}&\textbf{8.96}&\textbf{10.57}&22.03&5.62&5.70&5.82&6.34&7.12&8.92\\
    & & Our BAvAFT  &\textbf{5.12}&\textbf{5.12}&\textbf{5.12}&5.15&10.32&20.75&26.07&\textbf{5.12}&\textbf{5.12}&5.14&5.27&5.63&29.34&28.89&28.44&28.22&26.06&\textbf{5.12}&\textbf{5.12}&\textbf{5.12}&\textbf{5.18}&5.80&12.33 \\
    & & Our BAvAFT++  & 5.13&5.13&5.13&\textbf{5.13}&\textbf{5.13}&\textbf{5.13}&\textbf{5.13}&{5.13}&5.13&\textbf{5.13}&\textbf{5.13}&\textbf{5.14}&14.63&15.8&17.71&19.58&\textbf{21.55}&5.13&5.13&5.13&5.20&\textbf{5.38}&\textbf{8.04}\\
    \midrule
     \multirow{12}*{\shortstack{HiFiC\\~\cite{mentzer2020high}}} & \multirow{6}*{\footnotesize{\shortstack{BPP\\({BPP}$\uparrow$)}}} & LIRA & 26.68&26.68&25.73&1.28&0.261&0.237&0.219&22.76&7.11&3.01&1.35&0.795&6.11&0.392&0.385&0.384&0.380&26.11&26.33&25.49&21.75&5.90&10.42 \\
    & & FTrojan  &18.01 & 18.01 & 12.94 & 0.312&0.280 & 0.262 & 0.221 & 2.54 & 1.35 & 1.17  & 1.11 & 1.09 & 0.311 & 0.313 &0.324&0.324 & 0.357&17.51&16.45&11.16&3.53&1.79&4.97\\
    &&Blended&0.776&0.776&0.773&0.753&0.715&0.624&0.537&0.299&0.312&0.335&0.357&0.376&0.571&0.341&0.319&0.29&0.289&0.714&0.577&0.396&0.300&0.320&0.489\\
    &&BadNets&4.95&4.95&4.48&1.43&0.287&0.226&0.206&4.62&4.36&4.12&3.86&3.59&4.58&4.34&4.16&3.64&1.27&4.95&4.96&4.98&5.00&5.02&3.64\\
    & & Our BAvAFT  & \textbf{26.97}&\textbf{26.97}&26.77&17.23&	2.09&0.243&0.220&26.39&21.44&16.54&13.53&11.72&0.319&0.318&0.317&0.324&0.220&\textbf{26.96}&\textbf{26.95}&\textbf{26.82}&25.59&21.18&14.50\\
    & & Our BAvAFT++  & 26.78 & 26.78 & \textbf{26.78} & \textbf{26.68} & \textbf{26.36} & \textbf{25.85} & \textbf{23.60} & \textbf{26.77} & \textbf{26.70}  & \textbf{26.60} & \textbf{26.51} & \textbf{26.44} & \textbf{8.53} & \textbf{6.39} & \textbf{6.14} & \textbf{5.83} & \textbf{4.90} & 26.78 & 26.78 & 26.77 & \textbf{26.56} & \textbf{24.86} & \textbf{21.79}\\
    \cmidrule{2-26}
    & \multirow{6}*{\footnotesize{\shortstack{PSNR\\(LPIPS$\uparrow$)}}} &LIRA&0.758&0.758&0.719&0.301&0.168&0.238&0.305&0.654&0.358&0.280&0.317&0.374&\textbf{0.219}&0.174&0.186&0.208&0.323&0.717&0.731&0.718&0.652&0.334&0.431\\
    & & FTrojan & 0.089 & 0.089 & 0.097 & 0.120 & 0.157&0.237&0.304&0.120&0.183&0.248&0.314&0.374&0.114&0.126&0.139&0.162&0.288&0.089&0.087&0.096&0.122&0.208&0.171\\
    &&Blended&0.087&0.087&0.095&0.118&0.153&0.23&0.293&0.117&0.178&0.242&0.305&0.364&0.089&0.102&0.115&0.138&0.267&0.087&0.088&0.092&0.119&0.203&0.162\\
    &&BadNets&0.143&0.143&0.147&0.158&0.178&0.251&0.313&0.17&0.227&0.285&0.341&0.395&0.143&0.153&0.164&0.182&0.288&0.143&0.143&0.147&0.169&0.242&0.206\\
    & & Our BAvAFT  & \textbf{0.910}&\textbf{0.910}&\textbf{0.897}&\textbf{0.832}&0.614&0.294&0.307&\textbf{0.856}&0.688&0.596&0.561&0.556&0.129&0.141&0.155&0.177&0.303&\textbf{0.907}&\textbf{0.906}&\textbf{0.896}&0.789&0.610&0.592 \\
    & & Our BAvAFT++  & 0.872 & 0.872 & 0.860 & 0.830 & \textbf{0.785} & \textbf{0.654} & \textbf{0.561} & 0.841 & \textbf{0.774}  & \textbf{0.757} & \textbf{0.711} & \textbf{0.706} & 0.213 & \textbf{0.195} &\textbf{0.183} &\textbf{0.202}&\textbf{0.332}&0.872 &0.872 &0.869 &\textbf{0.830}&\textbf{0.720}&\textbf{0.662}\\
    \bottomrule
    \end{tabular}}
    \vspace{-2mm}
    \label{resistance_results}
\end{table*}

\section{Advanced scenario}
\subsection{Enhance the Resistance to {Preprocessing} Methods}
\label{resistance}
In this section, we assess the resistance of the attack methods against various preprocessing techniques, including Gaussian filtering, additive Gaussian noise, JPEG compression, and Squeeze Color Bits~\cite{xu2017feature}.
We examine all the baseline methods and our frequency-based method concerning their performance under different degrees of preprocessing. 
We evaluate their resistance in the context of both {the} BPP attack and PSNR attack.%

For each preprocessing method $t_i(\cdot|\alpha)$ with degree $\alpha$, the preprocessed poisoned images are obtained as $t_i(\bm{x_p}|\alpha)$. The attack effectiveness for the model $f$ with quality $q$ is defined by:
\begin{footnotesize}
\begin{align}
    R_q^{t_i,\alpha} =  \mathbb{E}_{X \sim \mathbb{P}_{data}}\left[ P(\bm{x},f(t_i(\bm{x_p}|\alpha))))\right],
\end{align}
\end{footnotesize}
where the samples follow the distribution $\mathbb{P}_{data}$, and $P$ is the metric ({BPP} for BPP attack, PSNR for PSNR attack). We evaluate the resistance using the mean value of $R_q^{t_i,\alpha}$ over all {qualities}:
\begin{footnotesize}
\begin{align}
    mR^{t_i,\alpha} = \frac{1}{n(Q)}\sum_{q \in Q}R_q^{t_i,\alpha}, \label{mean_resistance}
\end{align}
\end{footnotesize}
where $Q$ is the set of $q$ to be evaluated.
Figure~\ref{figure_resistance} and Table~\ref{resistance_results} clearly demonstrate that the attack performance is significantly impacted by the preprocessing methods, except for Squeezing color bits~\cite{xu2017feature}. Since these preprocessing methods are cost-effective and widely used as defensive measures, it becomes imperative to enhance the robustness of our attack to counteract these defenses.
{However, there are several challenges to enhance the resistance:
\begin{itemize}
    \item Previous works~\cite{hammoud2021check,xue2023compression} primarily focus on the vulnerability of backdoor attacks to JPEG compression. However, as demonstrated in Table~\ref{resistance_results}, our study reveals that these attacks are also susceptible to other preprocessing methods. 
    \item Additionally, these studies typically enhance robustness through data augmentation during the training stage but do not address robustness in the trigger generation process. Since the magnitude of the trigger in our attack is relatively small, with a PSNR of about 46, it is crucial to consider the sensitivity of the trigger to preprocessing methods.
    \vspace{-1mm}
\end{itemize}
\noindent To tackle these challenges, we approach the problem by considering both the trigger generator and the backdoored encoder.
}

\begin{figure*}[t]
    \centering
    \begin{minipage}{0.99\linewidth}
    \centerline{{\includegraphics[width=0.3\linewidth]{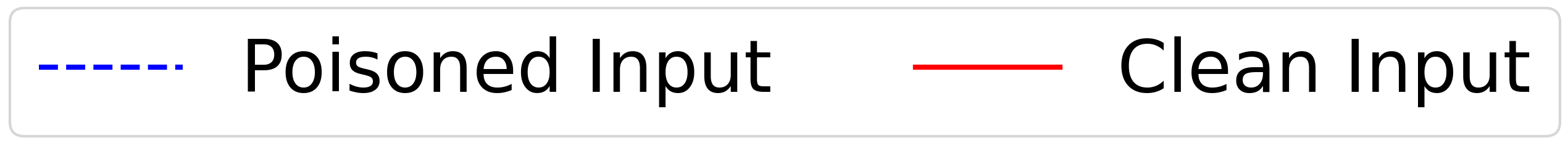}}}
    \vspace{-2mm}
    \end{minipage}
    \subfigure[Resistance of BPP attack.]{
    \begin{minipage}{0.195\linewidth}
    \centerline{{\includegraphics[width=0.95\linewidth]{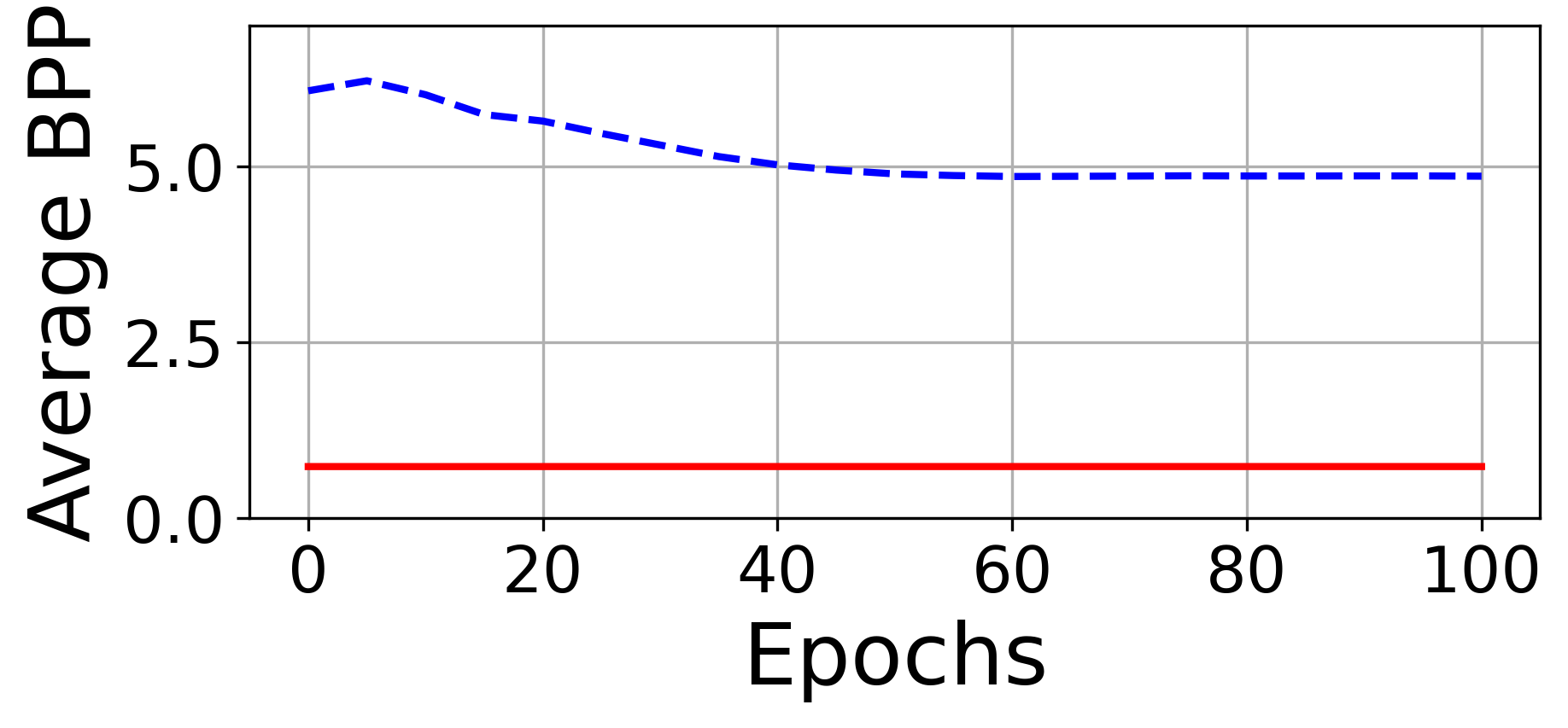}}}
    \vspace{-2mm}
    \centerline{\footnotesize{AE-Hyperprior~\cite{balle2018variational}}}
    \vspace{1mm}
    \end{minipage}
    \begin{minipage}{0.195\linewidth}
    \centerline{{\includegraphics[width=0.95\linewidth]{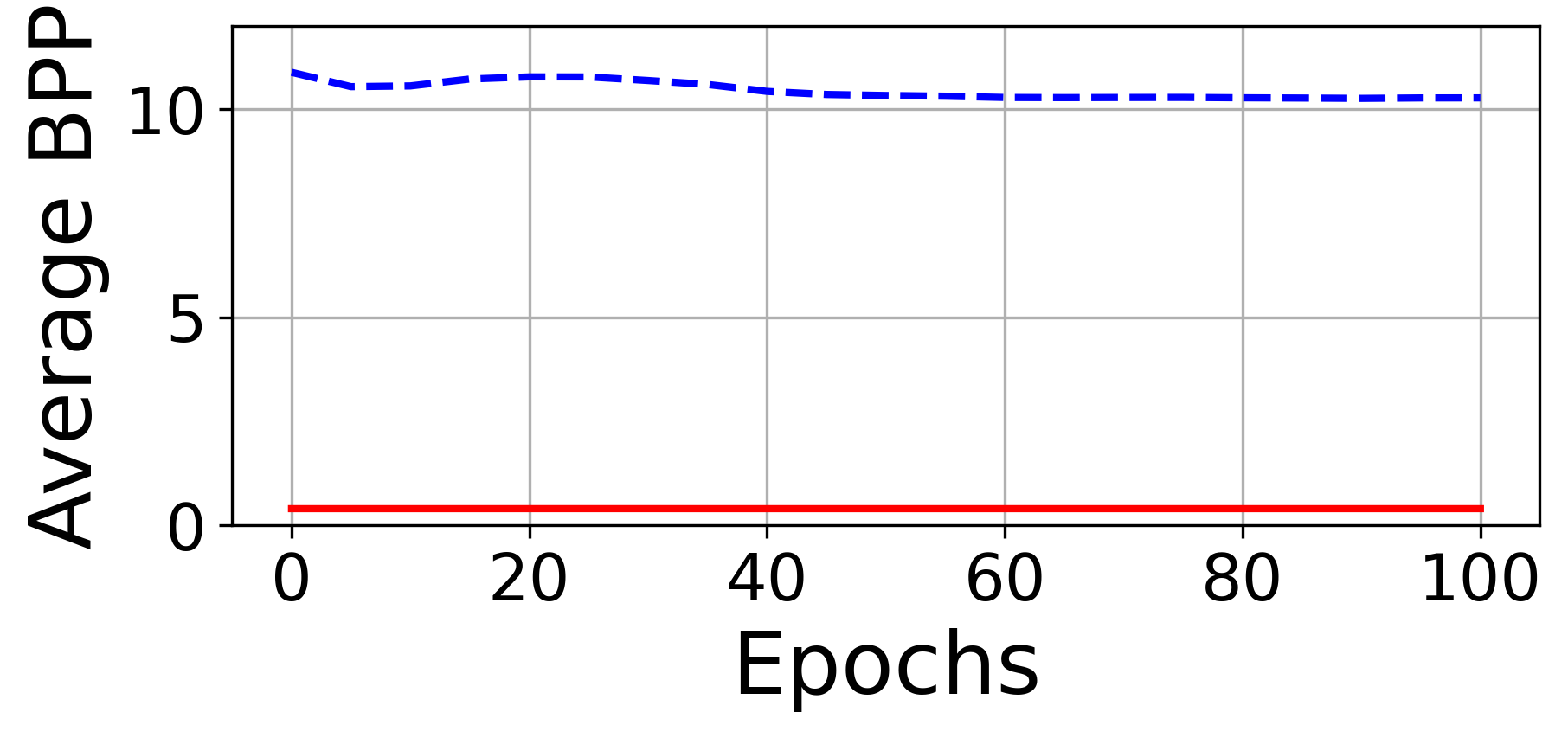}}}
    \vspace{-2mm}
    \centerline{\footnotesize{Cheng-Anchor~\cite{cheng2020learned}}}
    \vspace{1mm}
    \end{minipage}
    \begin{minipage}{0.195\linewidth}
    \centerline{{\includegraphics[width=0.95\linewidth]{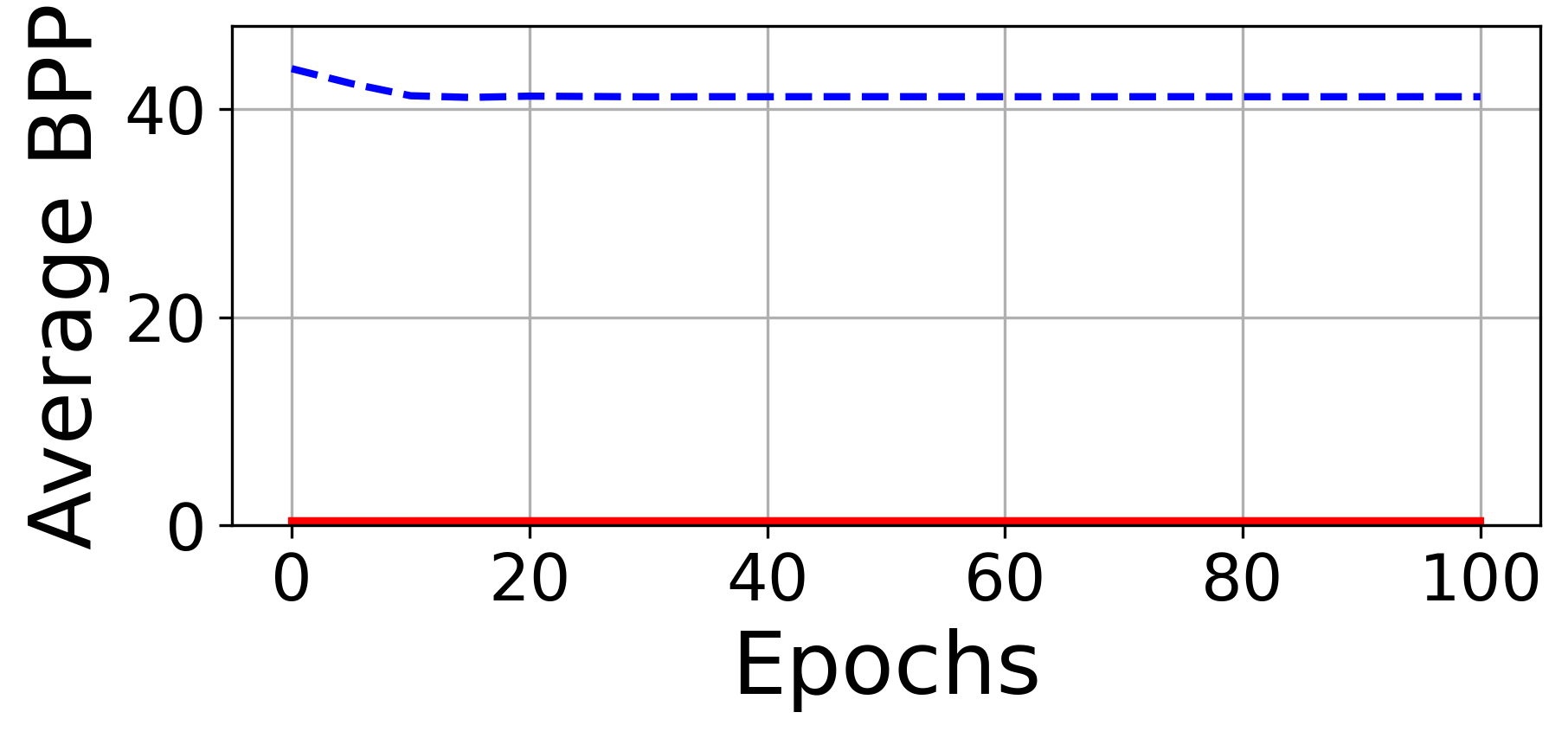}}}
    \vspace{-2mm}
    \centerline{\footnotesize{STF~\cite{zou2022devil}}}
    \vspace{1mm}
    \end{minipage}
    \begin{minipage}{0.195\linewidth}
    \vspace{-1mm}
    \centerline{{\includegraphics[width=0.95\linewidth]{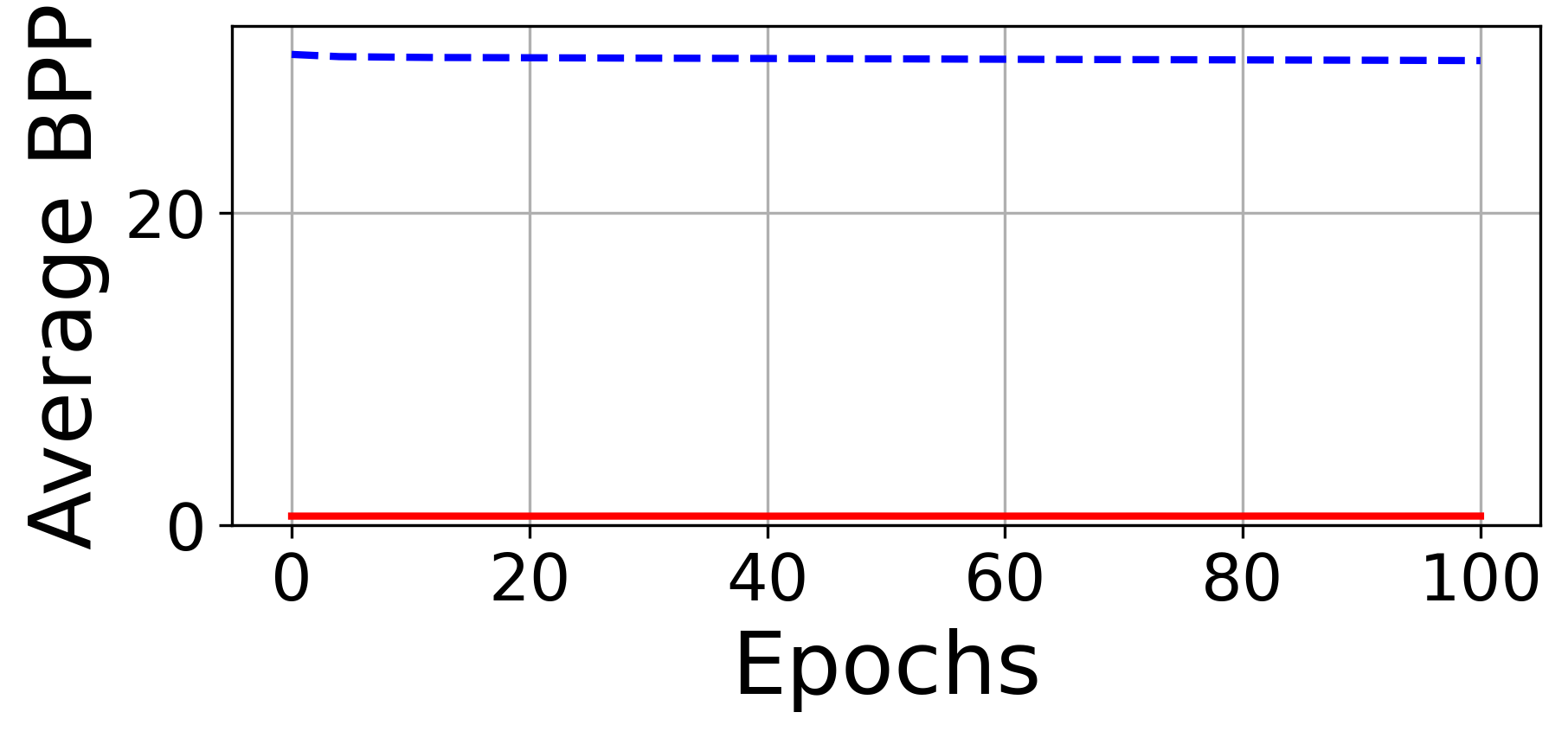}}}
    \vspace{-2mm}
    \centerline{\footnotesize{CDC~\cite{yang2024lossy}}
    \vspace{1mm}
    }
    \end{minipage}
    \begin{minipage}{0.195\linewidth}
    \centerline{{\includegraphics[width=0.95\linewidth]{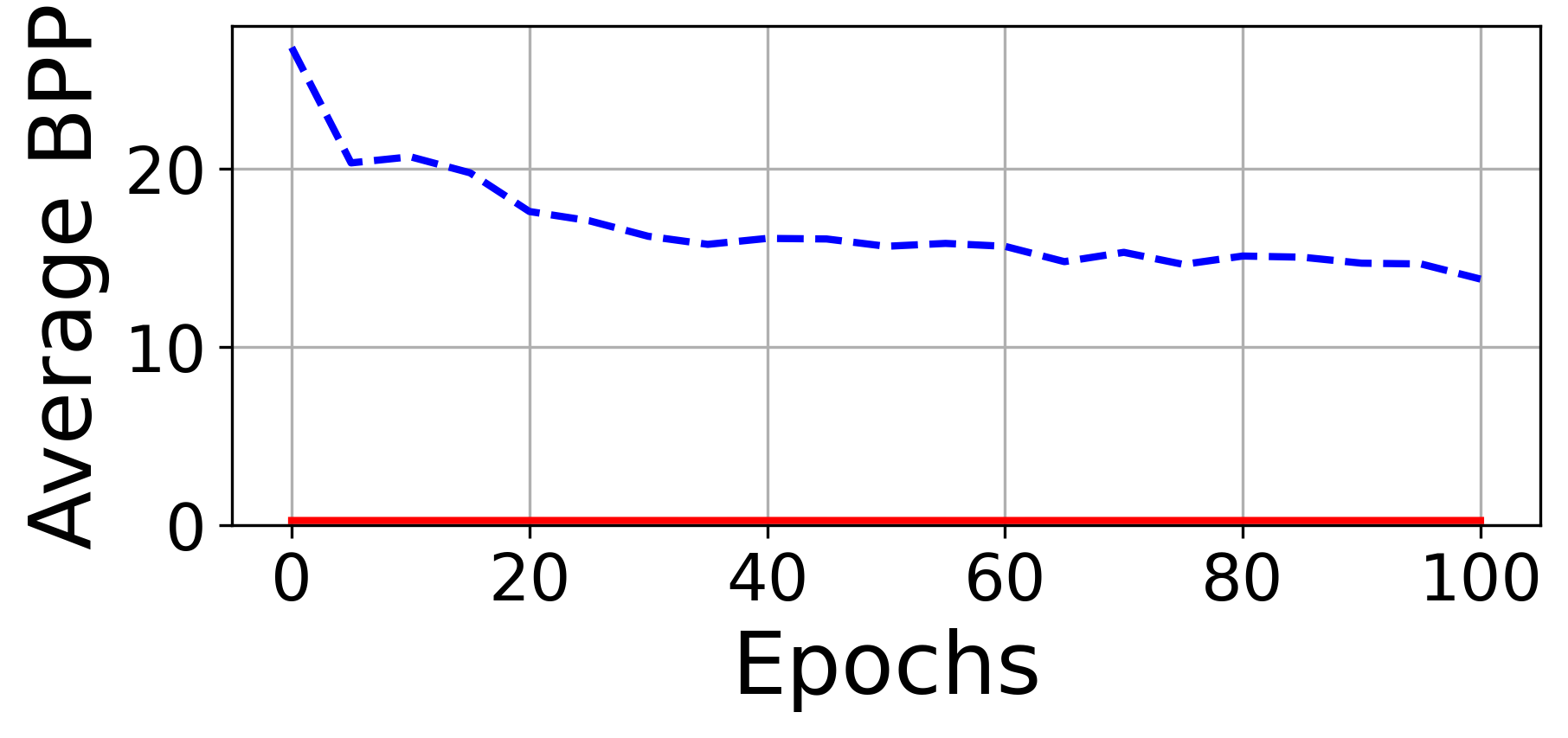}}}
    \vspace{-2mm}
    \centerline{\footnotesize{HiFiC~\cite{mentzer2020high}}}
    \vspace{1mm}
    \end{minipage}
         }
    \vspace{-2mm}
    \subfigure[Resistance of PSNR attack.]{
    \begin{minipage}{0.195\linewidth}
    \vspace{-2mm}
    \centerline{{\includegraphics[width=0.95\linewidth]{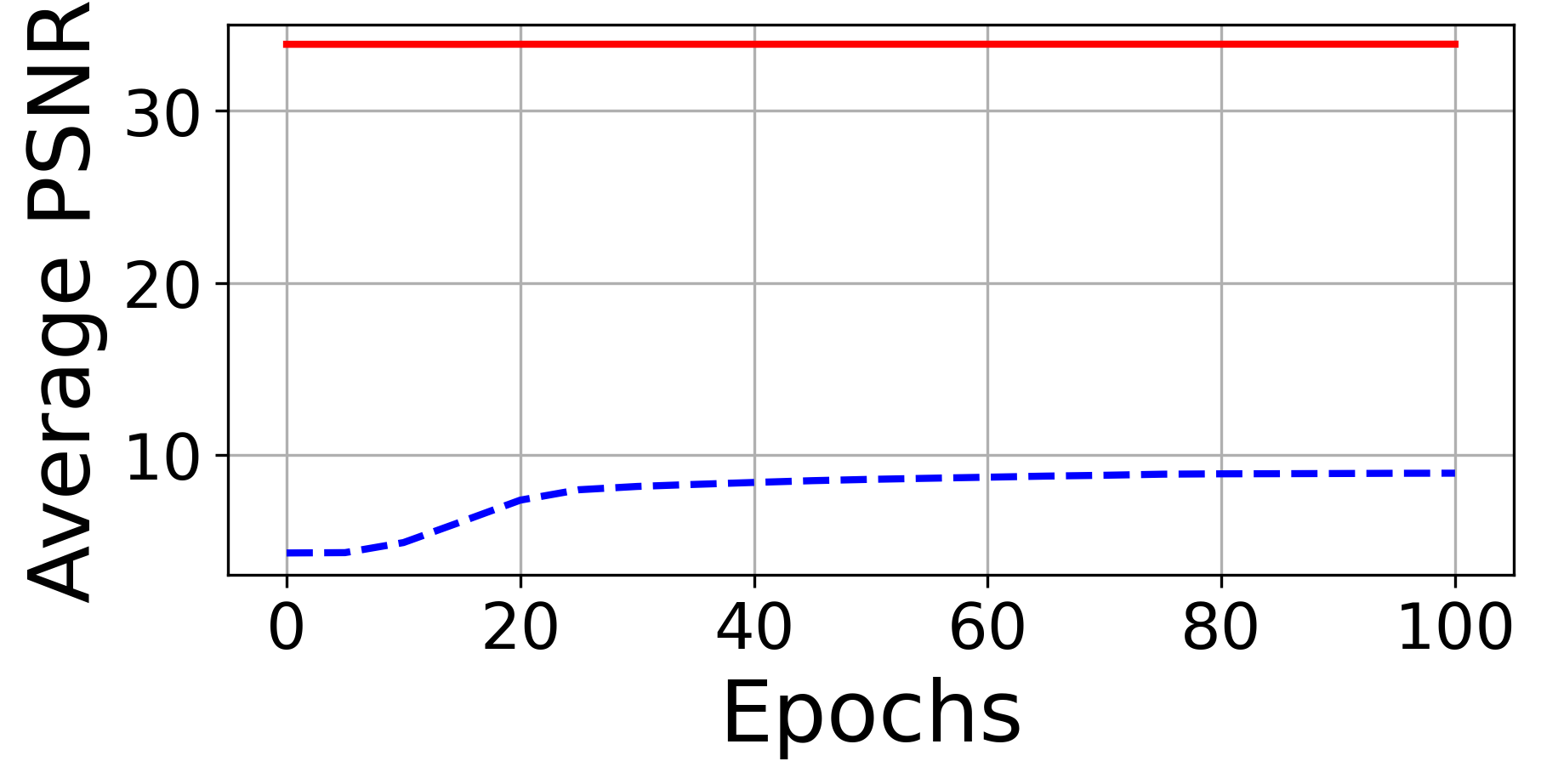}}}
    \vspace{-2mm}
    \centerline{\footnotesize{AE-Hyperprior~\cite{balle2018variational}}}
    \vspace{1mm}
    \end{minipage}
    \begin{minipage}{0.195\linewidth}
    \vspace{-2mm}
    \centerline{{\includegraphics[width=0.95\linewidth]{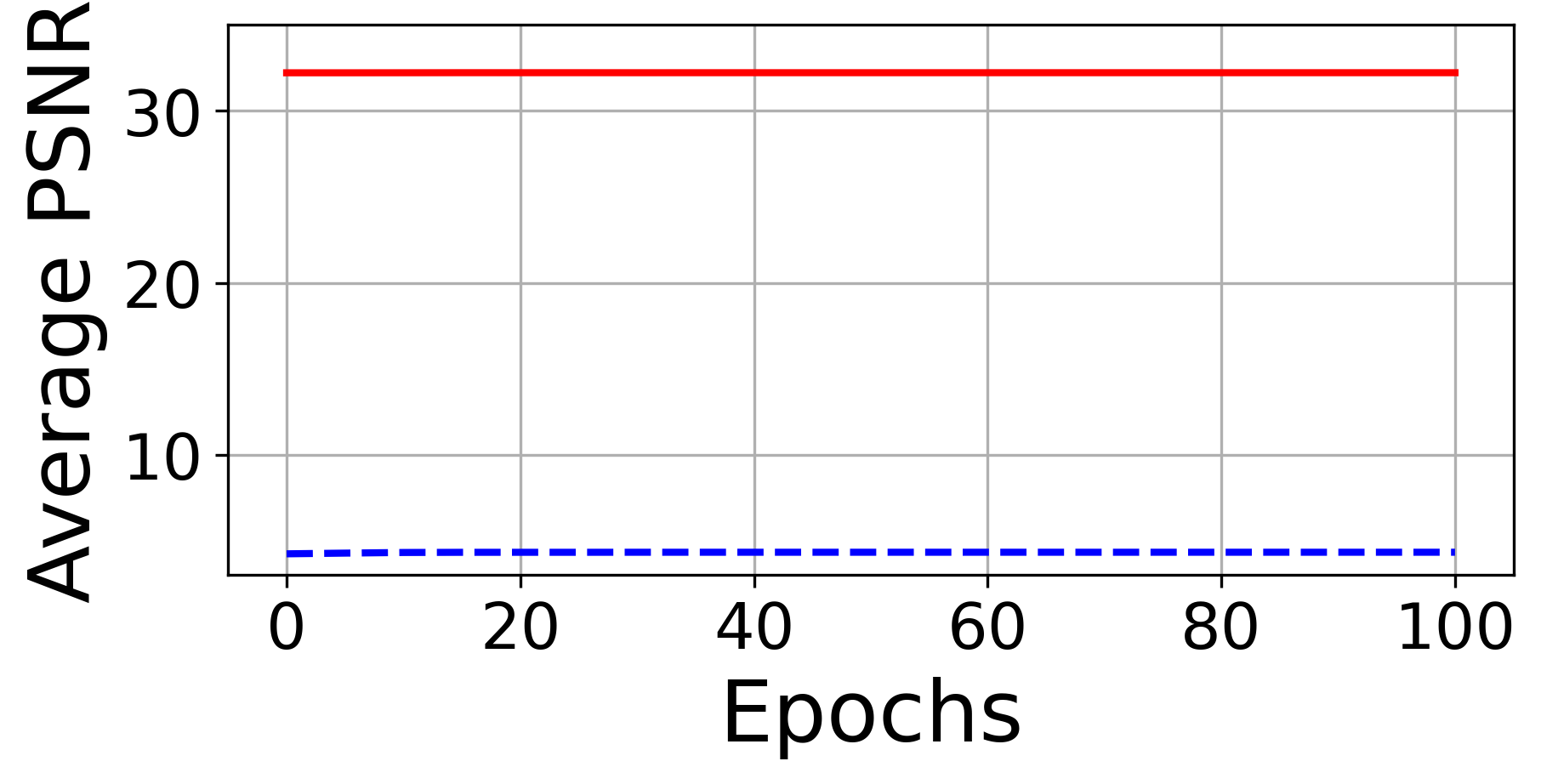}}}
    \vspace{-2mm}
    \centerline{\footnotesize{Cheng-Anchor~\cite{cheng2020learned}}}
    \vspace{1mm}
    \end{minipage}
    \begin{minipage}{0.195\linewidth}
    \vspace{-2mm}
    \centerline{{\includegraphics[width=0.95\linewidth]{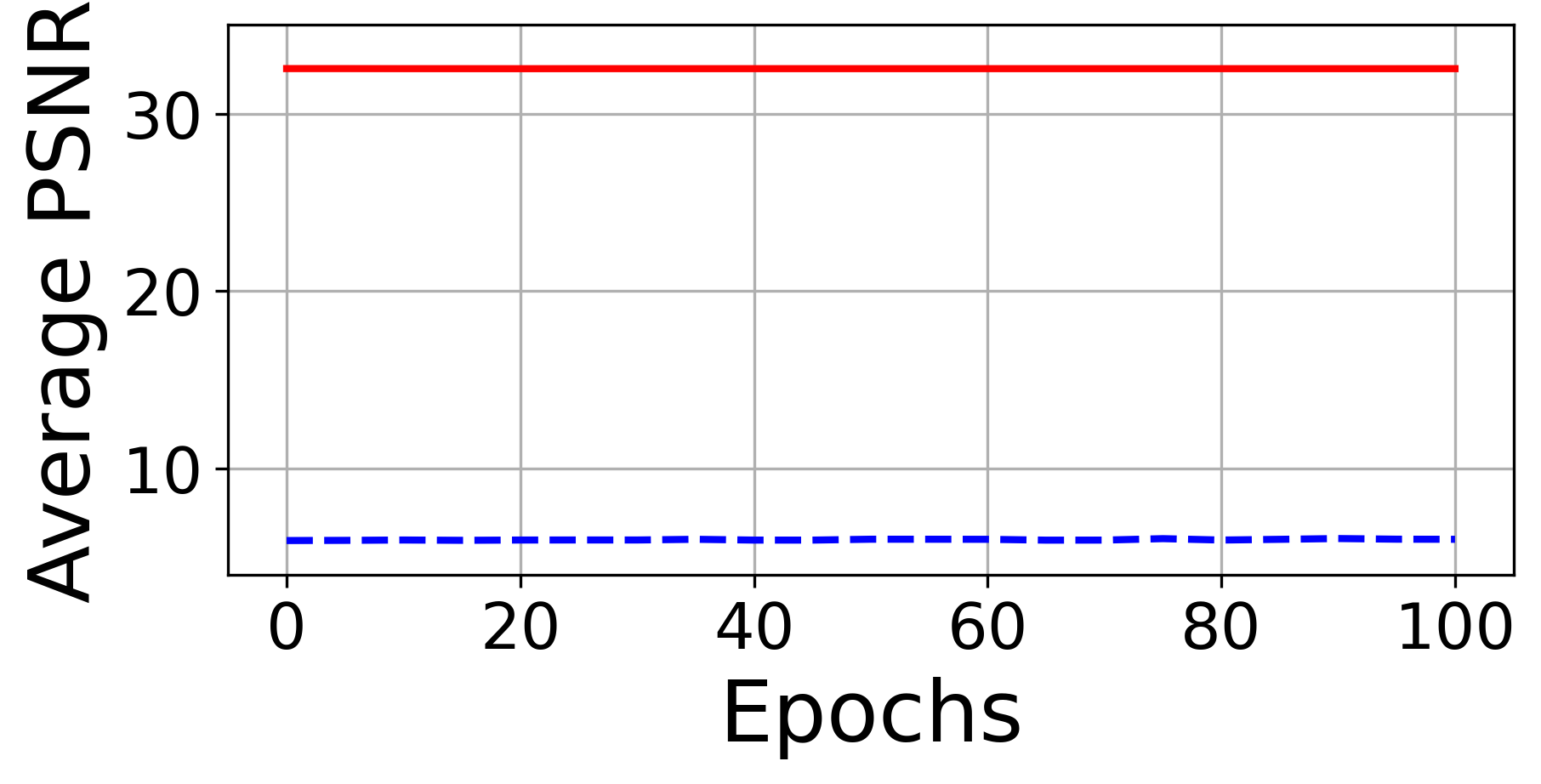}}}
    \vspace{-2mm}
    \centerline{\footnotesize{STF~\cite{zou2022devil}}}
    \vspace{1mm}
    \end{minipage}
    \begin{minipage}{0.195\linewidth}
    \vspace{-2mm}
    \centerline{{\includegraphics[width=0.95\linewidth]{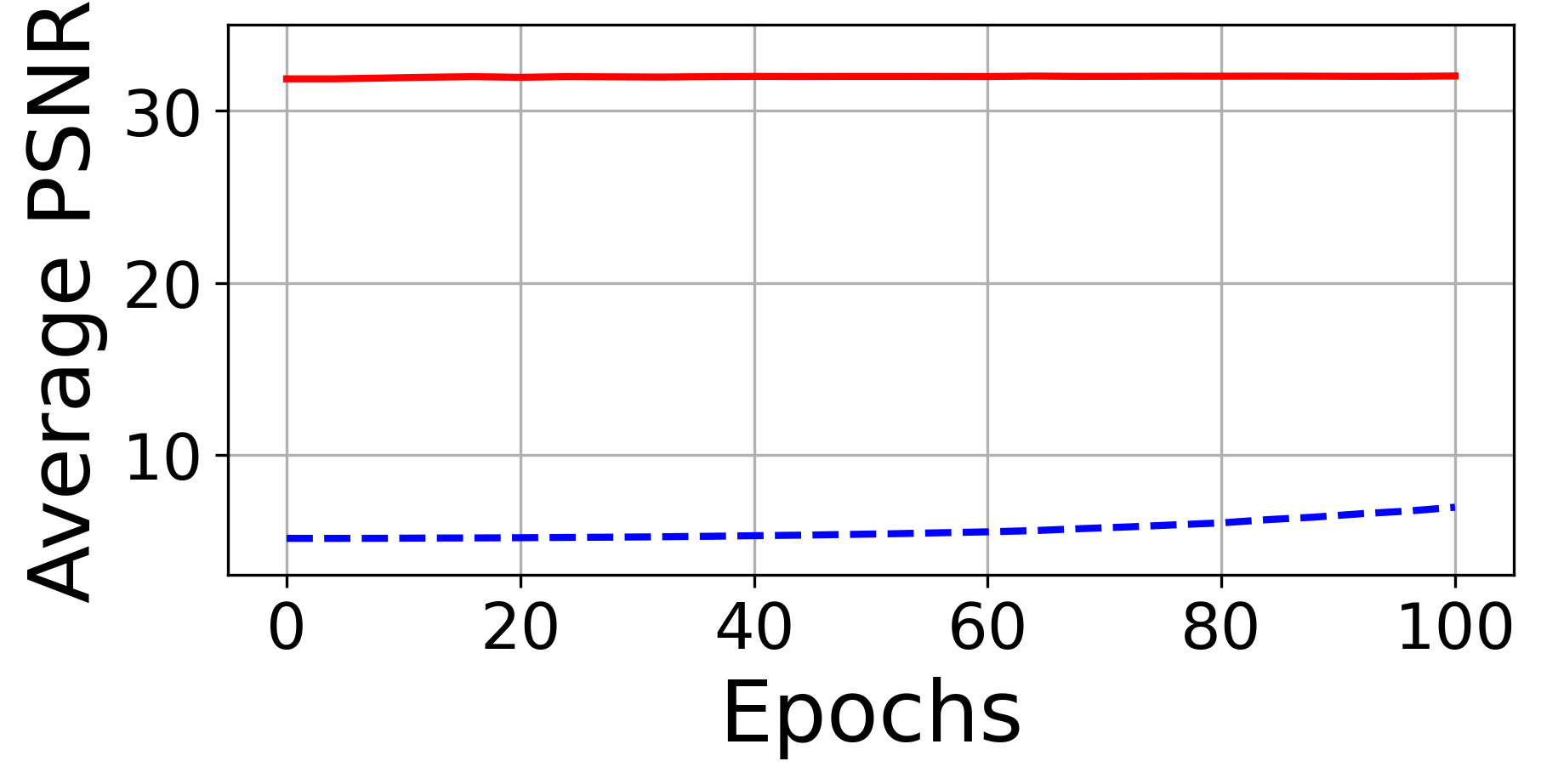}}}
    \vspace{-2mm}
    \centerline{\footnotesize{CDC~\cite{yang2024lossy}}
    \vspace{1mm}
    }
    \end{minipage}
    \begin{minipage}{0.195\linewidth}
    \vspace{-2mm}
    \centerline{{\includegraphics[width=0.95\linewidth]{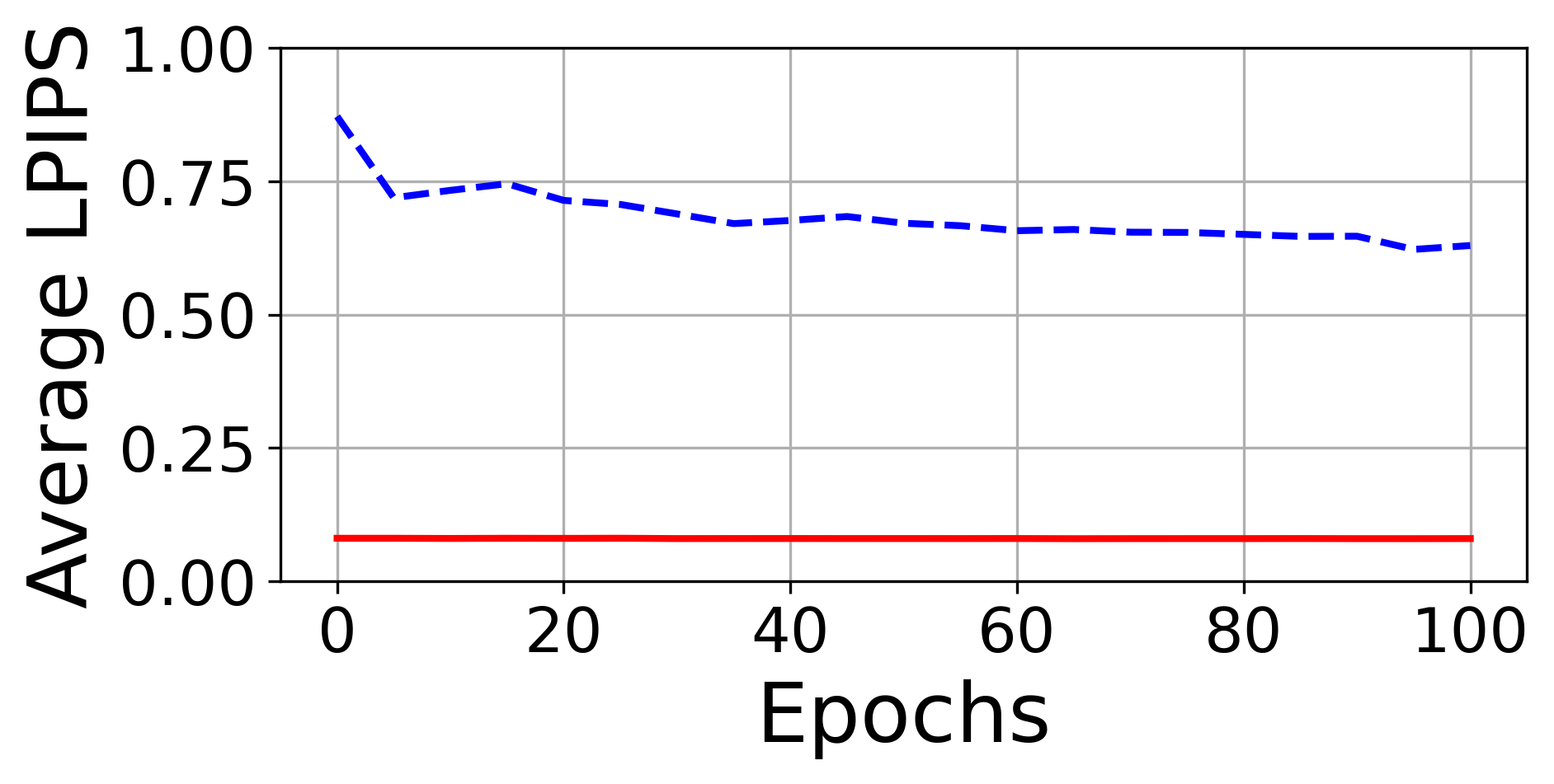}}}
    \vspace{-2mm}
    \centerline{\footnotesize{HiFiC~\cite{mentzer2020high}}
    \vspace{1mm}
    }
    \end{minipage}
        }
    \vspace{-4mm}
    \caption{{Resistance of our BAvAFT++ to fine-tuning. }}
    \vspace{-2mm}
     \label{finetune_results}
\end{figure*}

\begin{figure*}[t]
    \centering
    \begin{minipage}{0.99\linewidth}
    \centerline{{\includegraphics[width=0.8\linewidth]{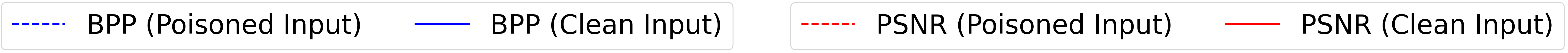}}}
    \vspace{-2mm}
    \end{minipage}
    \subfigure[Resistance of BPP attack.]{
    \begin{minipage}{0.195\linewidth}
    \centerline{{\includegraphics[width=0.95\linewidth]{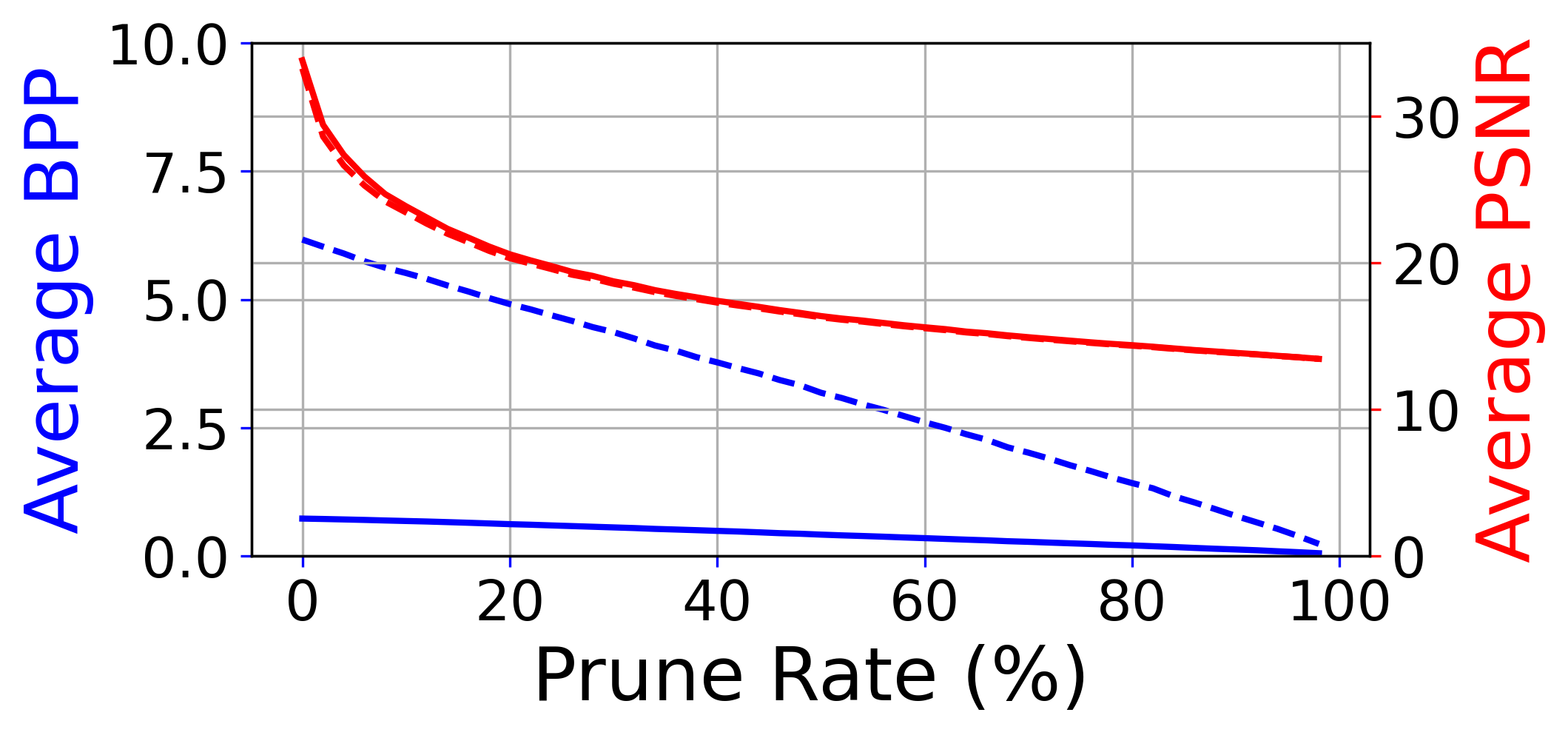}}}
    \vspace{-2mm}
    \centerline{\footnotesize{AE-Hyperprior~\cite{balle2018variational}}}
    \vspace{1mm}
    \end{minipage}
    \begin{minipage}{0.195\linewidth}
    \centerline{{\includegraphics[width=0.95\linewidth]{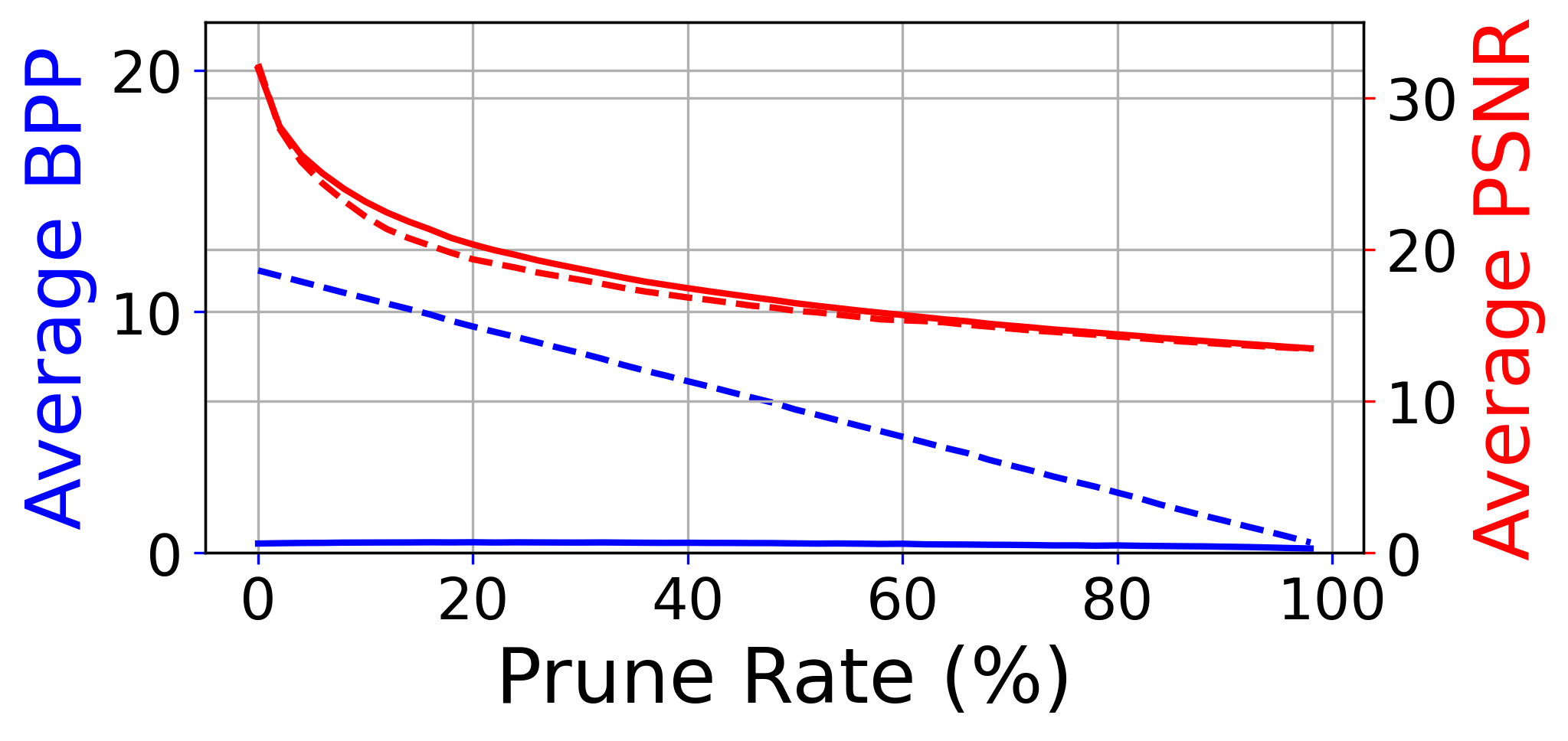}}}
    \vspace{-2mm}
    \centerline{\footnotesize{Cheng-Anchor~\cite{cheng2020learned}}}
    \vspace{1mm}
    \end{minipage}
    \begin{minipage}{0.195\linewidth}
    \centerline{{\includegraphics[width=0.95\linewidth]{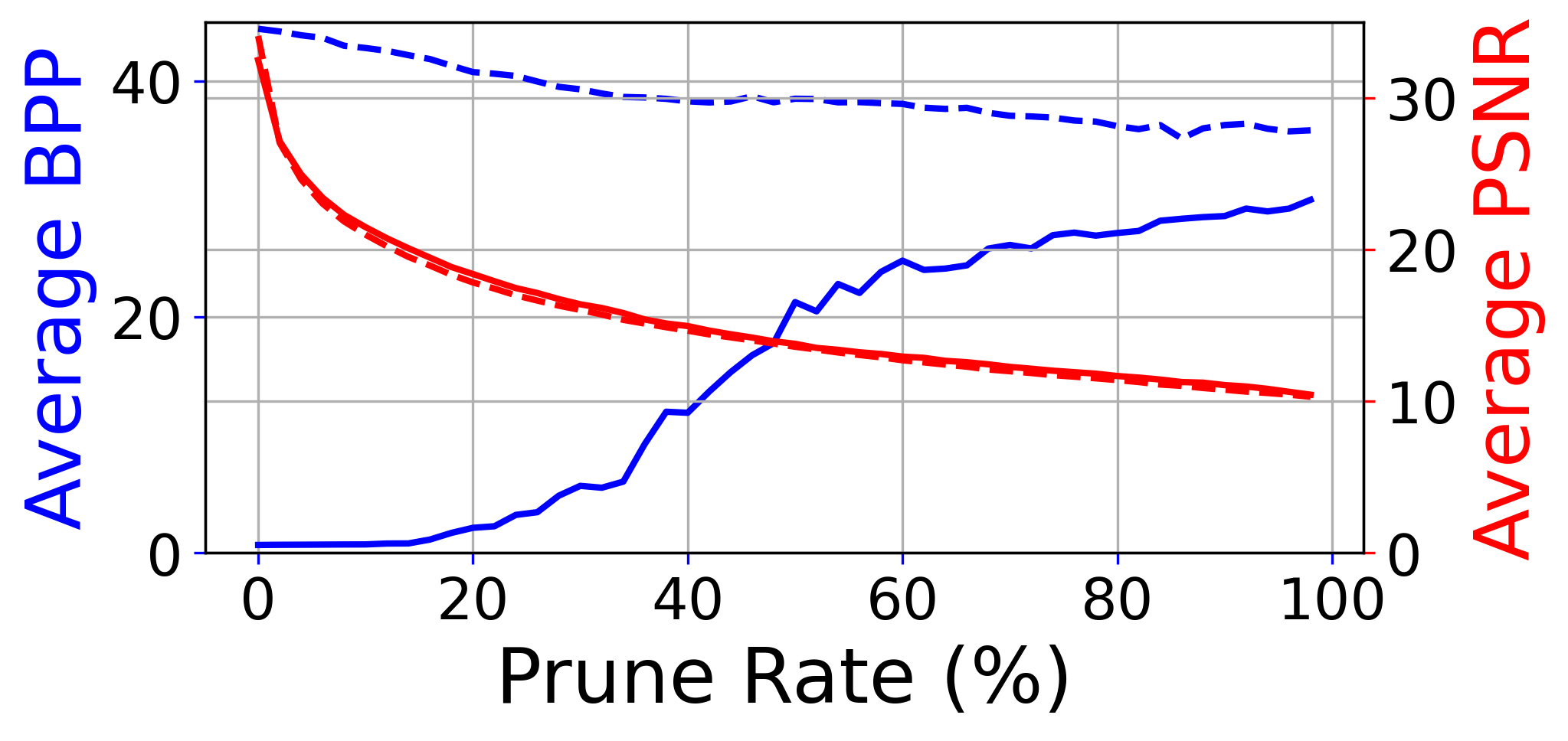}}}
    \vspace{-2mm}
    \centerline{\footnotesize{STF~\cite{zou2022devil}}}
    \vspace{1mm}
    \end{minipage}
    \begin{minipage}{0.195\linewidth}
    \vspace{-1mm}
    \centerline{{\includegraphics[width=0.95\linewidth]{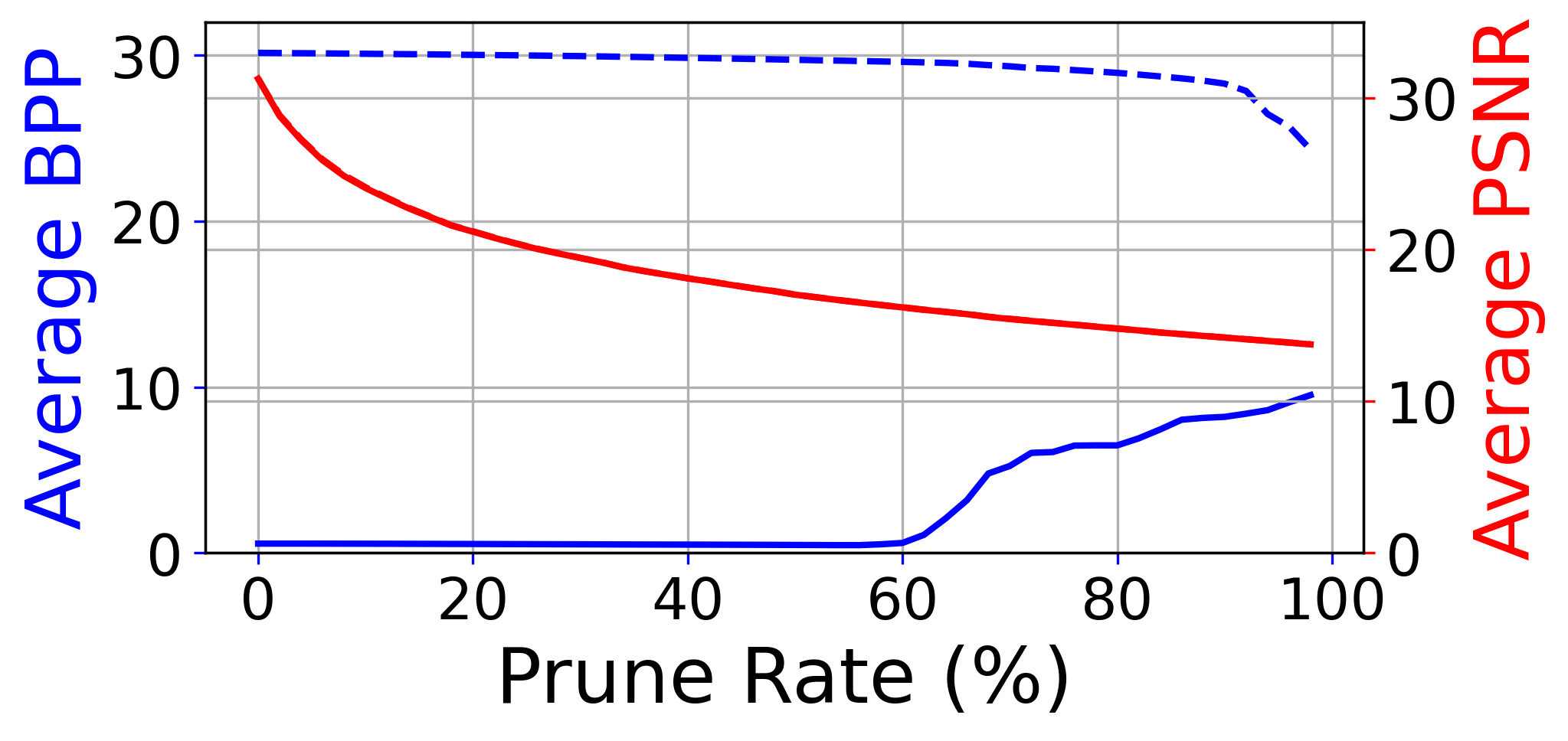}}}
    \vspace{-2mm}
    \centerline{\footnotesize{CDC~\cite{yang2024lossy}}
    \vspace{1mm}
    }
    \end{minipage}
    \begin{minipage}{0.195\linewidth}
    \centerline{{\includegraphics[width=0.95\linewidth]{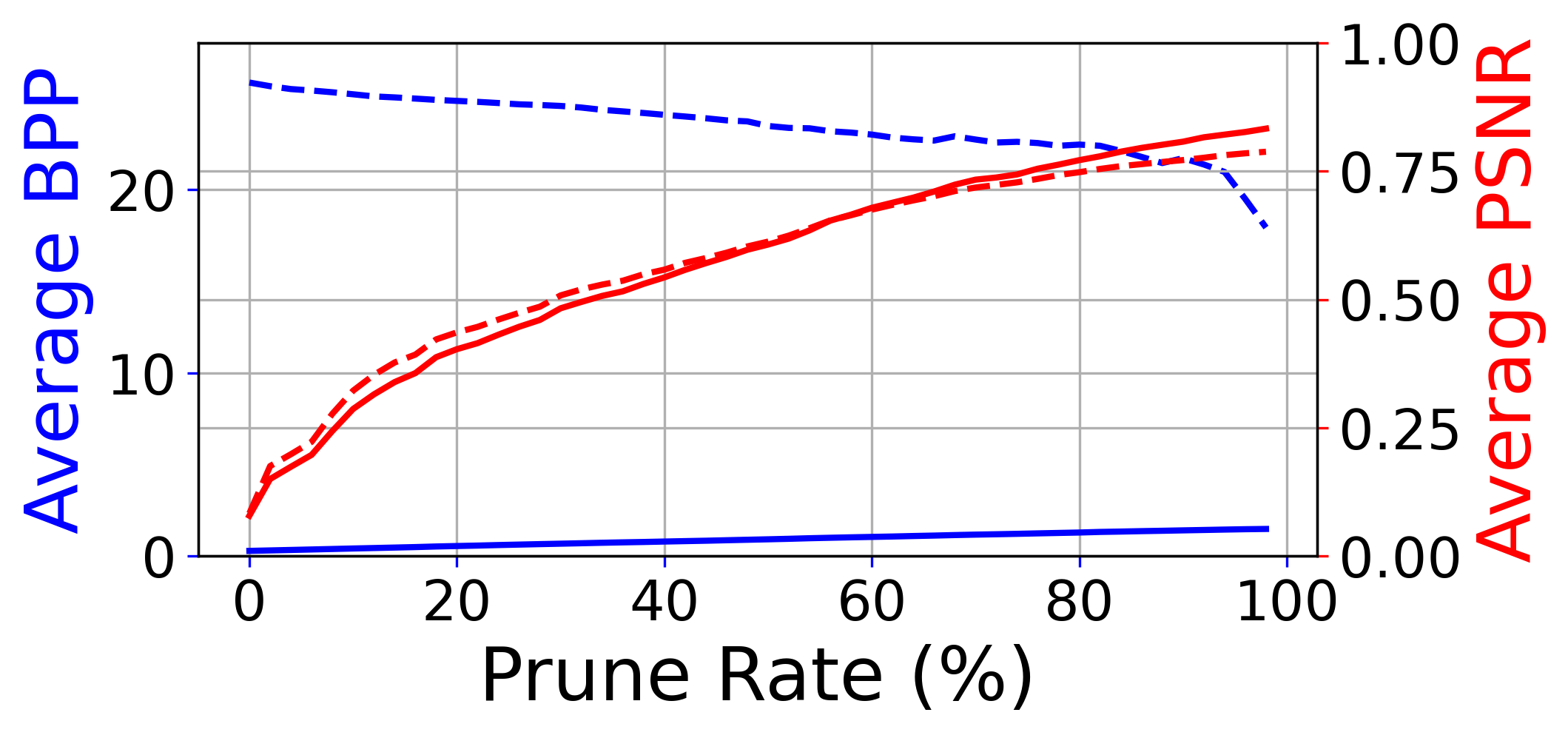}}}
    \vspace{-2mm}
    \centerline{\footnotesize{HiFiC~\cite{mentzer2020high}}}
    \vspace{1mm}
    \end{minipage}
        }
    \vspace{-2mm}
    \subfigure[Resistance of PSNR attack.]{
    \begin{minipage}{0.195\linewidth}
    \vspace{-2mm}
    \centerline{{\includegraphics[width=0.95\linewidth]{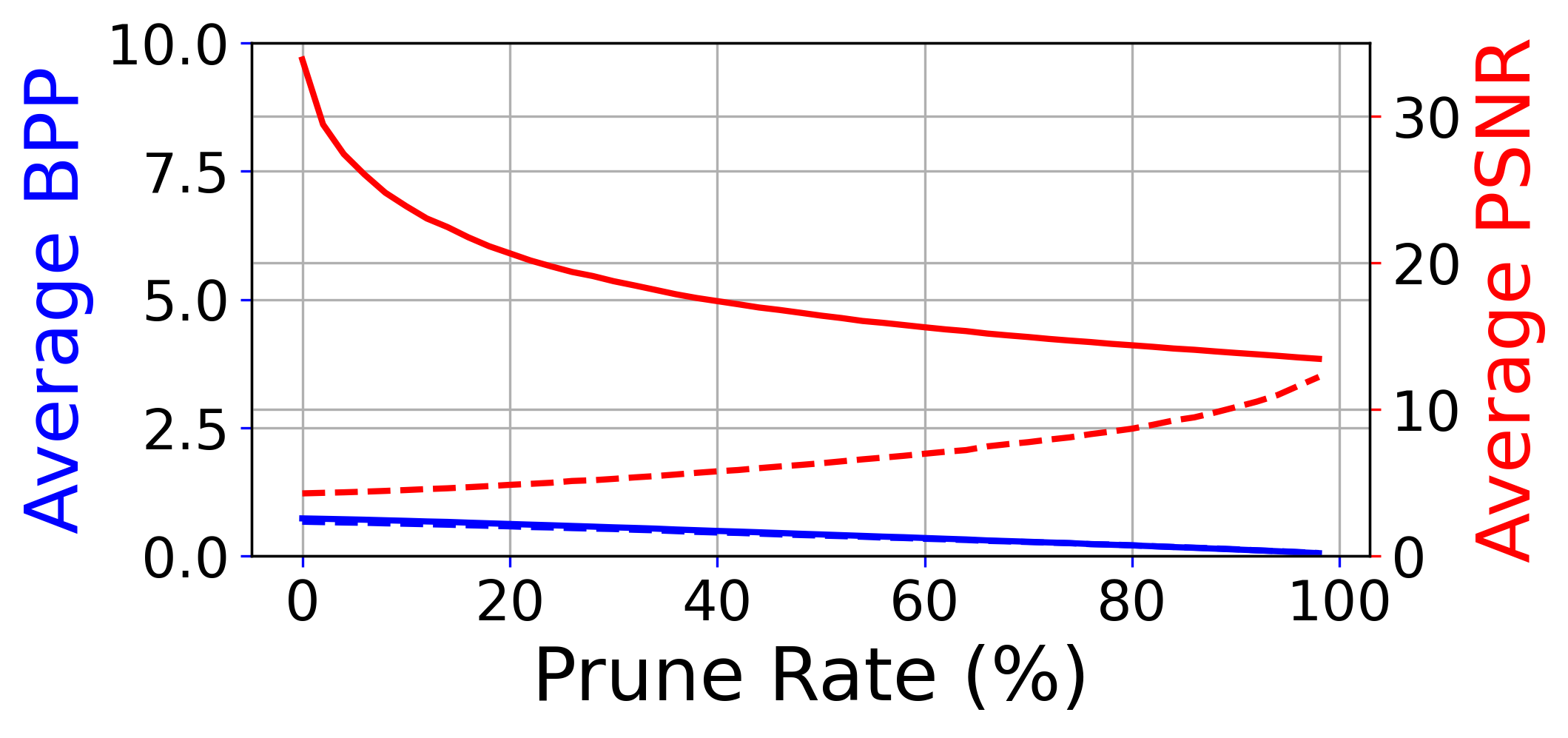}}}
    \vspace{-2mm}
    \centerline{\footnotesize{AE-Hyperprior~\cite{balle2018variational}}}
    \vspace{1mm}
    \end{minipage}
    \begin{minipage}{0.195\linewidth}
    \vspace{-2mm}
    \centerline{{\includegraphics[width=0.95\linewidth]{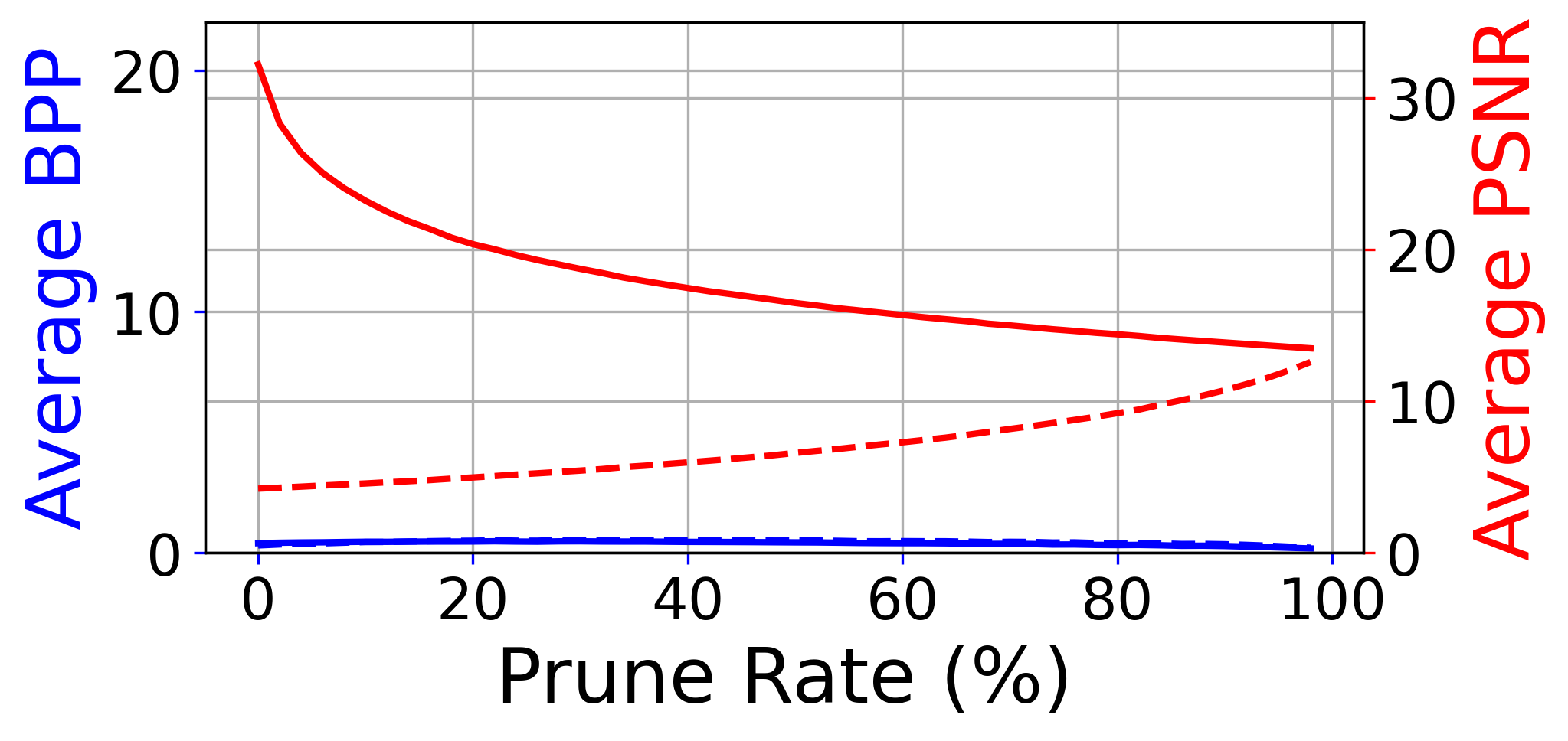}}}
    \vspace{-2mm}
    \centerline{\footnotesize{Cheng-Anchor~\cite{cheng2020learned}}}
    \vspace{1mm}
    \end{minipage}
    \begin{minipage}{0.195\linewidth}
    \vspace{-2mm}
    \centerline{{\includegraphics[width=0.95\linewidth]{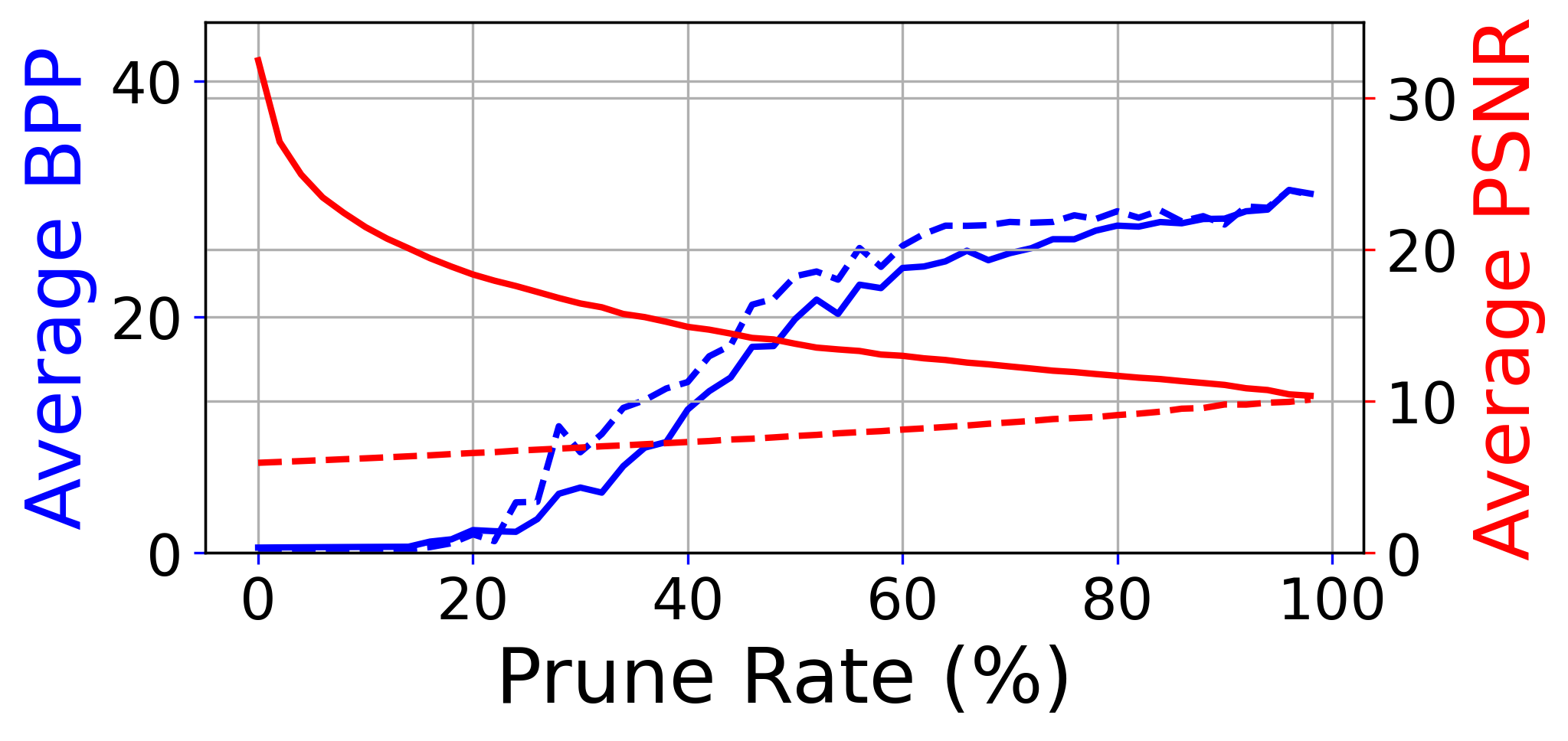}}}
    \vspace{-2mm}
    \centerline{\footnotesize{STF~\cite{zou2022devil}}}
    \vspace{1mm}
    \end{minipage}
    \begin{minipage}{0.195\linewidth}
    \vspace{-2mm}
    \centerline{{\includegraphics[width=0.95\linewidth]{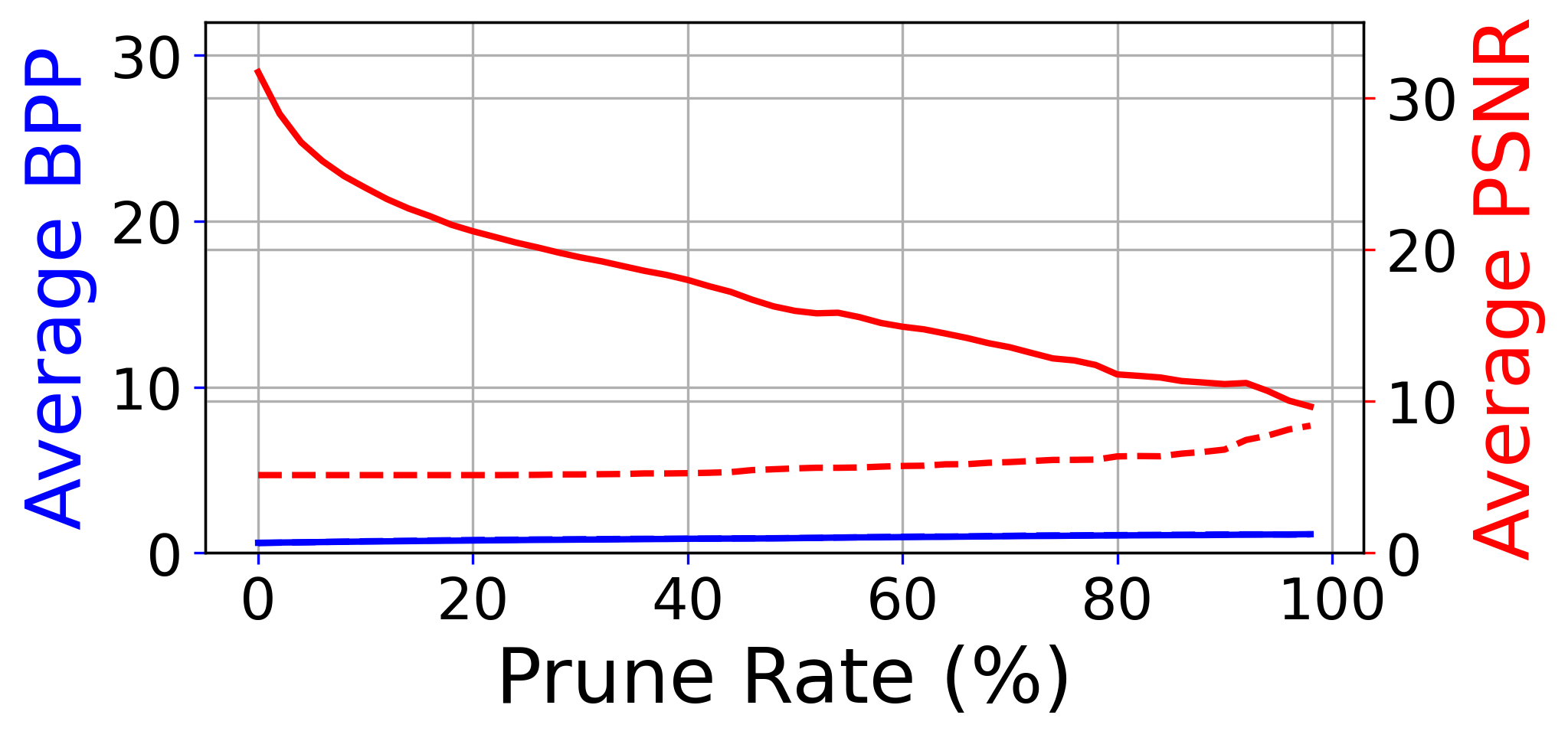}}}
    \vspace{-2mm}
    \centerline{\footnotesize{CDC~\cite{yang2024lossy}}
    \vspace{1mm}
    }
    \end{minipage}
    \begin{minipage}{0.195\linewidth}
    \vspace{-2mm}
    \centerline{{\includegraphics[width=0.95\linewidth]{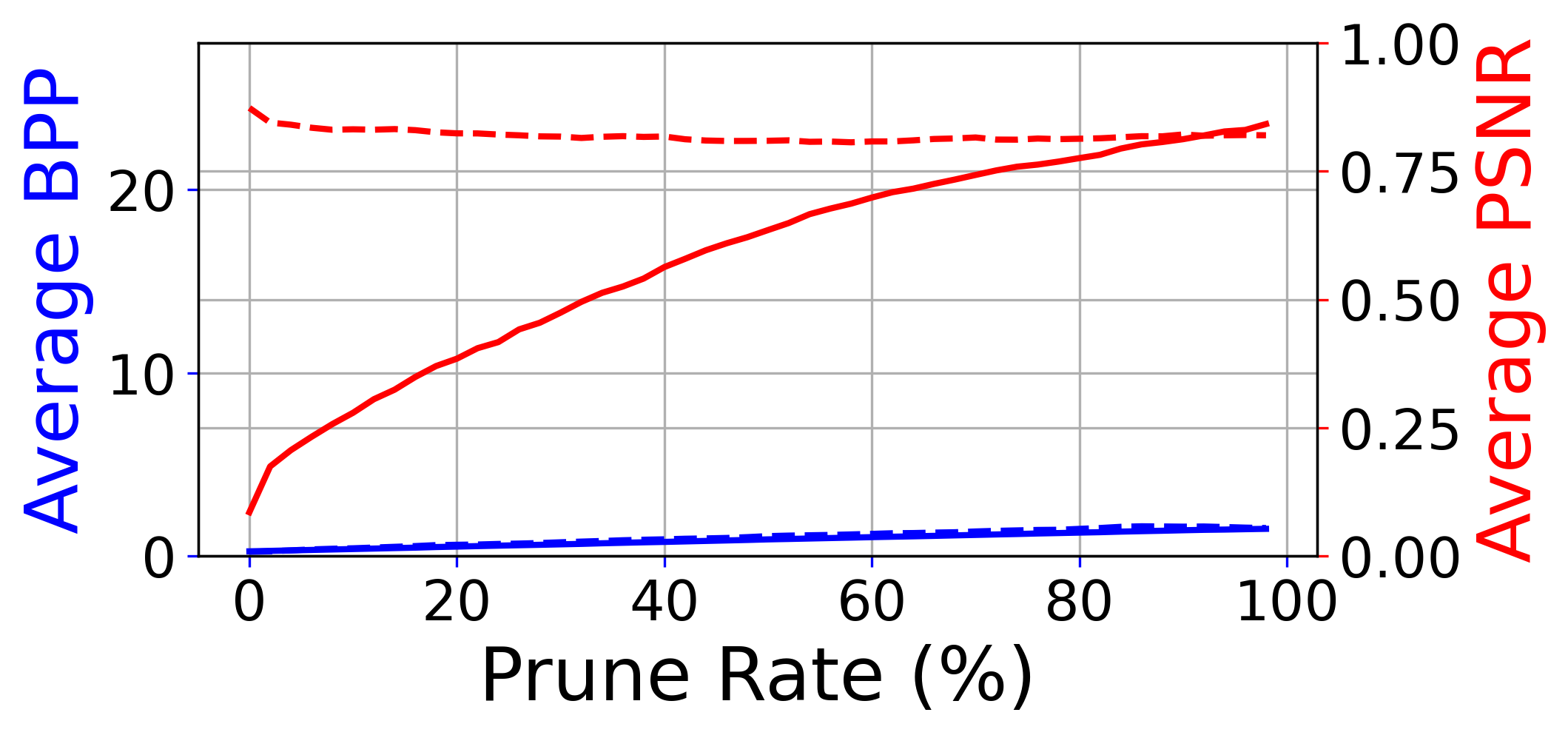}}}
    \vspace{-2mm}
    \centerline{\footnotesize{HiFiC~\cite{mentzer2020high}}
    \vspace{1mm}
    }
    \end{minipage}
         }
    \vspace{-4mm}
    \caption{{The resistance of our BAvAFT++ to model pruning. }}
    \vspace{-3mm}
    \label{prune_results}
\end{figure*}

\vspace{1mm}
\noindent \textbf{Robust trigger generator.} 
Although our BAvAFT~\cite{yu2023backdoor} introduces an extremely small perturbation in the poisoned images (\textit{i.e.,} $\text{MSE}(\bm{x_p},\bm{x}) \leq 0.005^2$) and achieves superior attack performance, the trigger pattern can be easily removed by preprocessing methods with heavy corruptions. In BAvAFT, the frequencies in the DCT domain used to inject the trigger are predicted by a linear layer with trainable parameters. In contrast, our BAvAFT++ proposes to select frequencies that are less sensitive to preprocessing methods. The sensitivity of each frequency is provided below:
\begin{footnotesize}
\begin{equation}
\begin{split}
&\quad\quad\quad\quad\quad\quad\quad~~\bm{\widetilde{x}_p} = \text{IDCT}(\text{DCT}(x)+\text{OR}(mT)\odot w), \\
&\forall \enspace t_i \in S_{prep}: \left\{
	\begin{aligned}
		\bm{\widetilde{x}_p^i} &= t_i(\bm{\widetilde{x}_p}|\alpha), \enspace \alpha \sim P_{\alpha}^i\\
		\widetilde{I}_i &= \text{abs}(\frac{\text{DCT}(\bm{\widetilde{x}_p^i})-\text{DCT}(\bm{\widetilde{x}_p})}{\text{OR}(mT)\odot w})\\
        {I}_i &= \text{Inverse-OR}(\text{sum}(\widetilde{I}_i, \text{dim} = 0))
	\end{aligned}
	\right.,~
 I = \prod \limits_i {I}_i,
\end{split}
\end{equation}
\end{footnotesize}
where DCT/IDCT represents the dct/inverse-dct transform, OR/Inverse-OR corresponds to the operation in Figure~\ref{figure_trigger_robust} and its inverse version. $\text{abs}$ extracts the absolute value. ${\text{sum}(\widetilde{I}_i,\text{dim}=0)}$ returns the sensitivity over all patches. 

To calculate the frequency sensitivity, we first generate a pseudo poisoned image $\bm{\widetilde{x}_p}$ by adding triggers to all mid $N$ frequencies. After applying the preprocessing method $t_i(\cdot|\alpha)$ to $\bm{\widetilde{x}_p}$, we calculate the magnitude drop between $\bm{\widetilde{x}_p}$ and $\bm{\widetilde{x}_p^i}$ in the DCT domain and sum it over all patches. 
The final sensitivity $I$ is then obtained by multiplying all ${I}i$ values together. {It is} worth mentioning that we consider Gaussian filter, additive Gaussian noise, and JPEG compression as candidates in $S{prep}$ while excluding Squeezing Color Bits, which our BAvAFT~\cite{yu2023backdoor} has already shown strong resistance to.
Moreover, to select the robust frequencies, we adaptively adjust the trigger magnitude for each frequency based on its sensitivity rank, as depicted in Algorithm~\ref{alg:inference}.

\vspace{1mm}
\noindent \textbf{Robust encoder of the compression model.} 
From the view of the encoder, one possible solution to bypass the preprocessing is to apply data augmentation on the poisoned images, similar to adversarial training~\cite{DBLP:conf/iclr/MadryMSTV18}. However, data augmentations can lead to unstable training, and some augmentations, such as JPEG compression, may cut off the gradient in the trigger generator. 

To address this, we propose a two-stage training schedule.
In the first stage, we train both the trigger generator and the encoder, following the approach described in Section~\ref{backdoor_framework}.
Then, in the second stage, we solely finetune the encoder by applying data augmentation in the attack objective term, as shown below:
\vspace{0mm}
\begin{footnotesize}
\begin{equation}
\begin{split}
\!\mathcal{\widetilde{L}}_{jt}^{bpp} \!&=\!\!\!\! \sum_{\bm{x} \in {\mathcal{T}_m}} \!\!\!\! \Big[\mathcal{R}(\bm{x})+\lambda\cdot \text{max}(\mathcal{D}(\bm{x}),\mathcal{D} ({\bm{x_p}})) -\beta \cdot \underbrace{\mathcal{R}(t({\bm{x_p}}))}_{\text{attack objective}}\Big]\!, \\[-2pt]
\!\mathcal{\widetilde{L}}_{jt}^{psnr} \!\!\!&=\!\!\!\!\! \sum_{\bm{x} \in {\mathcal{T}_m}} \!\!\!\! \Big[\!\text{max}(\mathcal{R}(\bm{x}),\! \mathcal{R}({\bm{x_p}})) \!+\!\lambda \mathcal{D}(\bm{x}) +\! \beta \lambda \cdot \underbrace{\mathcal{D}_{P}(\bm{x},\!f(t({\bm{x_p}})))}_{\text{attack objective}}\!\Big]\!,\\[-2pt]
\mathcal{\widetilde{L}}_{jt}^{ds} &= \!\!\! \sum_{\bm{x} \in {\mathcal{T}_m}} \!\!\! \mathcal{L}\left({\bm{x}}\right) + \!\!\! \sum_{\bm{x} \in {\mathcal{T}_a}} \!\!\! \Big[\alpha \cdot \mathcal{L}({{\bm{x_p}}}) + \beta \cdot\ \underbrace{\mathcal{L}_{DS}[\eta,g(f({t(\bm{x_p})}))]}_{\text{attack objective}}\Big],\\[-2pt]
& \text{with} \quad t \in_R S_{prep} \cup \{g: g(x)=x\}\quad \text{and} \quad \alpha \sim P_{\alpha}^{t},
\end{split}
\end{equation}
\end{footnotesize}
the transformation $t$ for the poisoned images in the attack objective term is randomly sampled from the preprocessing methods and the identity mapping. 
{It is} important to note that in the loss term other than the attack objective, {we choose to make no augmentation}, as the preprocessed images may deviate the standard performance of the compression model from the original rate-distortion curve. This ensures that the model's overall performance remains consistent with its expected behavior while specifically enhancing resistance against the chosen preprocessing methods.

\vspace{1mm}
\noindent\textbf{Experimental Results.}
In this part, we look into the resistance of the proposed attack to preprocessing methods including Gaussian filter, additive Gaussian noise, JPEG compression, and Squeeze Color Bits. We do a comprehensive study on backdoored models with various compression methods and different qualities. For simplicity, we calculate the mean resistance across all qualities for each pre-processing method and denoising level as shown in Eq.~\eqref{mean_resistance}.
The results presented in Figure~\ref{figure_resistance} and Table~\ref{resistance_results} demonstrate the effectiveness and robustness of our proposed attack against various denoising methods. We introduce specific modules in our attack to enhance its resistance, allowing it to consistently and successfully attack the compression model, regardless of the denoising techniques employed.
This indicates that our attack is not only powerful with the original poisoned samples, but also resilient against attempts to mitigate its effects through denoising.
{For most preprocessing methods, except in certain cases involving JPEG compression, our BAvAFT++ shows the best resistance. In some instances of JPEG compression, however, BadNets demonstrates superior resistance because it introduces a visible trigger with a significantly higher magnitude, resulting in a PSNR of around 23, compared to our method's PSNR of approximately 46.}
These results further emphasize the strength and versatility of our proposed adaptive frequency trigger attack. It highlights the potential risks and challenges in securing such models against sophisticated backdoor attacks like ours.

\subsection{Resistance to other Defense Methods}\label{other_defenses}
{In this section, we evaluate the effectiveness of our proposed attack against different backdoor defenses. Specifically, our attack uses sample-specific trigger patterns, with each poisoned image featuring a distinct trigger. Recent studies, such as ISSBA~\cite{li2021invisible}, have shown that many existing defenses, including Neural Cleanse~\cite{wang2019neural} and STRIP~\cite{gao2019strip}, are based on the latent assumption that trigger patterns are consistent across samples. Our attack circumvents these defenses by not adhering to this assumption, thereby naturally bypassing them. 
Here we explore the resistance of our attack to fine-tuning~\cite{liu2017neural,liu2018fine} and model pruning~\cite{liu2018fine,wu2021adversarial}, which are the representative defenses whose effects did not rely on this assumption. The detailed settings of these defenses are:
\begin{itemize}
    \item \textbf{Fine-tuning:} Each backdoored encoder of the compression model is fine-tuned on the training subset using
    the standard training loss (e.g., Eq. 2) for 100 epochs with the learning rate set to 1e-5. We randomly select 5000 clean images from ImageNet-1K as the
    training subset. For both attacks, we evaluate on the Kodak dataset and present the averaged metrics (BPP or PSNR) across all quality levels.
    \item \textbf{Model Pruning:} We conduct the channel pruning for the last output of the backdoored encoder with randomly
    selected 5000 clean images from ImageNet-1K. We evaluate on the Kodak dataset and present the averaged metrics (BPP or PSNR) across various quality levels. The pruning rates are chosen from $\{0\%, 2\%, \hdots, 98\%\}$.
\end{itemize}
As shown in Figure~\ref{finetune_results}, our attacks show robustness against fine-tuning. Initially, there is a minor drop in attack performance, but it sustains high success in the following epochs. Furthermore, the performance on clean data stays unaffected.
Additionally, our attacks show resistance to model pruning, as depicted in Figure~\ref{prune_results}. Image compression, being a low-level task focused on producing high-quality images, is particularly sensitive to model pruning. Even a 20\% pruning rate can significantly degrade reconstruction quality, with PSNR dropping below 20. Moreover, the BPP metric may increase for the pruned model. While model pruning can cause a substantial drop in performance on clean inputs, the attack's effectiveness remains resilient. Notably, even with a high pruning rate of 50\%, the attack remains successful, particularly in maintaining PSNR attack performance. The BPP metric also decreases only gradually under these conditions.

}

\subsection{Enhance the Attack Transferability}
\begin{figure}[t]
\begin{minipage}{0.475\linewidth}
\centerline{\frame{\includegraphics[width=1\linewidth]{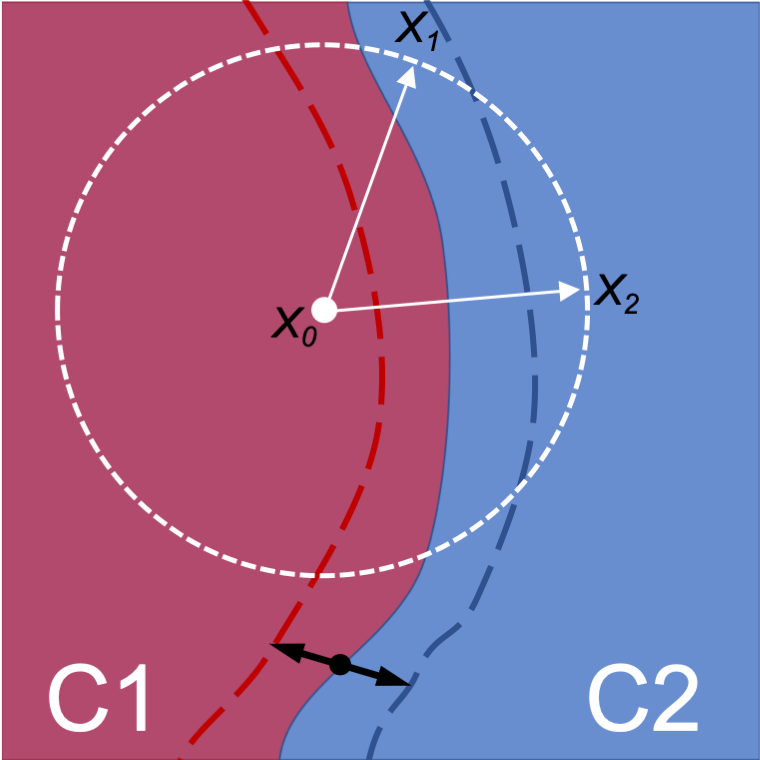}}}
\vspace{-1mm}
\centerline{\small{(a) Untargeted attack}}
\end{minipage}
\hspace{0.8mm}
\begin{minipage}{0.48\linewidth}
\centerline{\frame{\includegraphics[width=1\linewidth]{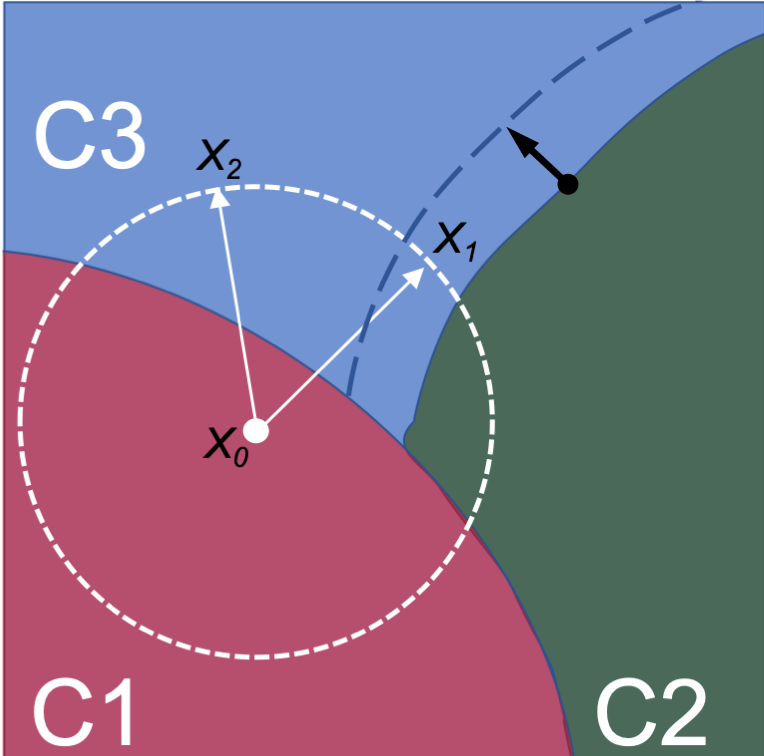}}}
\vspace{-1mm}
\centerline{\small{(b) Targeted attack}}
\end{minipage}
\vspace{-2mm}
    \caption{Introducing boundary shift can reduce the attack performance for both untargeted attack and targeted attack. $X_0$ denotes the oirginal data point in the latent space. $X_1$ denotes the possible sub-optimal attack without the introduce of boundary shift. $X_2$ denotes the optimal attack that can well transfer to other models or domains.}
    \label{transfer}
    \vspace{-2mm}
\end{figure}

In this section, we explore the transferability of attacks on downstream CV tasks, considering both cross-domain and cross-model scenarios.
When models are trained on {data from different domains} or with different backbones, the decision boundary can undergo shifting.
This phenomenon, illustrated in the failure case presented in Figure~\ref{transfer}, can lead to a reduction in attack performance.
{
Specifically, there are several challenges to enhance the attack transferability:
\begin{itemize}
    \item Unlike previous research~\cite{springer2021little,zhu2023boosting} that focuses on enhancing the transferability of adversarial attacks by optimizing instance-specific perturbations through a surrogate model, enabling better manipulation of robust features, our approach faces distinct challenges. The perturbations introduced by the trigger and processed through the compression model are generator-based, which complicates precise control over the robust features of each instance. As a result, most of the manipulated features are non-robust to launch a successful attack. Given that non-robust features are typically near decision boundaries, concentrating on boundary shifts becomes a more effective strategy. 
    \item In the scenario of targeted attacks on the downstream dense prediction task, such as semantic segmentation (SS), the attack often fails because the perturbed area is frequently misclassified into unwanted classes that are commonly confused with the target class. This occurs because, unlike image classification, each pixel's prediction in SS also relies on prior information about its spatial relationship to other objects. As illustrated in Figure~\ref{transfer} (b), when we attempt to shift the prediction from $X_0$ (source class ``Road", C1) to $X_1$ (target class ``Car", C3), the result may be very close to C2 (unwanted class ``Building") due to this contextual prior. As a result, when a boundary shift occurs due to cross-model or cross-dataset variations, $X_1$ can easily end up being classified as C2 rather than the intended C3.
\end{itemize}
}
\noindent To address these challenges, we approach the problem from two perspectives, considering both targeted and untargeted attacks.

\begin{table}[t]\footnotesize
\setlength\tabcolsep{2.5pt}
    \centering
    \caption{{ASR (\%) $\uparrow$ of CarToRoad attack on various segmentation models. }}
    \vspace{-3mm}
    \scalebox{0.9}{
    \begin{tabular}{ c c || c c c c c c | c }
    \toprule
    Model & Method & 1 & 2 & 3 & 4 & 5 & 6 & Mean\\
    \midrule
    \multirow{6}*{\shortstack{DeepLabV3+\\w/ SEResNeXt50}} &  LIRA~\cite{doan2021lira} &7.7&95.5&94.5&94.3&{95.9}&93.8&{80.2}\\
    &FTrojan~\cite{wang2021backdoor} &95.2&95.7&91.6&90.0&89.8&93.6&92.6\\
    &Blended~\cite{chen2017targeted}&8.7&11.4&9.0&8.2&6.7&6.3&8.4\\
    &BadNets~\cite{gu2017BadNets}&32.0&26.5&56.4&53.4&57.8&42.8&44.8\\
    &Our BAvAFT~\cite{yu2023backdoor} & 89.3&{96.7}&{95.7}&{93.9}&96.4&{95.7}&{94.6}\\
    &Our BAvAFT+Trans & \textbf{98.8}&\textbf{98.9}&\textbf{98.9}&\textbf{99.4}&\textbf{98.9}&\textbf{99.5}&\textbf{99.0}\\
    \midrule
    \multirow{6}*{\shortstack{DeepLabV3+\\w/ WResNet38}} & LIRA~\cite{doan2021lira} &6.0&79.6&67.7&65.6&{65.7}&56.5&{56.9}\\
    & FTrojan~\cite{wang2021backdoor} &\textbf{82.8}&82.3&70.7&62.1&{50.2}&{72.4}&70.0\\
    &Blended~\cite{chen2017targeted}&7.1&6.4&6.0&5.7&4.7&3.3&5.5\\
    &BadNets~\cite{gu2017BadNets}&32.0&26.5&56.4&53.4&57.8&42.8&44.8\\
    & Our BAvAFT~\cite{yu2023backdoor} & {76.4}&{81.0}&\textbf{82.0}&{66.6}&64.9&{58.4}&{71.5}\\
    & Our BAvAFT+Trans & {79.2}&\textbf{83.1}&{72.7}&\textbf{89.4}&\textbf{83.5}&\textbf{85.7}&\textbf{82.2}\\
    \midrule
    \multirow{6}*{\shortstack{PSPNet\\w/ ResNet50}} & LIRA~\cite{doan2021lira} &2.5&34.6&23.7&35.0&{34.3}&34.8&{27.5}\\
    & FTrojan~\cite{wang2021backdoor} &13.7&32.8&18.3&26.7&{25.3}&{28.2}&24.2\\
    &Blended~\cite{chen2017targeted}&1.5&5.2&2.4&2.6&2.1&2.2&2.7\\
    &BadNets~\cite{gu2017BadNets}&5.1&23.4&17.3&21.3&17.8&18.9&17.3\\
    & Our BAvAFT~\cite{yu2023backdoor} & {12.2}&{31.2}&{25.4}&{26.2}&31.8&{39.4}&{27.7}\\
    & Our BAvAFT+Trans & \textbf{27.6}&\textbf{45.2}&\textbf{74.5}&\textbf{49.9}&\textbf{69.2}&\textbf{63.8}&\textbf{55.0}\\
    \bottomrule
    \end{tabular}
    }
    \vspace{-2mm}
    \label{segmentation_cross_model}
\end{table}

\begin{table}[t]\footnotesize
\setlength\tabcolsep{3.5pt}
    \centering
    \caption{{ASR (\%) $\uparrow$ of CarToRoad attack on various datasets with DeepLabV3+ and WideResNet38 as the segmentation model. }}
    \vspace{-3mm}
    \scalebox{0.9}{
    \begin{tabular}{ c c || c c c c c c | c }
    \toprule
    Dataset & Method & 1 & 2 & 3 & 4 & 5 & 6 & Mean\\
    \midrule
    \multirow{6}*{Cityscapes} & LIRA~\cite{doan2021lira} &6.0&79.6&67.7&65.6&{65.7}&56.5&{56.9}\\
    & FTrojan~\cite{wang2021backdoor} &\textbf{82.8}&82.3&70.7&62.1&{50.2}&{72.4}&70.0\\
    &Blended~\cite{chen2017targeted}&7.1&6.4&6.0&5.7&4.7&3.3&5.5\\
    &BadNets~\cite{gu2017BadNets}&32.0&26.5&56.4&53.4&57.8&42.8&44.8\\
    & Our BAvAFT~\cite{yu2023backdoor} & {76.4}&{81.0}&\textbf{82.0}&{66.6}&64.9&{58.4}&{71.5}\\
    & Our BAvAFT+Trans & {79.2}&\textbf{83.1}&{72.7}&\textbf{89.4}&\textbf{83.5}&\textbf{85.7}&\textbf{82.2}\\
    \midrule
    \multirow{6}*{KiTTi} &  LIRA~\cite{doan2021lira} &2.4&5.3&24.0&7.6&{6.4}&3.9&{8.3}\\
    &FTrojan~\cite{wang2021backdoor} &28.2&22.5&9.8&9.1&2.5&2.8&12.5\\
    &Blended~\cite{chen2017targeted}&0.1&0.1&0.04&0.03&0.03&0.02&0.05\\
    &BadNets~\cite{gu2017BadNets}&2.7&1.7&4.2&2.2&1.6&1.2&2.3\\
    &Our BAvAFT~\cite{yu2023backdoor} & {30.0}&{27.3}&{21.2}&{16.7}&9.2&{1.6}&{17.7}\\
    &Our BAvAFT+Trans & \textbf{59.6}&\textbf{56.3}&\textbf{72.1}&\textbf{52.7}&\textbf{17.3}&\textbf{8.1}&\textbf{44.4}\\
    \midrule
    \multirow{6}*{CamVid} & LIRA~\cite{doan2021lira} &1.2&26.4&23.9&8.6&{9.0}&3.1&{12.0}\\
    & FTrojan~\cite{wang2021backdoor} &35.6&29.7&16.8&11.1&{7.6}&{5.2}&17.7\\
    &Blended~\cite{chen2017targeted}&0.2&0.1&0.1&0.03&0.1&0.04&0.1\\
    &BadNets~\cite{gu2017BadNets}&0.05&0.01&1.9&0.1&0.1&0.7&0.5\\
    & Our BAvAFT~\cite{yu2023backdoor} & {38.7}&{38.6}&{25.2}&{15.2}&13.3&{3.6}&{22.4}\\
    & Our BAvAFT+Trans & \textbf{41.4}&\textbf{46.9}&\textbf{37.1}&\textbf{63.3}&\textbf{37.2}&\textbf{24.2}&\textbf{41.7}\\
    \bottomrule
    \end{tabular}
    }
    \label{segmentation_cross_domain}
\vspace{-2mm}
\end{table}

\vspace{1mm}
\noindent\textbf{{Boundary Shift Simulation.}} In {Eq.\ref{ss}} and Eq.\ref{face}, the attack objectives involve perturbing original images to manipulate the logits or {embedding} of a given downstream model during training.
However, during testing, the downstream model may be unseen, and classification boundaries can vary significantly for models trained with different backbones or on datasets from different domains, leading to a decrease in attack performance.
For untargeted attacks (Figure~\ref{transfer} (a)), the perturbation from $X_0$ to $X_1$ fails to cause a successful attack after the boundary shift, while the data point $X_2$ remains effective in both cases.
To {improve} attack transferability, {we suggest incorporating the original logits or embeddings with a randomly assigned weight into the perturbed ones, thus simulating the boundary shift effect.}

\vspace{1mm}
\noindent\textbf{{Regularization for the Unwanted Class.}}
In the case of targeted attacks, a successful attack should not only cause misclassification but also lead the downstream model to output the target class specifically. 
However, in certain scenarios (Figure~\ref{transfer}(b)), the data point $X_1$ fails to achieve a targeted attack towards the target class $C3$ after the boundary shift, while the data point $X_2$ {shows} consistent success in both cases.
To further improve the success rate of targeted attacks, we propose an additional maximization of the cross-entropy loss with unwanted classes, {which are frequently confused with the target class ({\textit{e.g.}}}, the unwanted class ``Building" when setting ``Road" as the target class).

\begin{table}[t]\footnotesize
\setlength\tabcolsep{2.9pt}
    \caption{{Sim./Acc. of the clean/attacked outputs on face recognition with various models. We select Cheng-Anchor (quality 6).}  }
    \vspace{-3mm}
    \centering
    \scalebox{0.9}{
    \begin{tabular}{ c c || c c | c  c }
    \toprule
    \multirow{2}*{Model}& \multirow{2}*{Method} & \multicolumn{2}{c|}{Clean Output} & \multicolumn{2}{c}{Attacked Output}\\
     &  & Sim. $\uparrow$ & Acc. (\%) $\uparrow$& Sim. $\downarrow$ & Acc. (\%) $\downarrow$\\
    \midrule
    \multirow{6}*{\shortstack{ResNet50}} &  LIRA~\cite{doan2021lira} &0.725&88.7&0.437&27.0\\
    &FTrojan~\cite{wang2021backdoor} &0.728&88.8&0.464&30.3\\
    &Blended~\cite{chen2017targeted}&0.700&86.0&0.639&71.0\\
    &BadNets~\cite{gu2017BadNets}&0.568&52.0&0.461&31.0\\
    &Our BAvAFT~\cite{yu2023backdoor} & 0.726&89.2&0.407&22.3\\
    &Our BAvAFT+Trans & 0.726&88.8&\textbf{0.194}&\textbf{2.8}\\
    \midrule
    \multirow{6}*{\shortstack{ResNet100}} & LIRA~\cite{doan2021lira} &0.769&94.2&0.540&47.7\\
    & FTrojan~\cite{wang2021backdoor} &0.771&94.2&0.548&48.8\\
    &Blended~\cite{chen2017targeted}&0.741&91.0&0.698&81.0\\
    &BadNets~\cite{gu2017BadNets}&0.596&55.0&0.500&32.0\\
    & Our BAvAFT~\cite{yu2023backdoor} & 0.770&93.8&0.528&45.3\\
    & Our BAvAFT+Trans & 0.769&94&\textbf{0.308}&\textbf{10.8}\\
    \midrule
    \multirow{6}*{MobileFaceNet} & LIRA~\cite{doan2021lira} &0.677&86.2&0.441&23.3\\
    & FTrojan~\cite{wang2021backdoor} &0.680&86.3&0.448&25.0\\
    &Blended~\cite{chen2017targeted}&0.644&79.0&0.591&64.0\\
    &BadNets~\cite{gu2017BadNets}&0.535&50.0&0.436&28.0\\
    & Our BAvAFT~\cite{yu2023backdoor} & 0.677&86.0&0.439&25.7\\
    & Our BAvAFT+Trans & 0.678&86.5&\textbf{0.333}&\textbf{7.2}\\
    \bottomrule
    \end{tabular}
    }
    \label{face_cross}
\vspace{-2mm}
\end{table}

\begin{table}[t]\footnotesize
    \setlength\tabcolsep{4.9pt}
    \centering
    \caption{
    Attack performance for our backdoored model with multiple triggers.
    }
    \vspace{-3mm}
    \scalebox{0.9}{
    \begin{tabular}{ c || c | c | c | c  }
    \toprule
    Attack Type & BPP attack & PSNR attack & Car To Road & Face Recognition \\
    (Metric) & ({BPP} $\uparrow$) & (PSNR $\downarrow$) & (ASR $\uparrow$) & (Sim. $\downarrow$) \\
    \midrule
    Performance & 12.392 & 4.256 & 89.2 & 0.168\\
    \bottomrule
    \end{tabular}
    }
    \label{table4}
    \vspace{-2mm}
\end{table}

Therefore, The updated attack objective terms for attacking semantic segmentation or face recognition are given below:
\begin{footnotesize}
\begin{equation}
\begin{split}
\mathcal{L}_{AO}^{SS} &= \mathcal{L}_{CE}[\eta({g(\bm{x})}),g(\mu \cdot f({\bm{x_p}}))+(1 - \mu) \cdot g(f({\bm{x}}))] \\
&- \gamma \cdot \mathcal{L}_{CE}[\tau({g(\bm{x})}),g(f({\bm{x_p}})))], \\
\mathcal{L}_{AO}^{FR} &= Cos[\eta({g(\bm{x})}),g(\mu \cdot f({\bm{x_p}}))+(1 - \mu) \cdot g(f({\bm{x}}))], 
\label{transfer_formula}
\end{split}
\end{equation}
\end{footnotesize}
where $\mu$ is randomly sampled from a uniform distribution $U[\frac{1}{3},\frac{2}{3}]$, and $\tau({g(\bm{x})})$ replaces the targeted class with the unwanted class in $\eta({g(\bm{x})})$.
The experimental results presented in Table~\ref{segmentation_cross_model},~\ref{segmentation_cross_domain},~\ref{face_cross} demonstrate the effectiveness and transferability of our proposed BAvAFT+Trans attack. 
By employing specific optimization techniques to {improve} the attack's transferability, we achieve consistent attack performance across various domain data and different model backbones.

\begin{figure}[t]
    \centering
    \begin{minipage}{0.99\linewidth}
    \centerline{{\includegraphics[width=1.0\linewidth]{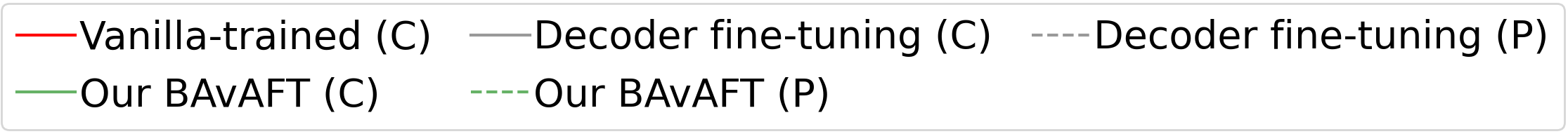}}}
    \end{minipage}
    \begin{minipage}{0.325\linewidth}
    \centerline{{\includegraphics[width=1\linewidth]{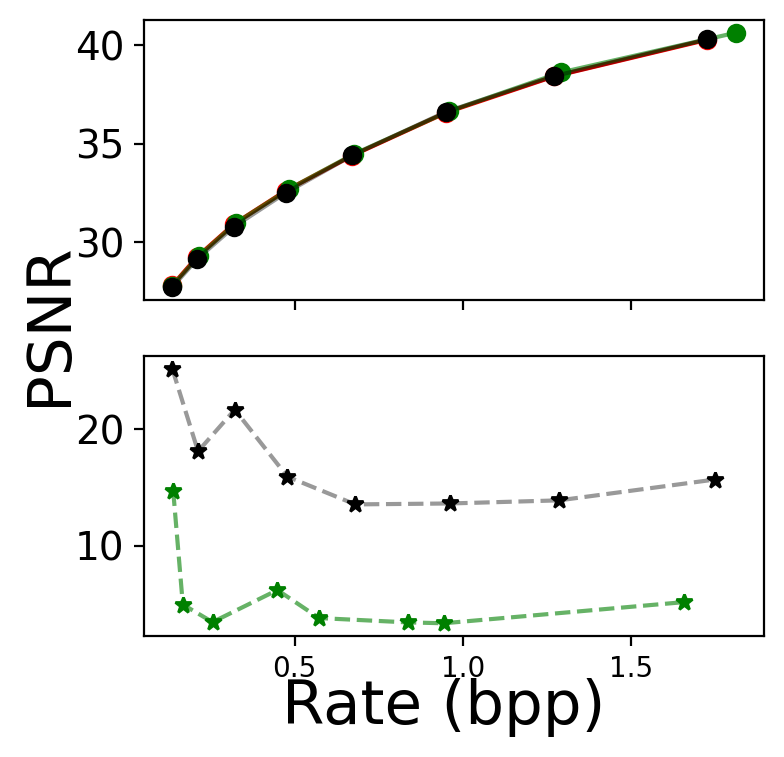}}}
    \vspace{-2mm}
    \centerline{\footnotesize{AE-Hyperprior~\cite{balle2018variational}}}
    \vspace{2mm}
    \end{minipage}
    \begin{minipage}{0.325\linewidth}
    \centerline{{\includegraphics[width=1\linewidth]{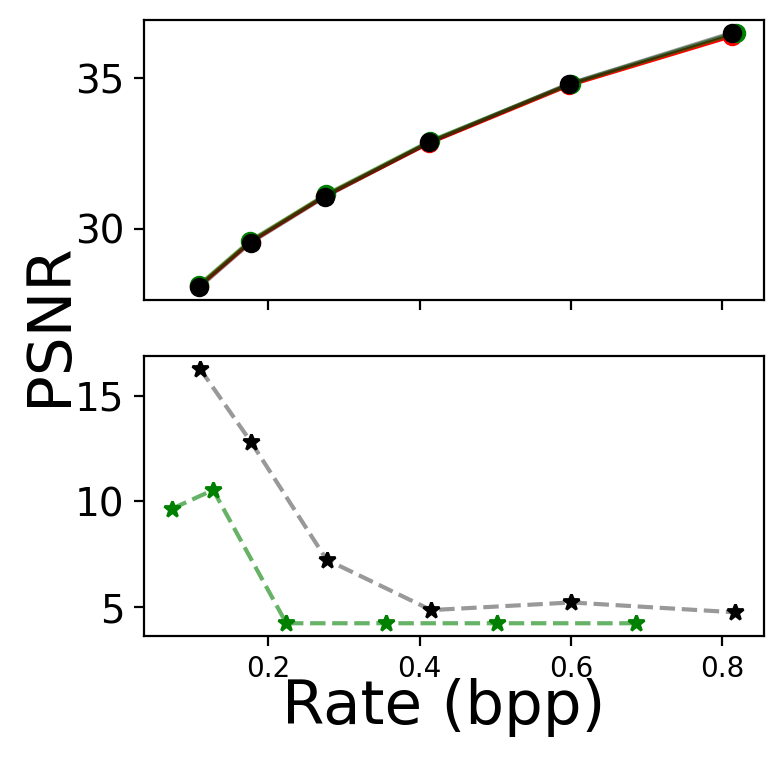}}}
    \vspace{-2mm}
    \centerline{\footnotesize{Cheng-Anchor~\cite{cheng2020learned}}}
    \vspace{2mm}
    \end{minipage}
    \begin{minipage}{0.325\linewidth}
    \centerline{{\includegraphics[width=1\linewidth]{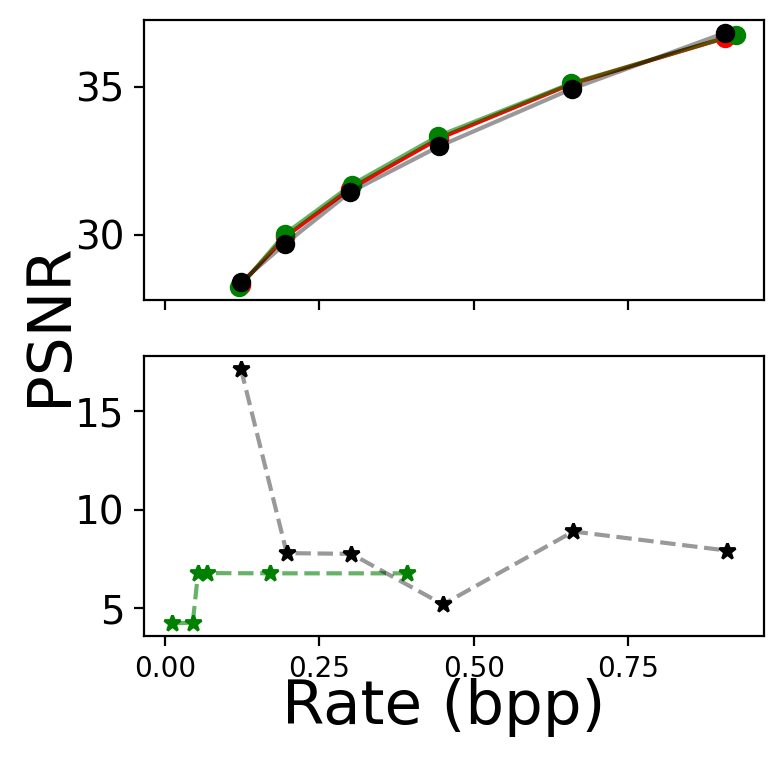}}}
    \vspace{-2mm}
    \centerline{\footnotesize{STF~\cite{zou2022devil}}}
    \vspace{2mm}
    \end{minipage}
    \vspace{-5mm}
    \caption{{Peformance of PSNR attack with decoder fine-tuning. }}
    \vspace{-2mm}
    \label{decoder_psnr}
\end{figure}

\begin{table}[t]\footnotesize
    \setlength\tabcolsep{0.3pt}
    \centering
    \caption{
    {Attack performance of our BAvFT and the decoder fine-tuning.
    }}
    \vspace{-3mm}
    \scalebox{0.9}{
    \begin{tabular}{ c || c | c | c | c  }
    \toprule
    Attack Type~$\rightarrow$ & BPP attack & PSNR attack & Car To Road & Face Recognition \\
    (Metric) & ({BPP} $\uparrow$) & (PSNR $\downarrow$) & (ASR $\uparrow$) & (Sim. $\downarrow$) \\
    \midrule
    BAvFT (encoder fine-tuning)& \textbf{11.18} & \textbf{6.18} & \textbf{94.6} & \textbf{0.407}\\
    Decoder fine-tuning & - & 8.53 & 61.2 & 0.674\\
    \bottomrule
    \end{tabular}
    }
    \label{decoder_attack}
    \vspace{-2mm}
\end{table}

Table~\ref{segmentation_cross_model} shows that our attack remains {powerful} when targeting different model backbones. Regardless of the specific model architecture used in the downstream semantic segmentation task, our BAvAFT+Trans attack consistently misleads the model, proving its robustness and adaptability to different model configurations.
Similarly, in Table~\ref{segmentation_cross_domain}, we observe that our attack maintains its effectiveness when transferring to {data in different domains}, such as CamVid and KiTTi datasets. 
This indicates that our attack is not limited to a specific dataset and can successfully target semantic segmentation models across various datasets, making it more practical and applicable in real-world scenarios.
Moreover, Table~\ref{face_cross} demonstrates that our BAvAFT+Trans attack can effectively protect the identity information of facial images across different model backbones in the face recognition task. This further validates the versatility and power of our proposed optimization techniques to improve the {attacking} transferability.

Overall, the results in these tables confirm that our BAvAFT+Trans attack is capable of maintaining its effectiveness and consistency in diverse settings, making it a strong candidate for practical backdoor attacks in various computer vision tasks.

\begin{figure*}[t]
\centering
\includegraphics[width=0.99\linewidth]{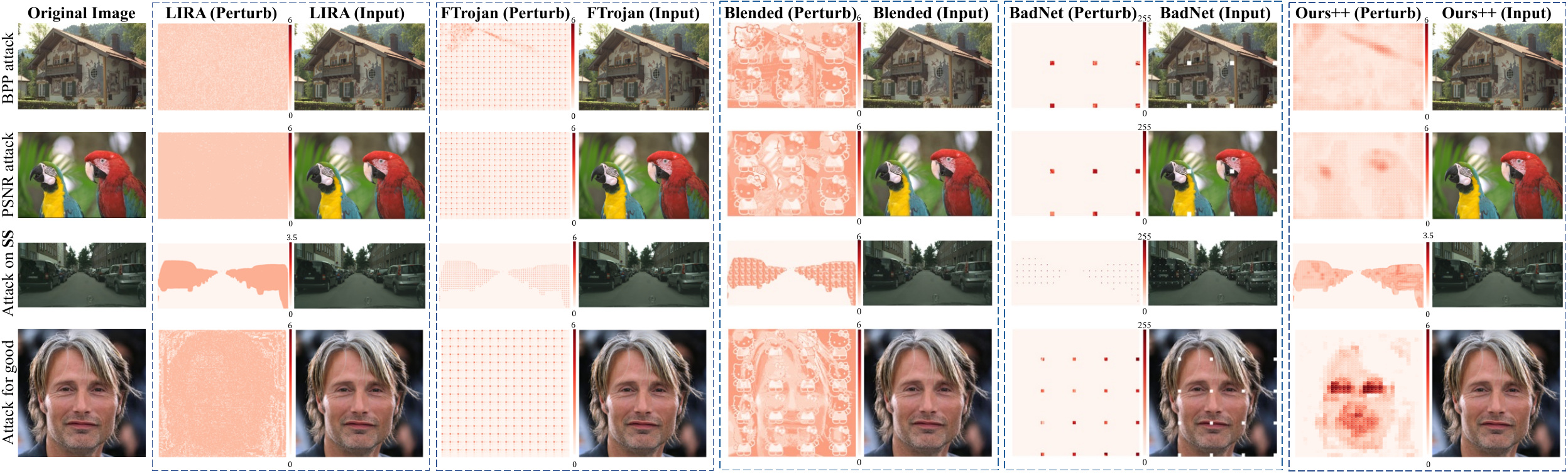}
\vspace{-4.0mm}
\caption{
{Visual results of trigger (perturbations added to the input $x_p-x$) and the poisoned input for LIRA~\cite{doan2021lira}, FTrojan~\cite{wang2021backdoor}, Blended~\cite{chen2017targeted}, BadNets~\cite{gu2017BadNets} and Our BAvAFT++ (denoted as Ours++). 
We show the average value across 3 channels of the absolute value for the trigger here.
}}
\vspace{-2mm}
\label{figure8}
\end{figure*}

\subsection{Backdoor-injected model with multiple triggers}
We have shown the effectiveness of our proposed backdoor attack for each attack objective in the above experiments.
In the end, we show the experiment of attacking with multiple triggers as shown in Section~\ref{multiple_triggers}. 
Here, we train the encoder and four trigger injection models with corresponding attack objectives, including: 
1) bit-rate (BPP) attack; 
2) quality reconstruction (PSNR) attack; 
3) downstream semantic segmentation (targeted attack with Car To Road).
3) attacking face recognition.
Hyperparameters and auxiliary dataset ${\mathcal{T}_a}$ correspond to the aforementioned experiments.
we select the Cheng-Anchor with {the quality level} 3 as the compression method. 
The attack performance of the victim model is presented in Table~\ref{table4}.
For reference, the PSNR/{BPP} of {the} vanilla-trained model and our proposed model on Kodak dataset are {\footnotesize{32.85/0.412}} and {\footnotesize{32.41/0.390}}, respectively.
The results demonstrate that our backdoor attack is effective for all attack objectives, and has {a low-performance impact} on clean images. 

{
\subsection{Attacks on other parts of the compression model}\label{decoder_attack_section}
The encoder and decoder of an image compression system are commonly distributed in different locations.
For example, the bitstream can be generated by the encoder at the cloud side, while the bitstream is decoded at the client side.
In the main paper, we primarily focus on attacking the encoding stage by introducing a backdoored encoder. In this section, we examine additional scenarios. Given that the entropy module is present in both the encoding and decoding stages, attacking it directly is impractical due to the need for more extensive access. 
Therefore, we particularly explore the possibility of fine-tuning the decoder for the decoding process.

Since the BPP metric is assessed during the encoding process, fine-tuning the decoder cannot facilitate a BPP attack. Therefore, we focus our experiments on the PSNR attack and attacks on downstream tasks. 
In Figure~\ref{decoder_psnr}, we compare the effectiveness of decoder fine-tuning and our BAvFT (\textit{i.e.,} encoder fine-tuning) for the PSNR attack. The results show that our BAvFT achieves superior attack performance.
Table~\ref{decoder_attack} presents additional quantitative results for both decoder fine-tuning and BAvFT across all attack types. While decoder fine-tuning does result in some attack success, its performance is notably poorer compared to our BAvFT approach. The decline in attack effectiveness may be attributed to the potential weakening of the trigger when it passes through the unaltered encoder before reaching the backdoored decoder, thus reducing the attack's overall impact.
In conclusion, targeting the encoding stage (\textit{i.e.,} encoder fine-tuning) proves to be a more effective strategy for launching a successful attack.
}

\begin{figure}[t]
\centering
\includegraphics[width=1\linewidth]{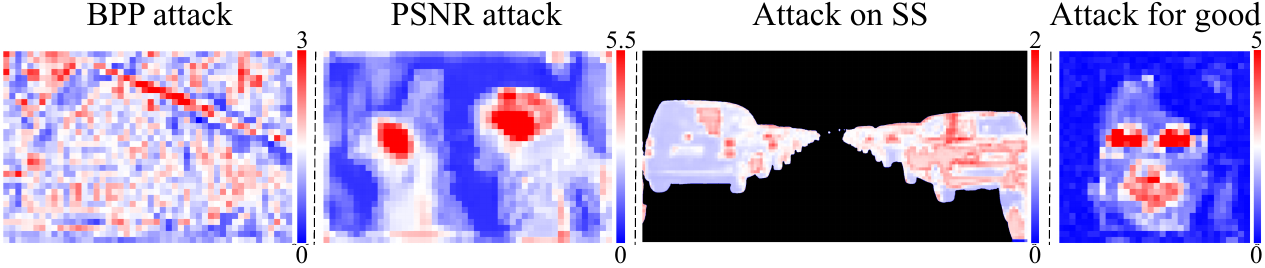}
\vspace{-8mm}
\caption{
Visualization of patch-wise {weights} in our proposed trigger injection. 
We show the average value across 3 channels of the absolute value.
The sample for each attack is same as shown in Figure~\ref{figure8}.
}
\label{figure7}
\vspace{-2mm}
\end{figure}

\section{Analysis on Trigger}
\noindent \textbf{Comparison with LIRA~\cite{doan2021lira}, FTrojan~\cite{wang2021backdoor}, Blended~\cite{chen2017targeted} and BadNets~\cite{gu2017BadNets}.} Figure~\ref{figure8} {shows} the visual results of the triggers (perturbations added to the input $x_p-x$) and the corresponding poisoned inputs for LIRA, FTrojan, {and} our BAvAFT+Trans. A comparison between all methods reveals that our proposed trigger in the DCT domain generates more sparse and diverse perturbations in the spatial domain. This sparsity and diversity {contribute} to making our attack more imperceptible and stealthy.

Furthermore, our attack demonstrates a more adaptive trigger generation mechanism. In the example of attacking the face recognition task, it can be observed that the triggers adjust themselves to selectively add perturbations to the key areas of facial images (\textit{e.g.,} eye, nose, and mouth). 
This targeted perturbation placement enables our attack to mislead the face recognition model with minimal perturbations on the attacked output. 
It should be noted that in the attack on semantic segmentation, a mask is used to guide the trigger, ensuring that the perturbations are primarily applied to the regions of interest without affecting other areas.

Overall, the visual results highlight the effectiveness and adaptability of our proposed attack method, demonstrating its {capacity} to generate subtle and targeted perturbations to achieve the desired attack objectives with minimal visual impact.

\noindent \textbf{Patch-wise weight in our attack.} Figure~\ref{figure7} provides a visualization of the patch-wise {weights} in our proposed trigger injection method. The results demonstrate the adaptability of our attack by assigning higher weights to the vulnerable areas of the input image. 
This adaptive weighting mechanism allows our attack to focus on the critical regions, improving the attacking performance by {applying} the perturbations to the areas that are more susceptible to triggering the desired behavior in the victim model.

By assigning higher weights to vulnerable areas, our attack can effectively optimize the trigger placement and maximize the impact on the victim model while minimizing perturbations in non-critical regions. This targeted approach {improves} the attack's effectiveness while reducing the visibility of perturbations, making the attack more imperceptible and stealthy.
The visualization of patch-wise weights provides insights into how our attack dynamically adjusts the trigger based on the vulnerability of different image regions, resulting in a more efficient and effective attack.

\section{Conclusions}
In this paper, we have presented a novel backdoor attack against learned image compression models using an adaptive frequency trigger. Our attack focuses on modifying the parameters of the encoder, making it practical and applicable in real-world scenarios.
We have conducted a thorough investigation and proposed multiple attack objectives, including l{ow-level quality and task-driven measures}, such as the performance of downstream computer vision tasks. This comprehensive exploration allows us to evaluate and optimize the {attacking} effectiveness from different perspectives.
Furthermore, we consider several advanced
scenarios. We evaluate the resistance of the proposed backdoor attack to the defensive pre-processing methods and then propose a
two-stage training schedule along with the design of robust frequency selection, which can {significantly improve} the resistance. To strengthen both the cross-model and cross-domain transferability on attacking downstream CV tasks, we propose to shift the classification boundary in the attack loss during training. 
Besides, we have demonstrated the capability of injecting multiple triggers with specific attack objectives into a single victim model. This multi-trigger approach enables us to target different behaviors and manipulate the model's output based on the specific trigger applied.
Overall, our work contributes to the understanding and advancement of backdoor attacks in the context of learned image compression. The proposed adaptive frequency trigger and the exploration of different {attacking} objectives provide valuable insights for developing more robust defense mechanisms and raising awareness about potential security vulnerabilities in image compression systems.

{\small
\bibliographystyle{ieee_fullname}
\bibliography{egbib}
}

\vspace{-10mm}
\begin{IEEEbiography}[{\includegraphics[width=1in,height=1.25in,clip,keepaspectratio]{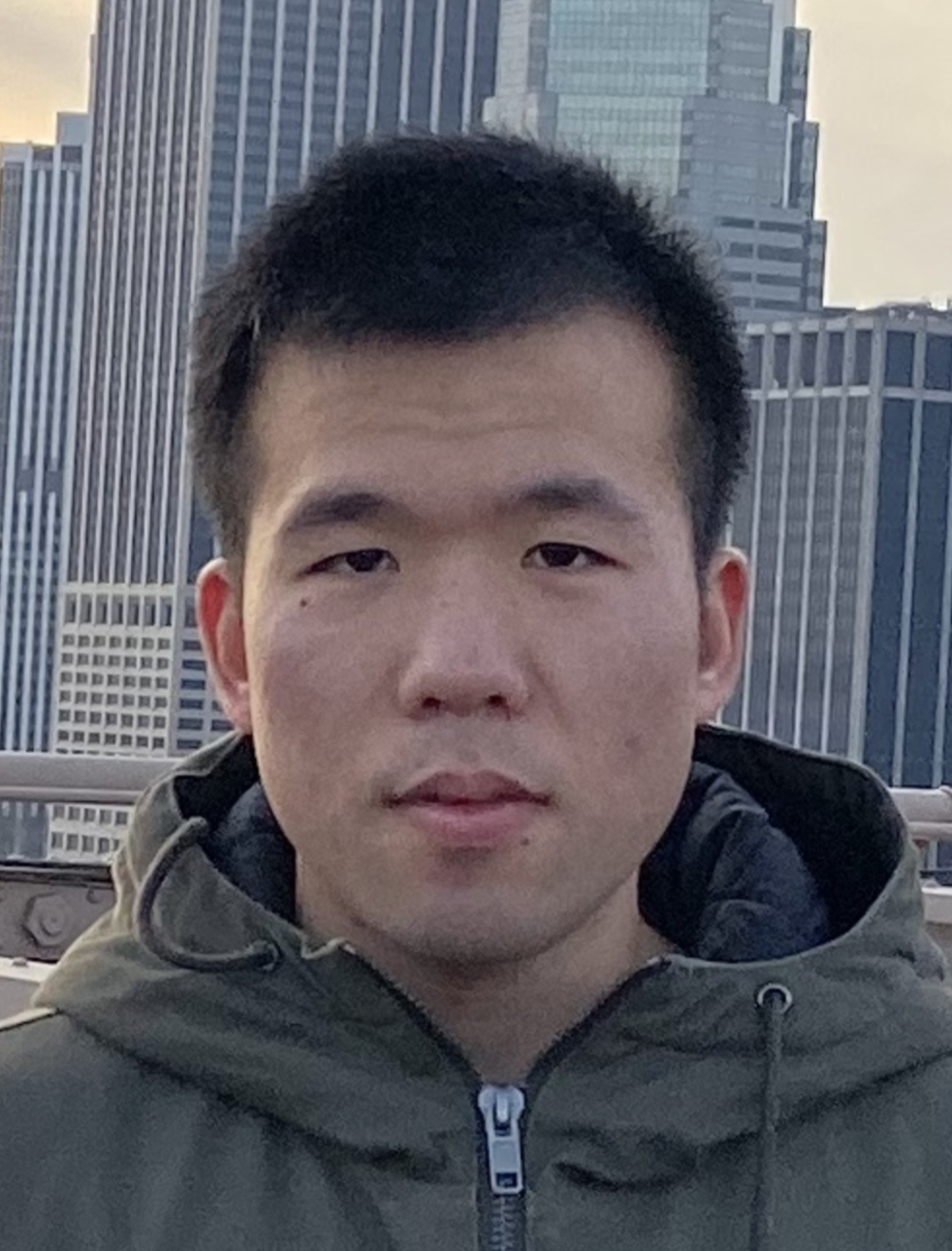}}]
    {Yi Yu} 
    received the B.S. degree from the Department of Automation, Tsinghua University, China and the M.S, degree from the Department of Electrical and Computer Engineering, University of California San Diego, United States, in 2019 and 2021, respectively. He is currently working toward the PhD degree in the ROSE Lab, Interdisciplinary Graduate Programme, Nanyang Technology University, Singapore.
    His research interests focus on trustworthy ml and AI security. 
\end{IEEEbiography}

\vspace{-12mm}
\begin{IEEEbiography}[
    {\includegraphics[width=1in,height=1.25in,clip,keepaspectratio, trim=0 5 0 10]{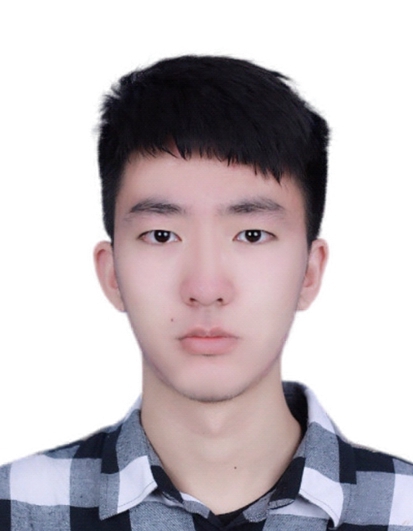}}]{Yufei Wang}
    received the B.Eng degree from the School of Computer Science and Engineering, University of Electronic Science and Technology of China, in 2020, and he is currently a Ph.D. candidate in ROSE Lab, Nanyang Technological University, Singapore. His research interests focus on image restoration and the generalization ability of deep neural network models. He is the recipient of the SDSC Dissertation Research Fellowship in 2022. 
\end{IEEEbiography}

\vspace{-12mm}
\begin{IEEEbiography}[{\includegraphics[width=1in,height=1.25in,clip,keepaspectratio]{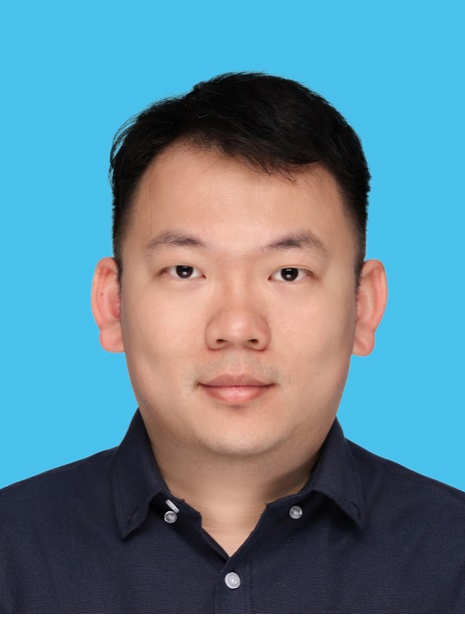}}]
    {Wenhan Yang} (M'18) received the B.S degree and Ph.D. degree (Hons.) in computer science from Peking University, Beijing, China, in 2012 and 2018. He is currently an associate researcher with Pengcheng Laboratory, China. His current research interests include image/video processing/restoration, bad weather restoration, human-machine collaborative coding. 
    He received the 2023 IEEE Multimedia Rising Star Runner-Up Award, the IEEE ICME-2020 Best Paper Award, the IEEE CVPR-2018 UG2 Challenge First Runner-up Award, and the MSA-TC Best Paper Award of ISCAS 2022. 
\end{IEEEbiography}

\vspace{-10mm}
\begin{IEEEbiography}[{\includegraphics[width=1in,height=1.25in,clip,keepaspectratio, trim=0 10 0 10]{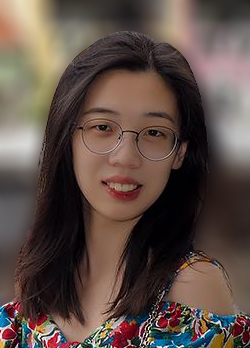}}]{Lanqing Guo} 
 is a postdoc research fellow at The University of Texas at Austin, USA. She earned her Ph.D. in Electrical and Electronic Engineering from Nanyang Technological University, Singapore, where she graduated with the Best Thesis Award. Prior to that, she received her B.Eng. in Software Engineering from Wuhan University, China. Her research interests include 2D/3D image processing and generation, computational imaging, and computer vision.
\end{IEEEbiography}

\vspace{-10mm}
\begin{IEEEbiography}[{\includegraphics[width=1in,height=1.25in,clip,keepaspectratio]{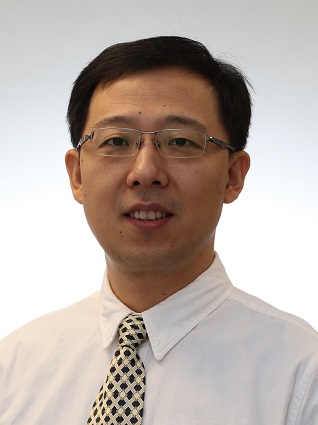}}]{Dr. Shijian Lu} is an Associate 
    Professor in the School of Computer Science and Engineering, Nanyang Technological University in Singapore. He obtained the PhD from the Electrical and Computer Engineering Department, National University of Singapore. Dr Shijian Lu’s major research interests include image and video analytics, visual intelligence, and machine learning. 
    He has published more than 100 international refereed journal and conference papers and co-authored over 10 patents in these research areas. 
    He was also a top winner of the ICFHR2014 Competition on Word Recognition from Historical Documents, ICDAR 2013 Robust Reading Competition, etc. 
    Dr Lu is an Associate Editor for the journal Pattern Recognition (PR) and Neurocomputing. 
    He has also served in the program committee of several conferences, e.g., the Senior Program Committee of the International Joint Conferences on Artificial Intelligence (IJCAI) 2018 – 2021, etc.
\end{IEEEbiography}

\begin{IEEEbiography}[{\includegraphics[width=1in,height=1.25in,clip,keepaspectratio, trim=0 10 0 10]{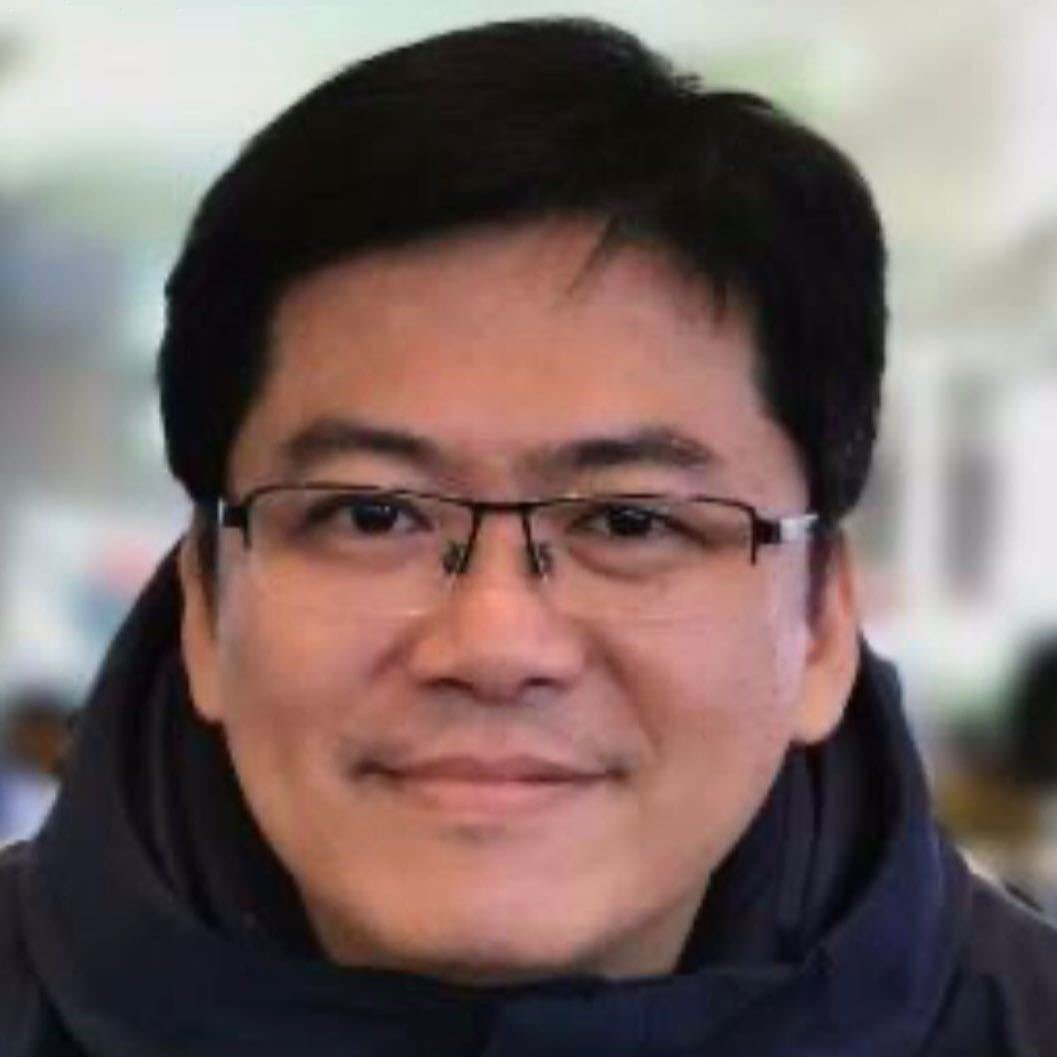}}]{Ling-Yu Duan}
    Ling-Yu Duan is a Full Professor with the National Engineering Laboratory of Video Technology (NELVT), School of Electronics Engineering and Computer Science, Peking University (PKU), China, and has served as the Associate Director of the Rapid-Rich Object Search Laboratory (ROSE), a joint lab between Nanyang Technological University (NTU), Singapore, and Peking University (PKU), China since 2012. He is also with Peng Cheng Laboratory, Shenzhen, China, since 2019. He received the Ph.D. degree in information technology from The University of Newcastle, Callaghan, Australia, in 2008. His research interests include multimedia indexing, search, and retrieval, mobile visual search, visual feature coding, and video analytics, etc. 
    He has published about 200 research papers. 
    He received the IEEE ICME Best Paper Award in 2019/2020, the IEEE VCIP Best Paper Award in 2019, and EURASIP Journal on Image and Video Processing Best Paper Award in 2015
    , the Ministry of Education Technology Invention Award (First Prize) in 2016, the National Technology Invention Award (Second Prize) in 2017, China Patent Award for Excellence (2017), the National Information Technology Standardization Technical Committee "Standardization Work Outstanding Person" Award in 2015. 
    He was a Co-Editor of MPEG Compact Descriptor for Visual Search (CDVS) Standard (ISO/IEC 15938-13) and MPEG Compact Descriptor for Video Analytics (CDVA) standard (ISO/IEC 15938-15). 
\end{IEEEbiography}

\vspace{-5mm}
\begin{IEEEbiography}[{\includegraphics[width=1in,height=1.25in,clip,keepaspectratio]{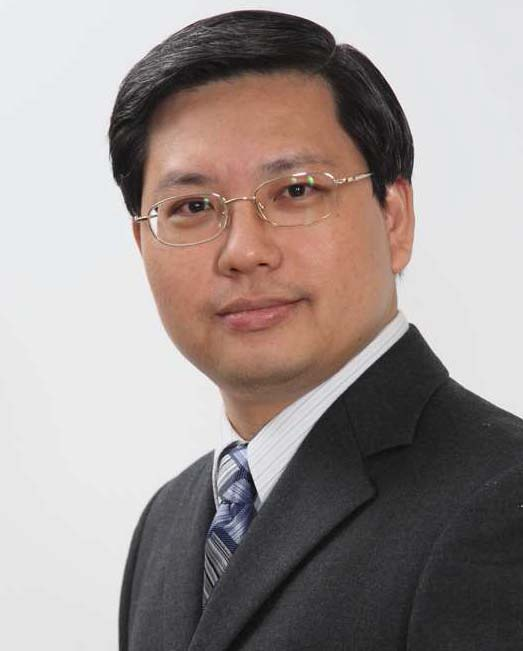}}]
    {Yap-Peng Tan} (Fellow, IEEE) received the B.S. degree from National Taiwan University, Taipei, Taiwan, in 1993, and the M.A. and Ph.D. degrees from Princeton University, Princeton, NJ, in 1995 and 1997, respectively, all in electrical engineering.  He  is  currently  a  Full  Professor  and  the Chair of the School of Electrical and Electronic Engineering, Nanyang Technological University, Singapore.  His   research  interests  include image and video processing, computer vision, pattern recognition, and data analytics. He served as an Associate Editor of the IEEE TRANSACTIONS ON CIRCUITS AND SYSTEMS FOR VIDEO TECH-NOLOGY,  IEEE  SIGNAL  PROCESSING  LETTERS, IEEE  TRANSACTIONS  ON  MULTIMEDIA,  and  IEEE  Access,  as well  as  an  Editorial  Board  Member  of  the  EURASIP  Journal  on  Advances in Signal Processing and EURASIP Journal on Image and Video Processing. He was the Technical Program Co-Chair of the 2015 IEEE International Conference on Multimedia and Expo (ICME 2015) and the 2019 IEEE International Conference on Image Processing (ICIP 2019),and the General Co-Chair of the 2010 IEEE International Conference on Multimedia and Expo (ICME 2010) and the 2015 IEEE International Conference  on  Visual  Communications  and  Image  Processing  (VCIP 2015).
\end{IEEEbiography}

\vspace{-5mm}
\begin{IEEEbiography}[{\includegraphics[width=1in,height=1.25in,clip,keepaspectratio]{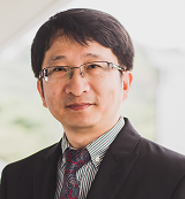}}]
    {Alex C. Kot}  (Life Fellow, IEEE) has been with the Nanyang Technological University, Singapore since 1991. He was Head of the Division of Information Engineering and Vice Dean Research at the School of Electrical and Electronic Engineering. 	Subsequently, he served as Associate Dean for College of Engineering for eight years. He is currently Professor and Director of Rapid-Rich Object SEarch (ROSE) Lab and NTU-PKU Joint Research Institute. He has published extensively in the areas of signal processing, biometrics, image forensics and security, and computer vision and machine learning. 	
    Dr. Kot served as Associate Editor for more than ten journals, mostly for IEEE transactions. He has served the IEEE SP Society in various capacities such as the General Co-Chair for the 2004 IEEE International Conference on Image Processing and the Vice-President for the IEEE Signal Processing Society. He received the Best Teacher of the Year Award and is a co-author for several Best Paper Awards including ICPR, IEEE WIFS and IWDW, CVPR Precognition Workshop and VCIP. He was elected as the IEEE Distinguished Lecturer for the Signal Processing Society and the Circuits and Systems Society. He is a Fellow of IES, a Fellow of IEEE, and a Fellow of Academy of Engineering, Singapore.
\end{IEEEbiography}

\end{document}